\documentclass[10pt,journal,compsoc]{IEEEtran}

\usepackage{epsfig}
\usepackage{graphicx}
\usepackage{amsmath}
\usepackage{amssymb}
\usepackage{subfig}
\usepackage{xspace}
\usepackage{bbold}
\usepackage{xcolor}
\usepackage{multirow}
\usepackage{url}
\usepackage{xspace}
\usepackage{review}

\newcommand{\vct}[1]{\ensuremath{\boldsymbol{#1}}}

\newcommand{\set}[1]{\ensuremath{\mathcal{#1}}}
\newcommand{\con}[1]{\ensuremath{\mathsf{#1}}}

\newcommand{\ind}[1]{\ensuremath{\mathbb 1_{#1}}}

\newcommand{\myparagraph}[1]{\vspace{5pt}\noindent\textbf{#1}}
\DeclareMathOperator{\Var}{Var}
\DeclareMathOperator{\E}{E}
\DeclareMathOperator{\DA}{DA}
\DeclareMathOperator{\mDA}{mDA}

\definecolor{darkgreen}{rgb}{0 0.5 0}

\newcommand{\eg}{\emph{e.g.}\xspace}

\setrevision{0}

\definecolor{darkgreen}{gray}{0.0}

\makeatletter
\DeclareRobustCommand\onedot{\futurelet\@let@token\@onedot}
\def\@onedot{\ifx\@let@token.\else.\null\fi\xspace}

\def\eg{\emph{e.g}\onedot} 
\def\ie{\emph{i.e}\onedot}

\def\etal{\emph{et al}\onedot}
\makeatother

\ifCLASSOPTIONcompsoc
  \usepackage[nocompress]{cite}
\else
  \usepackage{cite}
\fi
\ifCLASSINFOpdf
\else
\fi
\begin{document}
%
\title{
Inferring Latent Domains for \\ Unsupervised Deep Domain Adaptation}
%
%
%
%

\author{Massimiliano~Mancini,~\IEEEmembership{}
        Lorenzo~Porzi,~\IEEEmembership{}
        Samuel~Rota~Bul\'o,~\IEEEmembership{}
        
        Barbara~Caputo~\IEEEmembership{}
        and~Elisa~Ricci~\IEEEmembership{}
\IEEEcompsocitemizethanks{\IEEEcompsocthanksitem M. Mancini is with Department of Computer, Control, and Management Engineering, Sapienza University of Rome, Rome, Italy.(Email: mancini@diag.uniroma1.it)
\IEEEcompsocthanksitem L. Porzi and S. Rota Bul\'o are with Mapillary Research, Graz, Austria.(Email: \{lorenzo,samuel\}@mapillary.com)
\IEEEcompsocthanksitem B. Caputo is with DAUIN Department of Control and Computer  Engineering  of  Politecnico  di  Torino, Turin, Italy. (Email: barbara.caputo@polito.it)
\IEEEcompsocthanksitem M. Mancini and B. Caputo are with Italian Institute of Technology, Turin, Italy.
\IEEEcompsocthanksitem M. Mancini and E. Ricci are with Fondazione Bruno Kessler, Trento, Italy. (Email: eliricci@fbk.eu)
\IEEEcompsocthanksitem E. Ricci is with Department of Information Engineering and Computer Sciene, University of Trento, Trento, Italy}
}

\IEEEtitleabstractindextext{%
\begin{abstract}
Unsupervised Domain Adaptation (UDA) refers to the problem of learning a model in a target domain where labeled data are not available by leveraging information from annotated data in a source domain. 
Most deep UDA approaches operate in a single-source, single-target scenario, \textit{i.e.} they assume that the source and the target samples arise from a single distribution. However, in practice most datasets can be regarded as mixtures of multiple domains. In these cases, exploiting traditional single-source, single-target methods for learning classification models may lead to poor results. Furthermore, it is often difficult to  
provide the domain labels for all data points, \ie latent domains should be automatically discovered.
This paper introduces a novel deep architecture which addresses the problem of UDA by automatically discovering latent domains in visual datasets and exploiting this information to learn robust target classifiers.
Specifically, our architecture is based on 
two main components,  
\textit{i.e.} 
a side branch that automatically computes the assignment of each sample to its latent domain and novel layers that exploit domain membership information to appropriately align the distribution of the CNN internal feature representations to a reference distribution.
We evaluate our approach on publicly available benchmarks, showing that it outperforms state-of-the-art domain adaptation methods.
\end{abstract}

}

\begin{titlepage}
\null
\vfill
\renewcommand{\fboxsep}{10pt}
\fbox{\Large\begin{minipage}{\columnwidth}
\textbf{Disclaimer:}

This work has been accepted for publication in the IEEE Transactions on Pattern Analysis and Machine Intelligence:\vspace{4pt}
\newline
doi:     10.1109/TPAMI.2019.2933829
\newline
link:    https://ieeexplore.ieee.org/document/8792192
\newline
\newline
\textbf{Copyright:} 
\newline
\copyright~2019 IEEE. Personal use of this material is permitted. Permission from IEEE must be obtained for all other uses,  in  any  current  or  future  media,  including  reprinting/  republishing  this  material  for  advertising  or promotional purposes, creating new collective works, for resale or redistribution to servers or lists, or reuse of any copyrighted component of this work in other works.
\newline
\end{minipage}}
\vfill
\clearpage
\end{titlepage}
\maketitle

\IEEEdisplaynontitleabstractindextext

%
\IEEEpeerreviewmaketitle


%
%
%
%
\section{Introduction}
\IEEEPARstart{I}{n} the last few years, deep neural networks have enabled unprecedented progresses in many fields of artificial intelligence including computer vision, speech recognition, natural language processing and robotics. While deep models are extremely powerful by automatically learning discriminative representations of input data, one major limitation is that huge amounts of labeled data are typically required for training. However, compiling large scale annotated datasets is a tedious and extremely costly operation. To address this issue, over the years the research community has proposed different strategies, such as semi-supervised learning, transfer learning and domain adaptation.


Domain adaptation methods, in particular, are specifically designed in order to transfer knowledge from a \textit{source} domain to a domain of interest, \ie the \textit{target} domain. 
In the specific case of Unsupervised Domain Adaptation (UDA), labeled data are only available in the source domain, while no annotations are provided for the target samples.
Typically, knowledge transfer is achieved by 
learning domain-agnostic models or invariant feature representations. 
The problem of UDA has been widely studied and both theoretical results \cite{ben2010theory,mansour2009domain} and several algorithms have been developed, both considering  shallow models \cite{huang2006correcting,gong2013connecting,gong2012geodesic,long2013transfer,fernando2013unsupervised} and deep architectures \cite{long2015learning,tzeng2015simultaneous,ganin2014unsupervised,long2016unsupervised,ghifary2016deep,carlucci2017autodial,bousmalis2016domain}. While deep neural networks tend to produce more transferable and domain-invariant features with respect to shallow models, previous works have shown that the domain shift is only alleviated but not entirely removed \cite{donahue2014decaf}.

\begin{figure}[t]
  \centering
  \includegraphics[width=\columnwidth,trim={0cm 0cm 0cm 0cm},clip]{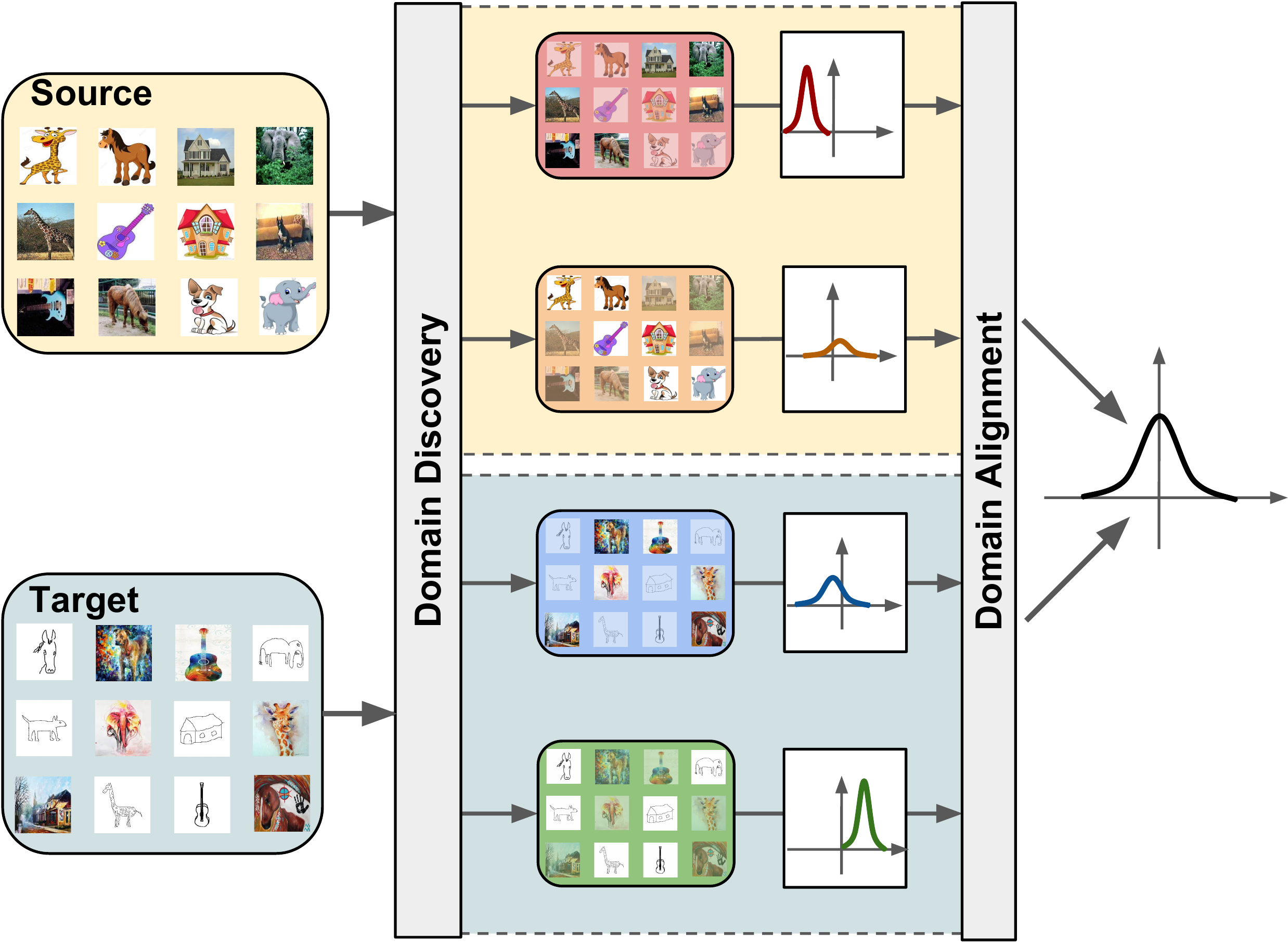}
  \caption{The idea behind the proposed framework. 
We introduce a novel deep architecture which, given a set of images, automatically discover multiple latent 
domains
and use this information to align the distributions of the internal CNN feature representations of sources and target domains
for the purpose of domain adaptation. In this way, more accurate target classifiers can be learned. Image better seen at magnification.
  }
  \vspace{-15pt}
  \label{fig:teaser}
\end{figure}

Most previous works on UDA focus on a single-source and single-target scenario. However, in many computer vision applications 
labeled training data are often generated from multiple distributions, \ie there are multiple source domains. Examples of multi-source DA problems arise when the source set corresponds to images taken with different cameras, collected from the web or associated to multiple points of views. In these cases, a naive application of single-source domain adaptation algorithms would not suffice, leading to poor results. Analogously, target samples may arise from more than a single distribution and learning multiple target-specific models may improve significantly the performance. Therefore, in the past several research efforts have been devoted to develop domain adaptation methods considering multiple source and target domains \cite{mansour2009domain,duan2009domain,sun2011two,xu2018deep}. However, these approaches assume that the multiple domains are known. A more challenging problem arises when training data correspond to latent domains, \ie we can make a reasonable estimate on the number of source and target domains available, but we have no information, or only partial, about domain labels.
To address this problem, known in the literature as \emph{latent domain discovery}, previous works have proposed methods which simultaneously discover hidden source domains and use them to learn the target classification models~\cite{hoffman2012discovering,gong2013reshaping,xiong2014latent}.

{This paper introduces the first approach based on deep neural networks able to automatically discover latent domains in multi-source, multi-target UDA setting.}
Our method is inspired by the recent works~\cite{carlucci2017autodial,carlucci2017just,mancini2018robust}, which revisit Batch Normalization (BN) layers \cite{ioffe2015batch} for the purpose of domain adaptation and generalization, introducing specific Domain Alignment layers (DA-layers). 
The main idea behind DA-layers is to cope with domain shift by aligning representations of source and target distributions to a reference Gaussian distribution. Our approach develops from the same intuition. However, to address the additional challenges of discovering and handling multiple latent domains, we propose a novel architecture which is able to (i) learn a set of assignment variables which associate source and target samples to a latent domain and (ii) exploit this information for aligning the distributions of the internal CNN feature representations and learn robust target classifiers (Fig.\ref{fig:teaser}).
Our experimental evaluation shows that the proposed approach alleviates the domain discrepancy and outperforms previous UDA techniques on popular benchmarks, such as Office-31~\cite{saenko2010adapting}, PACS \cite{li2017domain} and Office-Caltech~\cite{gong2012geodesic}.

To summarize, the contributions of this paper are threefold. Firstly, we present a novel deep learning approach for unsupervised domain adaptation which operates in a multi-source, multi-target setting. Secondly, we propose a novel architecture which is not only able to handle multiple domains, but also permits to automatically discover them by grouping source and target samples. Thirdly, our  experiments demonstrate that the  proposed framework is superior to 
many state of the art single- and multi-source/target UDA methods.

This paper extends our earlier work \cite{mancini2018boosting} in many aspects. In particular, we modify the deep architecture described in \cite{mancini2018boosting} for handling not only different source distributions but also multiple target domains. {We also develop a novel formulation of the loss function in \cite{mancini2018boosting} which allows not only to further boost the performances, but to also stabilize the estimates of the domain discovery branches, being more robust to mode collapsing issues (\eg all images assigned to a single domain).}  We also provide a more comprehensive review of related works. Finally, we significantly expand our experimental evaluation, considering more recent baseline methods and showing additional qualitative and quantitative results.

The remainder of this paper is organized as follows. We first  introduce related works in Section  \ref{sec:related} and then describe the proposed approach for latent domain discovery and unsupervised domain adaptation (Section \ref{sec:method}). The results of our extensive experimental evaluation are provided 
in Section \ref{sec:experiments}. We conclude the paper in Section \ref{sec:conclusions}.

\section{Related Work}
\label{sec:related}
In this section we review previous works on DA considering both methods based on hand-crafted features and more recent deep architectures, positioning our work in this context.

\myparagraph{DA methods with hand-crafted features.} Earlier DA approaches operate on hand-crafted features and attempt to reduce  
the discrepancy between the source and the target domains by adopting different strategies. For instance, instance-based methods \cite{huang2006correcting,yamada2012no,gong2013connecting} develop from the idea of learning classification/regression models by re-weighting source samples according to their similarity with the target data. A different strategy is exploited by feature-based methods, coping with domain shift by learning a common subspace for source and target data such as to obtain domain-invariant representations \cite{gong2012geodesic,long2013transfer,fernando2013unsupervised}. Parameter-based methods \cite{yang2007adapting} address the domain shift problem by discovering a set of shared weights between the source and the target models. However, they usually require labeled target data which is not always available.

While most earlier DA approaches focus on a single-source and single-target setting, some works have considered the related problem of learning classification models when the training data spans multiple domains \cite{mansour2009domain,duan2009domain,sun2011two}. The common idea behind these methods is that when source data arises from multiple distributions, adopting a single source classifier is suboptimal and improved performance can be obtained by leveraging information about multiple domains. However, these methods assume that the domain labels for all source samples are known in advance. In practice, in many applications the information about domains is hidden and latent domains must be discovered into the large training set.
Few works have considered this problem in the literature. Hoffman \etal \cite{hoffman2012discovering} address this task by modeling domains as  Gaussian distributions in the feature space and by estimating the membership of each training sample to a source domain using an iterative approach. 
Gong \etal \cite{gong2013reshaping} discover latent domains by devising a nonparametric approach which aims at
simultaneously achieving maximum distinctiveness among domains 
and ensuring that strong discriminative models are learned for each latent domain. In \cite{xiong2014latent} 
domains are modeled as manifolds and source images representations are learned decoupling information about semantic category and domain. By exploiting these representations the domain assignment labels are inferred using a mutual information based clustering method. 


\myparagraph{Deep Domain Adaptation.} Most recent works on DA consider deep architectures and robust domain-invariant features are learned using either supervised neural networks \cite{long2015learning,tzeng2015simultaneous,ganin2014unsupervised,ghifary2016deep,bousmalis2016domain,carlucci2017autodial}, deep autoencoders \cite{zeng2014deep} or generative adversarial networks \cite{bousmalis2016unsupervised,appleGAN}. Research efforts can be grouped in terms of the number of source domains available at training time.

In the single-source DA setting, we can identify two main 
 strategies. The first deals with \emph{features} and aims at learning deep domain invariant
representations. 
The idea is to introduce in the 
learning 
architecture different measures of domain
distribution shift at a single or multiple levels \cite{LongZ0J17,Sun:CORAL:AAAI16,carlucci2017autodial,carlucci2017just}
and then train the network to minimize these measures while also reducing a task-specific loss, for instance for classification or detection. 
In this way the network produces 
features invariant to the domain shift, but still discriminative
for the task at hand. 
Besides distribution evaluations, other domain shift measures used similarly 
are the error in the target sample reconstruction \cite{ghifary2016deep}, or 
various 
coherence metrics 
on the pseudo-labels assigned by the source models to the target data \cite{TRUDA-NIPS16_savarese, 
haeusser17,saito2017asymmetric}. Finally, a different
group of feature-based methods rely on adversarial loss functions \cite{tzeng2015simultaneous,ganin2016domain}. The method proposed in \cite{sankaranarayanan2017generate},
that push the network to be
unable to discriminate whether a sample coming from the source or 
from the target, 
is an interesting variant of \cite{ganin2016domain}, where the domain difference is still measured at the feature 
level but passing through an image reconstruction step.  
Besides integrating the domain discrimination objective
into end-to-end classification networks, 
it has 
also 
been shown that 
two-step networks may 
have practical advantages \cite{Hoffman:Adda:CVPR17,LOAD_ICRA}.
The second popular 
deep adaptive 
strategy focuses on \emph{images}.
The described adversarial logic that demonstrated its effectiveness for feature-based methods, has also
been extended to the goal of reducing the visual domain gap. Powerful GAN \cite{Goodfellow:GAN:NIPS2014} 
methods have been exploited to generate new images or perturb existing ones to resemble the visual style of a 
certain domain, thus reducing the discrepancy at pixel level \cite{Bousmalis:Google:CVPR17,appleGAN}.
Most of the works based on image adaptation aim at generating either target-like source images
or source-like target images, but it has been recently shown that integrating both the
transformation directions is highly beneficial \cite{russo17sbadagan}.

In practical applications one may be offered more than one source domain. This has triggered the study of multi-sources DA algorithms.
The multi-source setting 
was initially studied from a theoretical point of view, focusing on theorems indicating how to 
optimally sub-select the data to be used in learning the source models \cite{Crammer_JMLR08}, or 
proposing principled rules for combining the source-specific 
classifiers and obtain the ideal target class prediction \cite{mansour2009domain}. 
Several other works followed this direction in the shallow learning framework. 
When dealing with shallow-methods the na\"{\i}ve model learned by collecting all
the source data in single domain without any adaptation was usually showing low performance on the target.
It has been noticed that this behavior changes when moving to deep learning, where the larger number of 
samples as well as their variability supports generalization and usually provides good results on the target.
Only very recently two methods presented multi-source deep learning approaches that improve over this 
reference. The approach proposed in \cite{xu2018deep} builds over \cite{ganin2016domain}
by replicating the adversarial domain discriminator branch for each available source. Moreover
these discriminators are also used to get a perplexity score that indicates how the multiple
sources should be combined at test time, according to the rule in \cite{mansour2009domain}. 
A similar multi-way adversarial strategy is used also in \cite{MDAN_ICLRW18}, but this work comes 
with a theoretical support that frees it from the need of respecting a specific optimal source 
combination and thus from the need of learning the source weights.

While recent deep DA methods significantly outperform approaches based on hand-crafted features, the
vast majority of them only consider single-source, single-target settings.
Moreover, almost all work presented in the literature so far assume to have direct access to multiple source domains, where in many practical applications such knowledge might not be directly available, or costly to obtain in terms of time and human annotators. 
To our knowledge, this is the first work proposing a deep architecture for discovering latent source domains and exploiting them for improving classification performance on target data.

{Finally, our work is also related to domain generalization \cite{muandet2013domain,li2017deeper}. As in domain generalization, we assume the presence of multiple source domains. However, in domain generalization the domains are clearly separated and are leveraged to produce a model for any unseen target domain, \eg learning domain invariant representations \cite{li2017deeper,ghifary2016deep,sankaranarayanan2017generate,li2018learning}, or combining domain specific information \cite{xu2014exploiting,li2017domain,mancini2018robust,mancini2018best}. In our scenario instead, the multiple source domains are mixed and our goal is to latently discover them while producing a model for the single or multiples (possibly mixed) target domains. We highlight that discovering latent domains can be useful even in the context of domain generalization, as shown in \cite{xu2014exploiting}.}




\section{Method}
\label{sec:method}
\subsection{Problem Formulation and Notation}

We assume to have data belonging to one of several domains. Specifically, we consider  $\con{k_s}$ \emph{source} domains, characterized by unknown probability distributions $p_{\mathtt{xy}}^{s_1},\dots,p_{\mathtt{xy}}^{s_\con{k_s}}$ defined over $\set{X}\times\set{Y}$, where $\set X$ is the input space (\eg images) and $\set Y$ the output space (\eg object or scene categories) and, similarly, we assume $\con{k_t}$ \emph{target} domains characterized by $p_{\mathtt{xy}}^{t_1},\ldots,p_{\mathtt{xy}}^{t_\con{k_t}}$.
The numbers of source and target domains are not necessarily known a-priori, and are left as hyper-parameters of our method.

During training we are given a set of labeled sample points from the source domains, and a set of unlabeled sample points from the target domains, while we can have partial or no information about the domain of the source sample points.
We model the source data as a set $\set{S}=\{(x_1^s,y_1^s),\dots,(x_\con{n}^s,y_\con{n}^s)\}$ of i.i.d. observations from a mixture distribution $p_{\mathtt{xy}}^s=\sum_{i=1}^\con{k_s} \pi_{s_i} p_{\mathtt{xy}}^{s_i}$, where $\pi_{s_i}$ is the unknown probability of sampling from a source domain $s_i$.
Similarly, the target sample $\set{T}=\{x_1^t,\dots,x_\con{m}^t\}$ consists of i.i.d. observations from the marginal $p_\mathtt{x}^t$ of the mixture distribution over target domains.
Furthermore, we denote by $x_\set{S}=\{x_1^s,\dots,x_\con{n}^s\}$ and $y_\set{S}=\{y_1^s,\dots,y_\con{n}^s\}$, the source data and label sets, respectively.
We assume to know the domain label for a (possibly empty) sub-sample $\hat {\set S}\subset \set S$ from the source domains and we denote by $d_{\hat {\set S}}$ the domain labels in $\{s_1.\ldots,s_\con k\}$ of the sample points in $x_{\hat S}$. The set of source domains labels is given by $\set D_s=\{s_1,\ldots,s_\con {k_s}\}$, and similarly the set of target domains is denoted by $\set D_t$. 

Our goal is to learn a predictor that is able to classify data from the target domains. The major difficulties that this problem poses, and that we have to deal with, are: (i) the distributions of source and target domains can be drastically different, making it hard to apply a classifier learned on one domain to the others, (ii) we lack direct observation of target labels, and (iii) the assignment of each source and target sample point to its domain is unknown, or known for a very limited number of source sample points.

Several previous works~\cite{long2015learning,tzeng2015simultaneous,ganin2014unsupervised,ghifary2016deep,bousmalis2016domain,carlucci2017autodial} have tackled the related problem of domain adaptation in the context of deep neural networks, dealing with (i) and (ii) in the single domain case for both source and target data (\ie $\con {k_s}=1$ and $\con {k_t}=1$).
In particular, some recent works have demonstrated a simple yet effective approach based on the replacement of standard BN layers with specific \textit{Domain Alignment layers} ~\cite{carlucci2017just,carlucci2017autodial}.
These layers reduce internal domain shift at different levels within the network by normalizing features in a domain-dependent way, matching their distributions to a pre-determined one.
We revisit this idea in the context of multiple, unknown source and target domains and introduce a novel Multi-domain DA layer (mDA-layer) in Section~\ref{sec:dalayers}, which is able to normalize the multi-modal feature distributions encountered in our setting.
To do this, our mDA-layers exploit a side-output branch attached to the main network (see Section~\ref{sec:domain-prediction}), which predicts domain assignment probabilities for each input sample.
Finally, in Section~\ref{sec:loss} we show how the predicted domain probabilities can be exploited, together with the unlabeled target samples, to construct a prior distribution over the network's parameters which is then used to define the training objective for our network.

\subsection{Multi-domain DA-layers}
\label{sec:dalayers}

\begin{figure*}[t]
  \centering
  \includegraphics[width=0.99\textwidth,trim={0 5.8cm 0 0},clip]{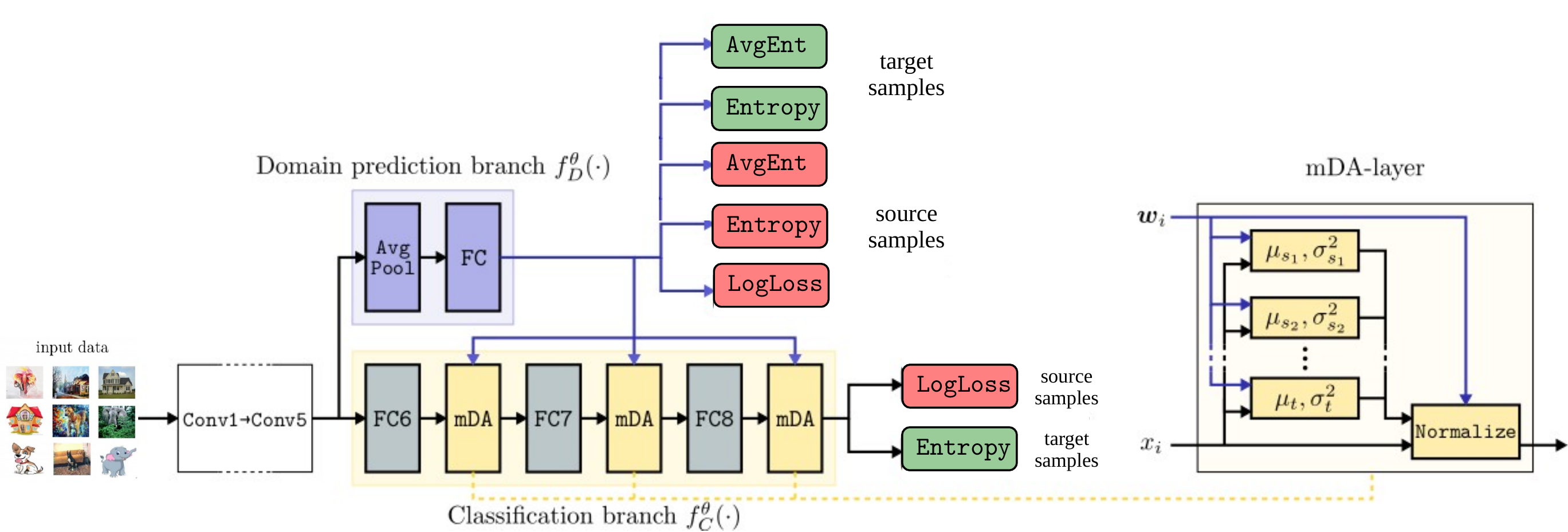}
  \caption{Schematic representation of our method applied to the AlexNet architecture (left) and of an mDA-layer (right).}
  \label{fig:method}
\end{figure*}

DA-layers~\cite{li2016revisiting,carlucci2017just,carlucci2017autodial} are motivated by the observation that, in general, activations within a neural network follow domain-dependent distributions.
As a way to reduce domain shift, the activations are thus normalized in a domain-specific way, shifting them according to a parameterized transformation in order to match their first and second order moments to those of a reference distribution, which is generally chosen to be normal with zero mean and unit standard deviation.
While previous works only considered settings with two domains, \ie source and target, the basic idea can in fact be applied to any number of domains, as long as the domain membership of each sample point is known.
Specifically, denoting as $q^d_\mathtt{x}$ the distribution of activations for a given feature channel and domain $d$, an input $x^d\sim q^d_\mathtt{x}$ to the DA-layer can be normalized according to
\[
  \DA(x^d; \mu_d, \sigma_d) = \frac{x^d - \mu_d}{\sqrt{\sigma_d^2 + \epsilon}},
\]
where $\mu_d = \E_{x\sim q^d_\mathtt{x}}[x]$, $\sigma^2_d = \Var_{x\sim q^d_\mathtt{x}}[x]$ are mean and variance of the input distribution, respectively, and $\epsilon>0$ is a small constant to avoid numerical issues.
In practice, when the statistics $\mu_d$ and $\sigma^2_d$ are computed over the current mini-batch, we obtain the application of standard batch normalization separately to the sample points of each domain.

As mentioned above, this approach requires full domain knowledge, because for each domain $d$, $\mu_d$ and $\sigma^2_d$ need to be calculated on a data sample belonging to the specific domain $d$.
In our case, however, we do not know the domain of the source/target sample points, or we have only partial knowledge about that.
To tackle this issue, we propose to model the layer's input distribution as a mixture of Gaussians, with one component per domain.
Specifically, we define a global input distribution $q_\mathtt{x} = \sum_d \pi_d q^d_\mathtt{x}$, where $\pi_d$ is the probability of sampling from domain $d$, and $q^d_\mathtt{x} = \mathcal{N}(\mu_d, \sigma^2_d)$ is the domain-specific distribution for $d$, namely a normal distribution with mean $\mu_d$ and variance $\sigma_d^2$.
Given a mini-batch $\set{B}=\{x_i\}_{i=1}^\con{b}$, a maximum likelihood estimate of the parameters $\mu_d$ and $\sigma_d^2$ is given by
\begin{equation}
\label{eqn:mixture-params}
\begin{aligned}
  \mu_d &= \sum_{i=1}^\con{b} \alpha_{i,d} x_i, &
  \sigma_d^2 &= \sum_{i=1}^\con{b} \alpha_{i,d} (x_i - \mu_d)^2,
\end{aligned}
\end{equation}
where
\begin{equation}
\label{eqn:weights}
  \alpha_{i,d} = \frac{q_\mathtt{d|x}(d\mid x_i)}{\sum_{j=1}^\con{b} q_\mathtt{d|x}(d\mid x_j)},
\end{equation}
and $q_\mathtt{d|x}(d\mid x_i)$ is the conditional probability of $x_i$ belonging to domain $d$, given $x_i$.
Clearly, the value of $q_\mathtt{d|x}$ is known for all sample points for which we have domain information.
In all other cases, the missing domain assignment probabilities are inferred from data, using the \emph{domain prediction} network branch which will be detailed in Section~\ref{sec:domain-prediction}.
Thus, from the perspective of the alignment layer, these probabilities become an additional input, which we denote as $w_{i,d}$ for the predicted probability of $x_i$ belonging to $d$.

By substituting $w_{i,d}$ for $q_\mathtt{d|x}(d\mid x_i)$ in 
\eqref{eqn:weights}, we obtain a new set of empirical estimates for the mixture parameters, which we denote as $\hat{\mu}_d$ and $\hat{\sigma}^2_d$.
These parameters are used to normalize the layer's inputs according to
\begin{equation}
\label{eqn:normalization}
  \mDA(x_i, \vct{w}_i; \vct{\hat{\mu}}, \vct{\hat{\sigma}}) = \sum_{d\in \set D} w_{i,d} \frac{x_i - \hat{\mu}_d}{\sqrt{\hat{\sigma}_d^2 + \epsilon}},
\end{equation}
where $\vct{w}_i=\{w_{i,d}\}_{d\in\set D}$, $\vct{\hat{\mu}}=\{\hat{\mu}_d\}_{d\in\set D}$, $\vct{\hat{\sigma}}=\{\hat{\sigma}^2_d\}_{d\in\set D}$ and $\set D$ is the set of source/target latent domains.
As in previous works \cite{carlucci2017autodial,carlucci2017just,ioffe2015batch}, during back-propagation we calculate the derivatives through the statistics and weights, propagating the gradients to both the main input and the domain assignment probabilities.

\subsection{Domain prediction}
\label{sec:domain-prediction}
Our mDA-layers receive a set of domain assignment probabilities for each input sample point, which needs to be predicted, and different mDA-layers in the network, despite having different input distributions, share consistently the same domain assignment for the sample points. 
As a practical example, in the typical case in which mDA-layers are used in a CNN to normalize convolutional activations, the network would predict a single set of domain assignment probabilities for each input image, which would then be fed to all mDA-layers and broadcasted across all spatial locations and feature channels corresponding to that image.
We compute domain assignment probabilities using a distinct section of the network, which we call the \emph{domain prediction} branch, while we refer to the main section of the network as the \emph{classification} branch.
The two branches share the bottom-most layers and parameters as depicted in Figure~\ref{fig:method}.

The domain prediction branch is implemented as a minimal set of layers followed by two softmax operations with $\con{k_s}$ and $\con{k_t}$ outputs for the source and target latent domains, respectively (more details follow in Section~\ref{sec:experiments}). The rationale of keeping the domain prediction separated between source and target derives from the knowledge that we have about the source/target membership of a sample point that we receive in input, while it remains unknown the specific source or target domain it belongs to.
Furthermore, for each sample point $x_i$ with known domain membership $\hat d$, we fix in each mDA-layer $w_{i,d}=1$ if $d=\hat d$, otherwise $w_{i,d}=0$ .

We split the network into a domain prediction branch and classification branch at some low level layer.
This choice is motivated by the observation~\cite{aljundi2016lightweight} that features tend to become increasingly more domain invariant going deeper into the network, meaning that it becomes increasingly harder to compute a domain membership as a function of deeper features.
In fact, as pointed out in~\cite{carlucci2017autodial}, this phenomenon is even more evident in networks that include DA-layers.

\subsection{Training the network}
\label{sec:loss}
{In order to exploit unlabeled data within our discriminative setting, we follow the approach sketched in~\cite{carlucci2017autodial}, where unlabeled data is used to define a regularizer over the network's parameters.
By doing so, we obtain a loss for $\theta$ that takes the following form:
\begin{equation}
\label{eqn:loss_general}
\begin{aligned}
  L(\theta) = L_\text{cls}(\theta)+L_\text{dom}(\theta)\,,
\end{aligned}
\end{equation}
where $L_\text{cls}$ is a loss term that penalizes based on the final classification task, while $L_\text{dom}$ accounts for the domain classification task.}

{\myparagraph{Classification loss $L_\text{cls}$.}
The classification loss consists of two components, accounting for the the supervised sample from the source domain $\set S$ and the unlabeled target sample $\set T$, respectively:
\begin{equation}
\label{eqn:loss_cls}
\begin{aligned}
  L_\text{cls}(\theta)=&- \frac{1}{\con{n}} \sum_{i=1}^\con{n} \log f_C^\theta(y_i^s; x_i^s)+\frac{\lambda_C}{\con{m}} \sum_{i=1}^\con{m} H(f_C^\theta(\cdot;x_i^t)).
\end{aligned}
\end{equation}
The first term on the right-hand-side is the average log-loss related to the supervised examples in $\set S$, where $f_C^\theta(y_i^s; x_i^s)$ denotes the output of the \emph{classification branch} of the network for a source sample, \ie the predicted probability of $x_i^s$ having class $y_i^s$. The second term on the right-hand-side of \eqref{eqn:loss_cls} is the entropy $H$ of the classification distribution $f_C^\theta(\cdot; x_i^t)$, averaged over all unlabeled target examples $x_i^t$ in $\set T$, scaled by a positive hyper-parameter $\lambda_C$.}

{\myparagraph{Domain loss $L_\text{dom}$.}
Akin to the classification loss, the domain loss presents a component exploiting the supervision deriving from the known domain labels in $\hat{\set S}$ and a component exploiting the domain classification distribution on all sample points lacking supervision. However, the domain loss has in addition a term that tries to balance the distribution of sample points across domains, in order to avoid predictions to collapse into trivial solutions such as constant assignments to a single domain. 
Accordingly, the loss takes the following form:
\begin{multline}
\label{eqn:loss_dom}
  L_\text{dom}(\theta)=-\frac{\lambda_D}{|\hat{\set{S}}|} \sum_{x_i \in x_{\hat {\set{S}}}} \log f_{D_s}^\theta(d_i; x_i)\\
  -\lambda_B H(\bar f_{D_s}^\theta(\cdot)) +\frac{\lambda_E}{|\set S\setminus\hat{\set S}|} \sum_{x\in x_{\set{S}\setminus\hat{\set S}}} H(f_{D_s}^\theta(\cdot; x))\\
  -\lambda_B H(\bar f_{D_t}^\theta(\cdot)) +\frac{\lambda_E}{\con m} \sum_{i=1}^\con m H(f_{D_t}^\theta(\cdot; x_i^t)).
\end{multline}
Here, $f_{D_s}^\theta$ and $f_{D_t}^\theta$ denote the outputs of the \emph{domain prediction branch} for data points from the source and target domains, respectively, while
$\bar f_{D_s}^\theta$  and a$\bar f_{D_t}^\theta$ denote the distributions of predicted domain classes across $\set S$ and $\set T$, respectively, \ie
\[
\bar f_{D_s}^\theta(y) = \frac{1}{\con n}\sum_{i=1}^\con nf_{D_s}^\theta(y; x^s_i),\quad
\bar f_{D_t}^\theta(y) = \frac{1}{\con m}\sum_{i=1}^\con mf_{D_t}^\theta(y; x^t_i)\,.
\]
The first term in \eqref{eqn:loss_dom} enforces the correct domain prediction on the sample points with known domain and it is scaled by a positive hyper-parameter $\lambda_D$.
The terms scaled by the positive hyper-parameter $\lambda_E$ enforce domain predictions with low uncertainty for the data points with unknown domain labels, by minimizing the entropy of the output distribution. 
Finally, the terms scaled by the positive hyper-parameter $\lambda_B$ enforce balanced distributions of predicted domain classes across the source and target sample, by maximizing the entropy of the averaged distribution of domain predictions. 
Interestingly, since the classification branch has a dependence on the domain prediction branch via the mDA-layers, by optimizing the proposed loss, the network learns to predict domain assignment probabilities that result in a low classification loss.
In other words, the network is free to predict domain memberships that do not necessarily reflect the real ones, as long as this helps improving its classification performance.}

{We optimize the loss in \eqref{eqn:loss_general} with stochastic gradient descent. Hence, the samples $\set S$, $\set T$, $\hat{\set S}$ that are considered in the computation of the gradients are restricted to a random subsets contained in the mini-batch. In Section~\ref{sec:experiments-setup} we provide more details on how each mini-batch is sampled.}

\section{Experiments}
\label{sec:experiments}
\subsection{Experimental Setup}
\label{sec:experiments-setup}
\subsubsection{Datasets}
\label{sec:experiments-datasets}
In our evaluation we consider several common DA benchmarks: the combination of USPS~\cite{friedman2001elements}, MNIST~\cite{lecun1998gradient} and MNIST-m~\cite{ganin2014unsupervised}; the Digits-five benchmark in~\cite{xu2018deep}; Office-31~\cite{saenko2010adapting}; Office-Caltech~\cite{gong2012geodesic} and PACS~\cite{li2017deeper}.

\myparagraph{MNIST, MNIST-m and USPS}
are three standard datasets for digits recognition.
USPS \cite{friedman2001elements} is a dataset of digits scanned from U.S. envelopes, MNIST~\cite{lecun1998gradient} is a popular benchmark for digits recognition and MNIST-m~\cite{ganin2014unsupervised} its counterpart obtained by blending the original images with colored patches extracted from BSD500 photos~\cite{arbelaez2011contour}. 
Due to their different representations (\eg colored vs gray-scale), these datasets have been adopted as a DA benchmark by many previous works \cite{ganin2014unsupervised,bousmalis2016domain,bousmalis2016unsupervised}.
Here, we consider a multi source DA setting, using MNIST and MNIST-m as sources and USPS as target, training on the union of the training sets and testing on the test set of USPS.

\myparagraph{Digits-five} is an experimental setting proposed in \cite{xu2018deep} which considers 5 datasets of digits recognition.
In addition to MNIST, MNST-m and USPS, it includes SVHN~\cite{netzer2011reading} and Synthetic numbers datasets~\cite{ganin2016domain}.
SVHN~\cite{netzer2011reading} contains pictures of real-world house numbers, collected from Google Street View.
Synthetic numbers~\cite{ganin2016domain} is built from computer generated digits, including multiple sources of variations (\ie position, orientation, background, color and amount of blur), for a total of 500 thousands images.
We follow the experimental setting described in~\cite{xu2018deep}: the train\slash{}test split comprises a subset of 25000 images for training and 9000 for testing for each of the domains, except for USPS for which the entire dataset is used.
As in~\cite{xu2018deep}, we report the results when either SVHN or MNIST-m are used as targets and all the other domains are taken as sources.

\myparagraph{Office-31} is a standard DA benchmark which contains images of 31 object categories collected from 3 different sources: Webcam (W), DSLR camera (D) and the Amazon website (A).
Following~\cite{xiong2014latent}, we perform our tests in the multi-source setting, where each domain is in turn considered as target, while the others are used as source.

\myparagraph{Office-Caltech}~\cite{gong2012geodesic} is obtained by selecting the subset of $10$ common categories in the Office31 and the Caltech256~\cite{griffin2007caltech} datasets.
It contains $2533$ images, about half of which belong to Caltech256.
The different domains are Amazon (A), DSLR (D), Webcam (W) and Caltech256 (C).
In our experiments we consider the set of source\slash{}target combinations used in~\cite{gong2013reshaping}.

\myparagraph{PACS}~\cite{li2017deeper} is a recently proposed DA benchmark which is especially interesting due to the significant domain shift between its domains.
It contains images of 7 categories (\textit{dog, elephant, giraffe, guitar, horse}) and 4 different visual styles: \ie Photo (P), Art paintings (A), Cartoon (C) and Sketch (S).
We employ the dataset in two different settings.
First, following the experimental protocol in~\cite{li2017deeper}, we train our model considering 3 domains as sources and the remaining as target, using all the images of each domain.
Differently from \cite{li2017deeper} we consider a DA setting (\ie target data is available at training time) and we do not address the problem of domain generalization.
Second, we use 2 domains as sources and the remaining 2 as targets, in a multi-source multi-target scenario.
In this setting the results are reported as average accuracy between the 2 target domains.

\addnote[domain-labels]{1}{In all experiments and settings, we assume to have no domain labels (\ie $\hat{\set{S}}=\emptyset$), unless otherwise stated.}

\subsubsection{Networks and training protocols}
\label{sec:experiments-networks}
We apply our approach to four different CNN architectures: the MNIST and SVHN networks described in~\cite{ganin2014unsupervised,ganin2016domain}, AlexNet~\cite{krizhevsky2012imagenet} and ResNet~\cite{he2016deep}.
We choose AlexNet due to its widespread use in many relevant DA works~\cite{ganin2014unsupervised,carlucci2017autodial,long2015learning,long2016unsupervised}, while ResNet is taken as an exemplar for modern state-of-the-art architectures employing batch-normalization layers.
Both AlexNet and ResNet are first pre-trained on ImageNet and then fine-tuned on the datasets of interest.
The MNIST and SVHN architectures are chosen for fair comparison with previous works considering digits datasets~\cite{ganin2016domain,xu2018deep}. 
Unless otherwise noted, we optimize our networks using Stochastic Gradient Descent with momentum $0.9$ and weight decay $5\times10^{-4}$.

For the evaluation on MNIST, MNIST-m and USPS datasets, we employ the MNIST network described in~\cite{ganin2014unsupervised}, adding an mDA-layer after each convolutional and fully-connected layer.
The domain prediction branch is attached to the output of \texttt{conv1}, and is composed of a convolution with the same meta-parameters as \texttt{conv2}, a global average pooling, a fully-connected layer with 100 output channels and finally a fully-connected classifier.
Following the protocol described in~\cite{carlucci2017autodial,ganin2014unsupervised}, we set the initial learning rate $l_0$ to 0.01 and we anneal it through a schedule $l_p$ defined by $l_p= \frac{l_0}{(1+\gamma p)^\beta}$ where $\beta=0.75$, $\gamma=10$ and $p$ is the training progress increasing linearly from 0 to 1.
We rescale the input images to $32\times 32$ pixels, subtract the per-pixel image mean of the dataset and feed the networks with random crops of size $28\times 28$.
A batch-size of 128 images per domain is used.

For the Digits-five experiments we employ the SVHN architecture of~\cite{ganin2016domain}, which is the same architecture adopted by~\cite{xu2018deep}, augmented with mDA-layers and a domain prediction branch in the same way as the MNIST network described in the previous paragraph.
We train the architecture for 44000 iterations, with a batch-size of 32 images per domain, an initial learning rate of $10^{-4}$ which is decayed by a factor of 10 after 80\% of the training process.
We use Adam as optimizer with a weight decay $5\times10^{-5}$, and pre-process the input images like in the MNIST, MNIST-m, USPS experiments.

For the experiments on Office-31 and Office-Caltech we employ the AlexNet architecture.
We follow a setup similar to the one proposed in~\cite{carlucci2017autodial,carlucci2017just}, fixing the parameters of all convolutional layers and inserting mDA-layers after each fully-connected layer and before their corresponding activation functions.
The domain prediction branch is attached to the last pooling layer \texttt{pool5}, and is composed of a global average pooling, followed by a fully connected classifier to produce the final domain probabilities.
The training schedule and hyper--parameters are set following~\cite{carlucci2017autodial}. 

For the experiments on the PACS dataset we consider the ResNet architecture in the 18-layers setup described in~\cite{he2016deep}, denoted as ResNet18.
This architecture comprises an initial $7\times 7$ convolution, denoted as \texttt{conv1}, followed by 4 main modules, denoted as \texttt{conv2} -- \texttt{conv5}, each containing two residual blocks.
To apply our approach, we replace each Batch Normalization layer in the residual blocks of the network with an mDA-layer.
The domain prediction branch is attached to \texttt{conv1}, after the pooling operation.
The branch is composed of a residual block with the same structure as \texttt{conv2}, followed by global average pooling and a fully connected classifier.
In the multi-target experiments we add a second, identical domain prediction branch to discriminate between target domains.
We also add a standard BN layer after the final domain classifiers, which we found leads to a more stable training process in the multi-target case.
In both cases, we adopt the same training meta-parameters as for AlexNet, with the exception of weight-decay which is set to $10^{-6}$ and learning rate which is set to $5\cdot10^{-4}$.
The network is trained for 600 iterations with a batch-size of 48, equally divided between the domains, and the learning rate is scaled by a factor 0.1 after 75\% of the iterations.

Regarding the hyper--parameters of our method,  we set the number of source domains $\con{k}$ equal to $Q-1$, where $Q$ is the number of different datasets used in each single experiment.
In the multi-source multi-target scenarios, since we always have the domains equally split between source and target, we consider $\con{k}$ equal $Q/2$ for both source and target.
Following \cite{carlucci2017autodial}, in the experiments with AlexNet we fix $\lambda_C=\lambda_E=0.2$ with $\lambda_B=0.1$.
Similarly, for the experiments on digits classification, we set $\lambda_C=\lambda_E=0.1$ and $\lambda_B=0.05$ for MNIST, MNIST-m and USPS, and $\lambda_C=0.01$ and $\lambda_E=\lambda_B=0.05$ for Digits-five, with $\lambda_E=0.01$ if $\lambda_B=0$, which we found leading to a more stable minimization of the loss of the domain branch.
In the experiments involving ResNet18 we select the values $\lambda_C=0.1$ and $\lambda_E=\lambda_B=0.0001$ through cross-validation, following the procedure adopted in \cite{long2013transfer,carlucci2017autodial}.
Similarly, in the multi-target ResNet18 experiments we select $\lambda_C=\lambda_E=\lambda_B=0.1$. 
When domain labels are available for a subset of source samples, we fix $\lambda_D=0.5$. 

We implement\footnote{Code available at: \url{https://github.com/mancinimassimiliano/latent_domains_DA.git}} all the models with the Caffe~\cite{jia2014caffe} framework and our evaluation is performed using an NVIDIA GeForce 1070 GTX GPU. 
We initialize both AlexNet and ResNet18 from models pre-trained on ImageNet, taking AlexNet from the Caffe model zoo, and converting ResNet18 from the original Torch model\footnote{\scriptsize\url{ https://github.com/HolmesShuan/ResNet-18-Caffemodel-on-ImageNet}}. 
{For all the networks and experiments, we add mDA layers and their variants in place of standard BN layers}.

\subsection{Results}
In this section, we first analyze the proposed approach, demonstrating the advantages of considering multiple sources\slash{}targets and discovering latent domains.
We then compare the proposed method with state-of-the-art approaches.
For all the experiments we report the results in terms of accuracy, repeating the experiments at least 5 times and averaging the results.
In the multi-target experiments, the reported accuracy is the average of the accuracies over the target domains. {As for standard deviations, since we do not tune the hyperparameters of our model and baselines by employing the accuracy on the target domain, their values can be high in some settings. For this reason, in order to provide a more appropriate analysis of the significance of our results, we propose to adopt the following approach. In particular, let us model the accuracy of an algorithm as a random variable $X_a$ with unknown distribution. The accuracy of a single run of the algorithm is an observation from this distribution. Therefore, in order to compare two algorithms we consider the two sets of associated observations $A=\{a_1,\dots,a_n\}$ and $B=\{b_1,\dots,b_m\}$ and estimate the probability that one algorithm is better than the other as: 
\[
p(X_a>X_b)=\frac{\sum_{a \in A}\sum_{b \in B} \delta(a>b)}{|A|\cdot|B|}
\] 
where $\delta$ is the Dirac function. 
In the following we use this metric to compare our approach with respect to a baseline where no latent domain discovery process is implemented (specifically, the method DIAL~\cite{carlucci2017just}, see below) considering five runs for each experiment. For sake of clarity, we denote this probability estimate as $p^*$. 
}

{In the following we first analyze the performances of the proposed approach with $\lambda_B=0$ (denoted as Ours $\lambda_B=0$), \ie the algorithm we presented in \cite{mancini2018boosting}, and then we describe the impact of the loss term we introduce in this paper setting $\lambda_B>0$ (denoted simply as Ours).}

\subsubsection{Experiments on the Digits datasets}
In a first series of experiments, reported in Table~\ref{tab:digits}, we test the performance of our approach on the MNIST, MNIST-m to USPS benchmark (M-Mm to U).
The comparison includes: (i) the baseline network trained on the union of all source domains (\textit{Unified sources}); (ii) training separate networks for each source, and selecting the one the performs the best on the target (\textit{Best single source}); (iii) DIAL~\cite{carlucci2017just}, trained on the union of the sources (\textit{DIAL \cite{carlucci2017just} - Unified sources}); (iv) DIAL, trained separately on each source and selecting the best performing model on the target (\textit{DIAL \cite{carlucci2017just} - Best single source}).
We also report the results of our approach in the ideal case where the multiple source domains are known and we do not need to discover them (\textit{Multi-source DA}).
For our approach with $\lambda_B=0$, we consider several different values of $\con{k}$, \ie the number of discovered source domains.

By looking at the table several observations can be made.
First, there is a large performance gap between models trained only on source data and DA methods, confirming that deep architectures by themselves are not enough to solve the domain shift problem~\cite{donahue2014decaf}.
Second, in analogy with previous works on DA~\cite{mansour2009domain,duan2009domain,sun2011two}, we found that considering multiple sources is beneficial for reducing the domain shift with respect to learning a model on the unified source set.
Finally, and more importantly, when the domain labels are not available, our approach is successful in discovering latent domains and in exploiting this information for improving accuracy on target data, partially filling the performance gap between the single source models and \textit{Multi-source DA}. 
Interestingly, the performance of our algorithm changes only slightly for different values of $\con{k}$, motivating our choice to always fix $\con{k}$ to the known number of domains in the next experiments. {Importantly, comparing our approach with DIAL we achieve higher accuracy in most of the runs, \ie  $p^*=0.65$}. {In this experiment, the introduction of the loss term forcing a uniform assignment among clusters (denoted as Ours) leads to comparable performances to our method with $\lambda_B=0$. This behaviour can be ascribed to the fact that the separation among different domains is quite clear in this case and adding constraints to the domain discovery process is not required. In the following, we show that the proposed loss is beneficial in more challenging datasets.} 

\begin{table}[t]
			\caption{Digits datasets: comparison of different models in the multi-source scenario. MNIST (M) and MNIST-m (Mm) are taken as source domains, USPS (U) as target.} 
		\centering
		\scalebox{.9}{
		\begin{tabular}{ l | c} 
			\hline
			Method & M-Mm to U \\
            	\hline
                Unified sources &  57.1 \\
                Best single source & 59.8  \\
			DIAL \cite{carlucci2017just} - Unified sources &81.7 
            \\
          {DIAL \cite{carlucci2017just} - Best single source} & 81.9 \\
Ours $\lambda_B=0$ $\con{k}=2$& 	82.5 \\
Ours $\lambda_B=0$ $\con{k}=3$& 	82.2 \\
Ours $\lambda_B=0$ $\con{k}=4$& 	82.7 \\
Ours $\lambda_B=0$ $\con{k}=5$& 82.4 \\
Ours ($\con{k}=2$)&82.4  \\\hline\hline
{Multi-source DA}&84.2
\\ \hline 
		\end{tabular}
        }
		\label{tab:digits}
\end{table}

In a second set of experiments (Table~\ref{tab:digits-5}), we compare our approach with previous and recently proposed single and multi-source unsupervised DA approaches.
Following~\cite{xu2018deep}, we perform experiments on the Digits-five dataset, considering two settings with SVHN and MNIST-m as targets.
As in the previous case, we evaluate the performance of the baseline network (with and without BN layers) and of DIAL when trained on the union of the sources, and, as an upper bound, our Multi-source DA with perfect domain knowledge.
Moreover, we consider the Deep Cocktail Network (DCTN)~\cite{xu2018deep} multi-source DA model, as well as the ``source only'' baseline and the single source DA models reported in~\cite{xu2018deep}: Reverse gradient (RevGrad) \cite{ganin2014unsupervised} and Domain Adaptation Networks (DAN) \cite{long2015learning}.
For all single source DA models we consider two settings: ``Unified Sources'', where all source domains are merged, and ``Multi-Source'', where a separate model is trained for each source domain, and the final prediction is computed as an ensemble.
As we can see, the Unified Sources DIAL already achieves remarkable results in this setting, outperforming DCTN, and Multi-source DA only provides a modest performance increase.
As expected, the performance of our approach lies between these two ($p^*$ equal to 0.56 and 0.64 for SVHN and MNIST-m respectively, with $\lambda_B=0$).

\subsubsection{Experiments on PACS}
\myparagraph{Comparison with state of the art.}
In our main PACS experiments we compare the proposed approach with the baseline ResNet18 network, and with ResNet18 + DIAL~\cite{carlucci2017just}, both trained on the union of source sets.
As in the digits experiments, we also report the performance of our method when perfect domain knowledge is available (Multi-source DA).
Table \ref{tab:pacs} shows our results.
In general, DA models are especially beneficial when considering the PACS dataset, and multi-source DA networks significantly outperform the single source one.
Remarkably, our model is able to infer domain information automatically without supervision.
In fact, its accuracy is either comparable with Multi-source DA (Photo, Art and Cartoon) or in between DIAL and Multi-source DA (Sketch). {The average $p^*$ is $0.67$.
Looking at the partial results, it is interesting to note that the improvements of our approach and Multi-source DA w.r.t. DIAL are more significant when either the Sketch or the Cartoon domains are employed as target set (average $p^*=0.81$).
Since these domains are less represented in the ImageNet database, we believe that the corresponding features derived from the pre-trained model are less discriminative, and DA methods based on multiple sources become more effective. Setting $\lambda_B>0$, allows to obtain a further boost of performances in the Sketch scenario, where the source domains are closer in appearances. In the other settings, the domain shift is mostly among the Sketch domain and all the others and it can be easily captured by our original formulation in \cite{mancini2018boosting}.}

\begin{table}[t]
			\caption{Digits-five~\cite{xu2018deep} setting, comparison of different single source and multi-source DA models. The first row indicates the target domain with the others used as sources.} 
		\centering
		\scalebox{.85}{
		\begin{tabular}{l | l | c  c | c } 
			\hline
			&Method  & SVHN & MNIST-m & Mean\\
            	\hline
               \multirow{6}{*}{Unified sources} & Source only (ours)&74.1&64.4&69.3\\ 
                &Source only+BN (ours)&77.7&59.4&68.6\\
                &Source only from~\cite{xu2018deep}&72.2&64.1&68.2\\
            	&RevGrad \cite{ganin2014unsupervised}&68.9&71.6&70.3\\
                &DAN \cite{long2015learning}&71.0&66.6&68.8\\
                &DIAL \cite{carlucci2017just}&82.2&68.8&75.5\\
                \hline
              &Ours $\lambda_B=0$ &{82.4}&{69.1}&{75.8}\\
               &Ours &\textbf{82.6}&\textbf{70.1}&\textbf{76.4}\\
                \hline\hline
                 \multirow{6}{*}{Multi-source} &Only Source \cite{xu2018deep}&64.6&60.7&62.7\\
               & RevGRAD \cite{ganin2014unsupervised}&61.4&71.1&66.3\\
                &DAN \cite{long2015learning}&62.9&62.6&62.8\\
                &DCTN \cite{xu2018deep}&77.5&70.9&74.2\\
                &Multi-source DA&84.1&69.4&76.8\\
                \hline
		\end{tabular}
        }
		\label{tab:digits-5}
\end{table}

\begin{table}[t]
			\caption{PACS dataset: comparison of different methods using the ResNet architecture. The first row indicates the target domain, while all the others are considered as sources. } 
		\centering
		\scalebox{.98}{
		\begin{tabular}{ l | c  c  c  c | c  } 
			\hline
			Method & Sketch & Photo & Art & Cartoon & Mean \\
            	\hline
                ResNet \cite{he2016deep} &60.1&92.9&74.7&72.4&75.0\\
DIAL \cite{carlucci2017just} &66.8&\textbf{97.0} &87.3 &85.5&84.2 \\
Ours $\lambda_B=0$ &{69.6}&\textbf{97.0}&\textbf{87.7}&\textbf{86.9}&{85.3}\\
Ours   &\textbf{70.7} &\textbf{97.0} &87.4 &86.3 &\textbf{85.4} \\\hline\hline
Multi-source DA & 71.6 & 96.6  & 87.5 & 87.0 & 85.7 \\ \hline 
\hline
		\end{tabular}
        }
		\label{tab:pacs}
\end{table}

\begin{figure*}[t]
 \centering
  \subfloat[Photo as target]{\includegraphics[width=0.25\textwidth,height=0.10\textheight]{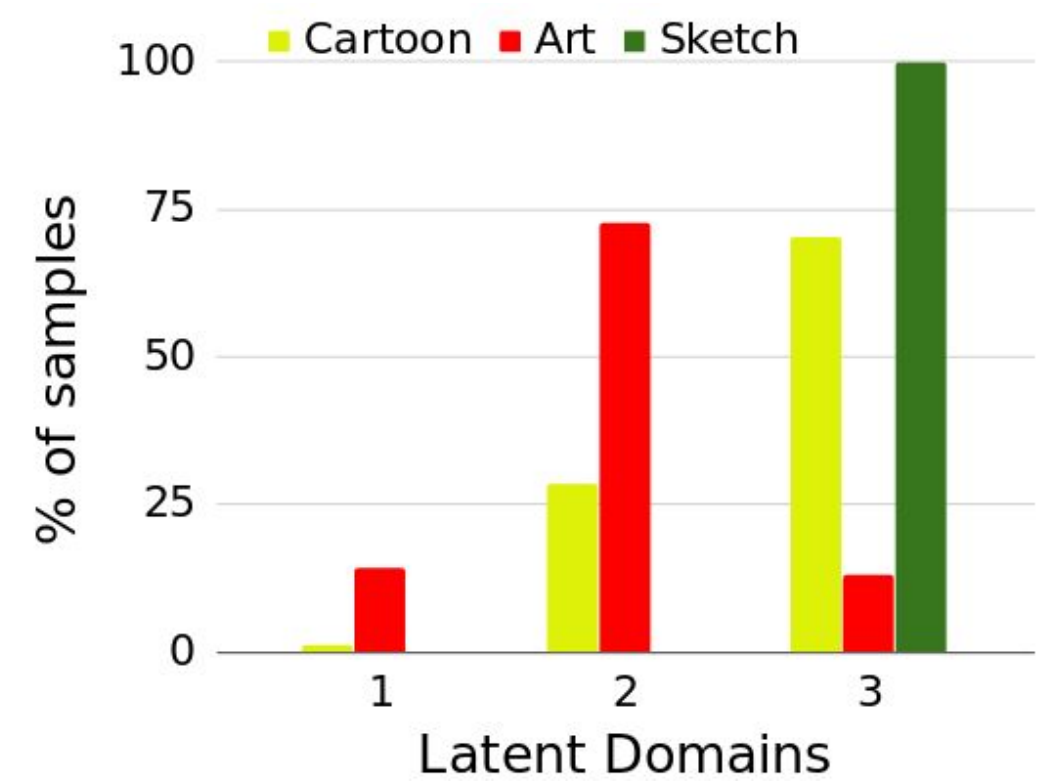}\label{fig:assignment-photo}}
  \subfloat[Art as target]
  {\includegraphics[width=0.25\textwidth,height=0.10\textheight]{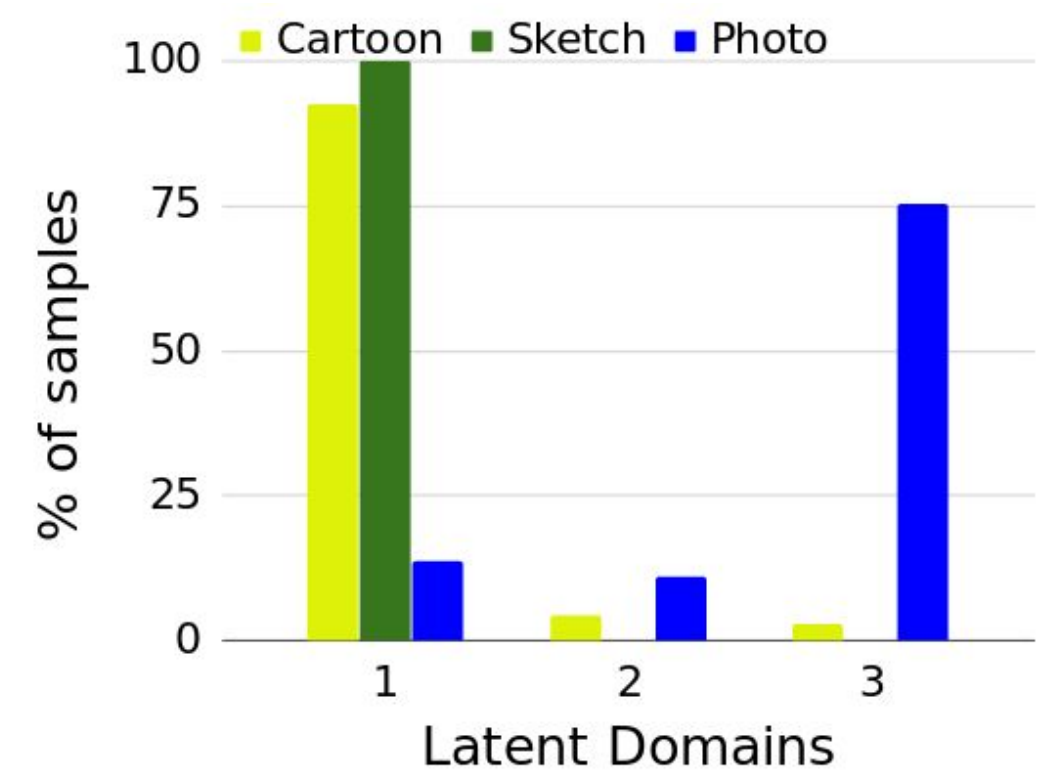}\label{fig:assignment-art}}
  \subfloat[Cartoon as target]{\includegraphics[width=0.25\textwidth,height=0.10\textheight]{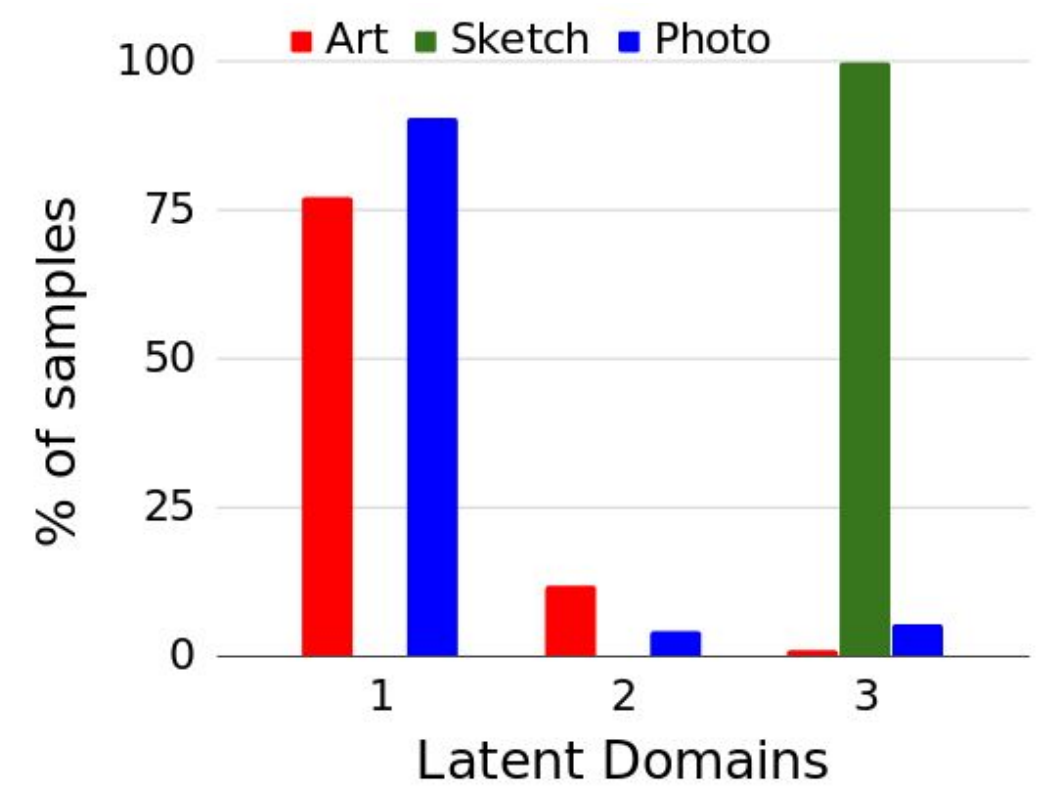}
  \label{fig:assignment-cartoon}} 
  \subfloat[Sketch as target]{\includegraphics[width=0.25\textwidth,height=0.10\textheight]{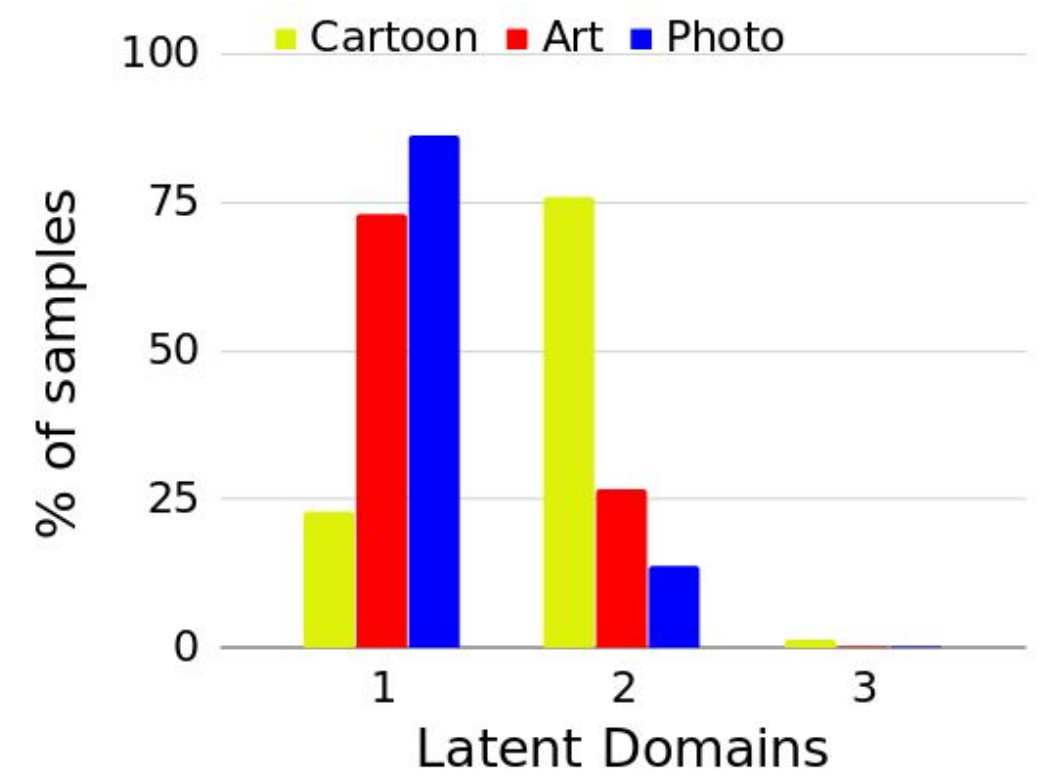}\label{fig:assignment-sketch}}
  \caption{Distribution of the assignments produced by the domain prediction branch for each latent domain in all possible settings of the PACS dataset. Different colors denote different source domains.
  }
  \label{fig:max-assignment}
\end{figure*}

\begin{figure*}[t]
  \centering
  \subfloat[Photo as target]{\includegraphics[width=0.45\textwidth,height=0.2\textwidth,trim={0cm 0cm 36.1cm 0cm},clip]{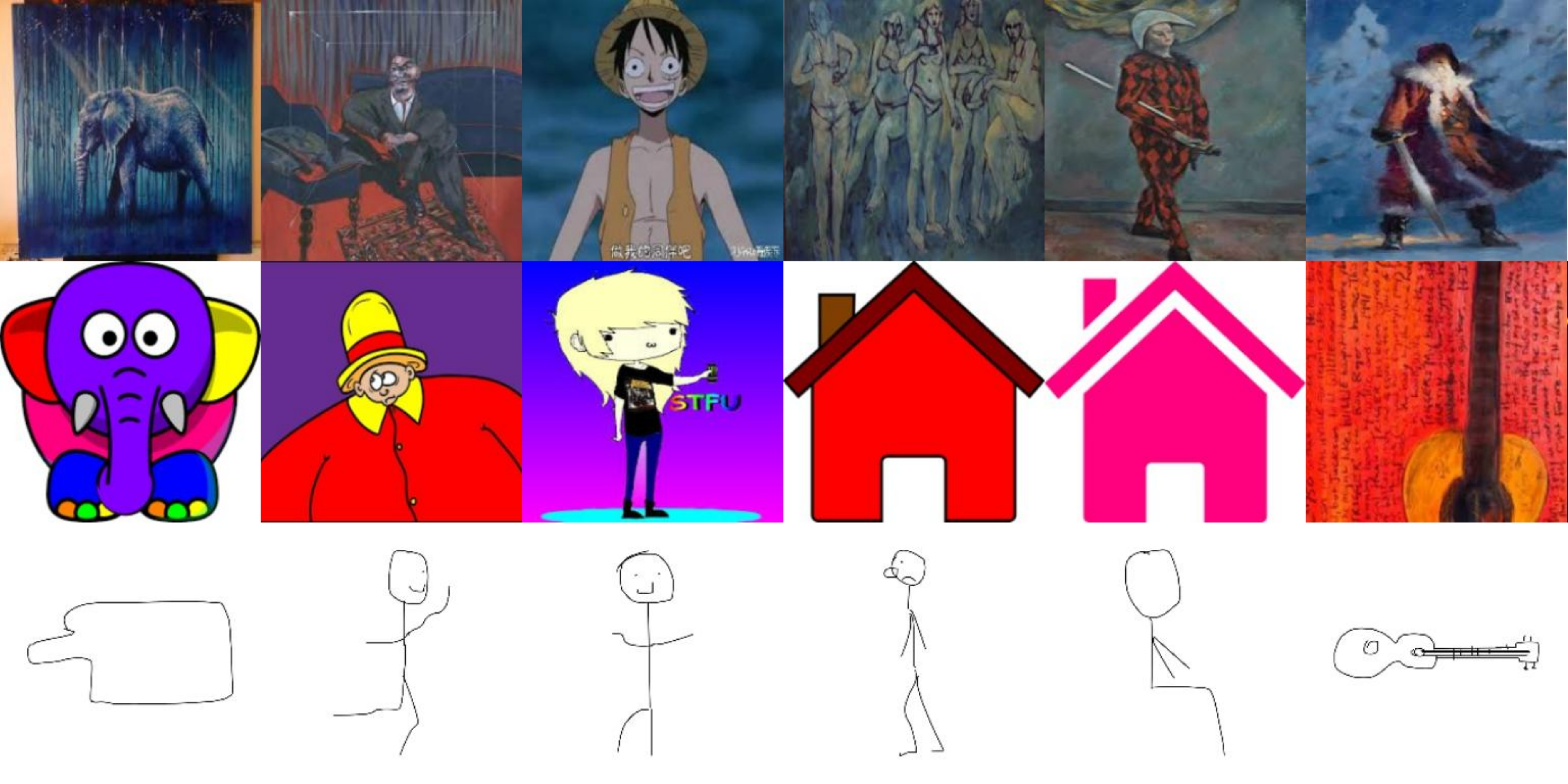}\label{fig:topk-photo}}\hfill
  \subfloat[Art as target]
  {\includegraphics[width=0.45\textwidth,height=0.2\textwidth,trim={0cm 0cm 36.1cm 0cm},clip]{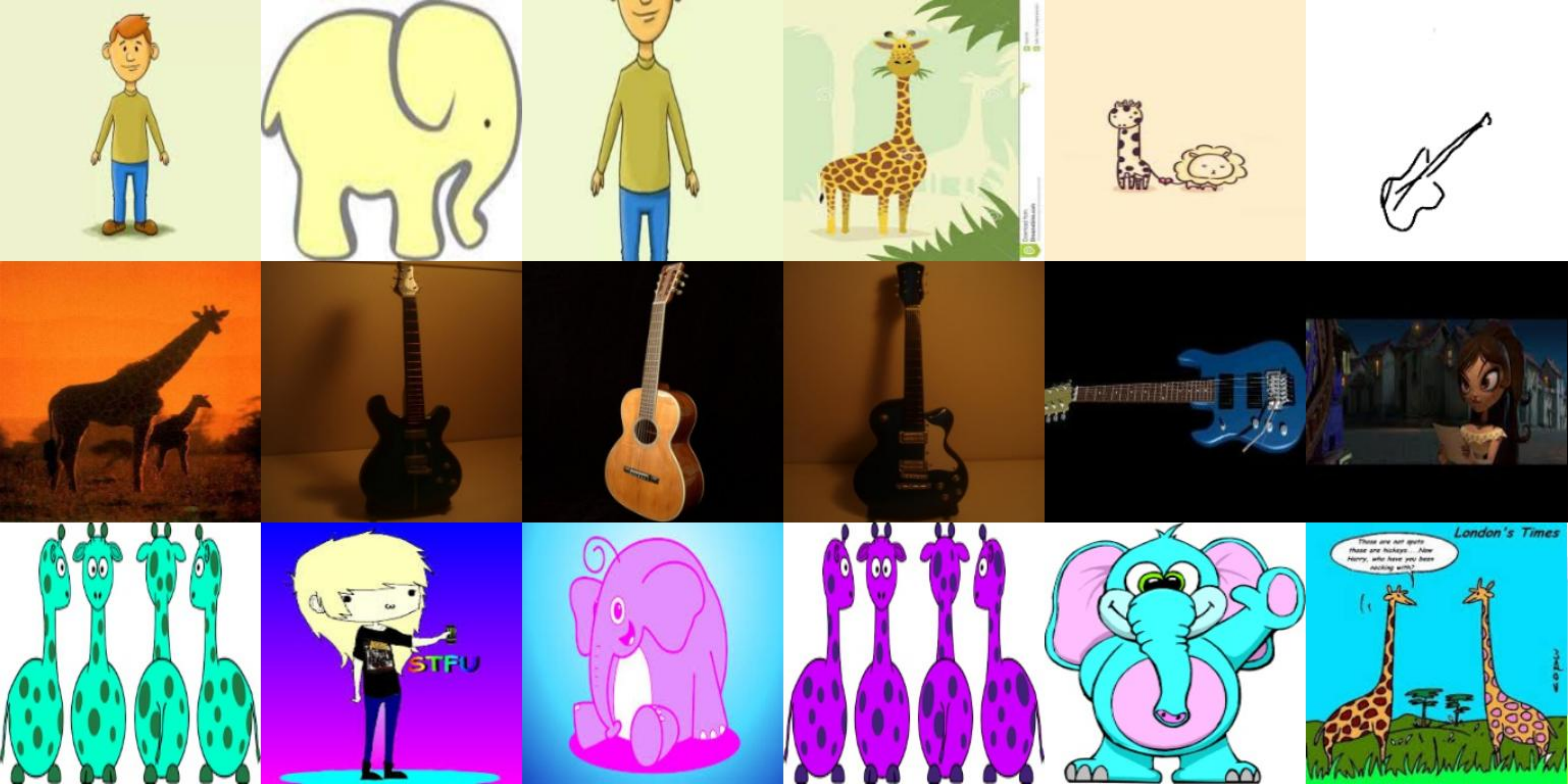}\label{fig:topk-art}}\\  
  \subfloat[Cartoon as target]{\includegraphics[width=0.45\textwidth,height=0.2\textwidth,trim={0cm 0cm 36.1cm 0cm},clip]{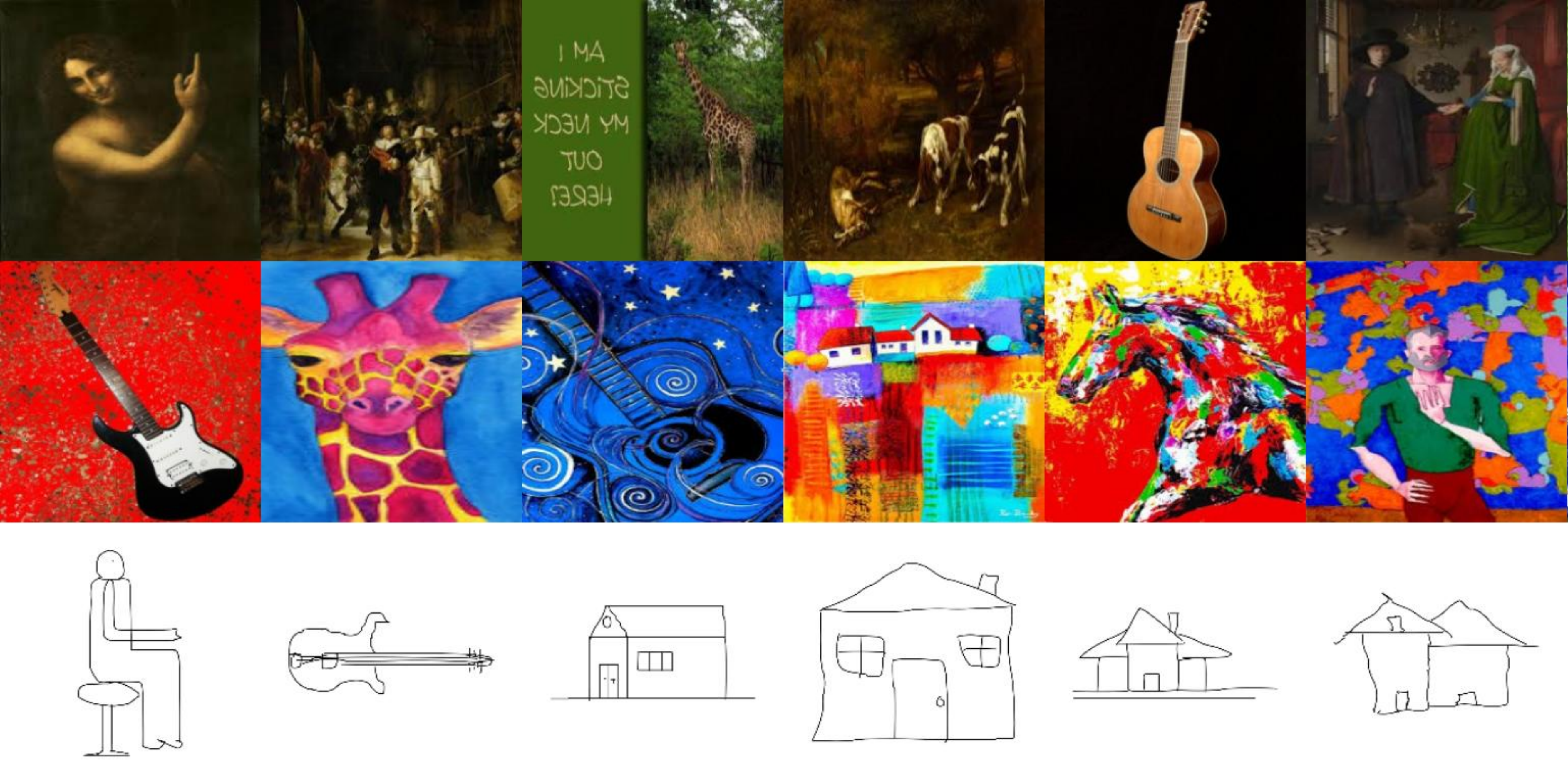}\label{fig:top-k-cartoon}}   \hfill
  \subfloat[Sketch as target]{\includegraphics[width=0.45\textwidth,height=0.2\textwidth,trim={0cm 0cm 36.1cm 0cm},clip]{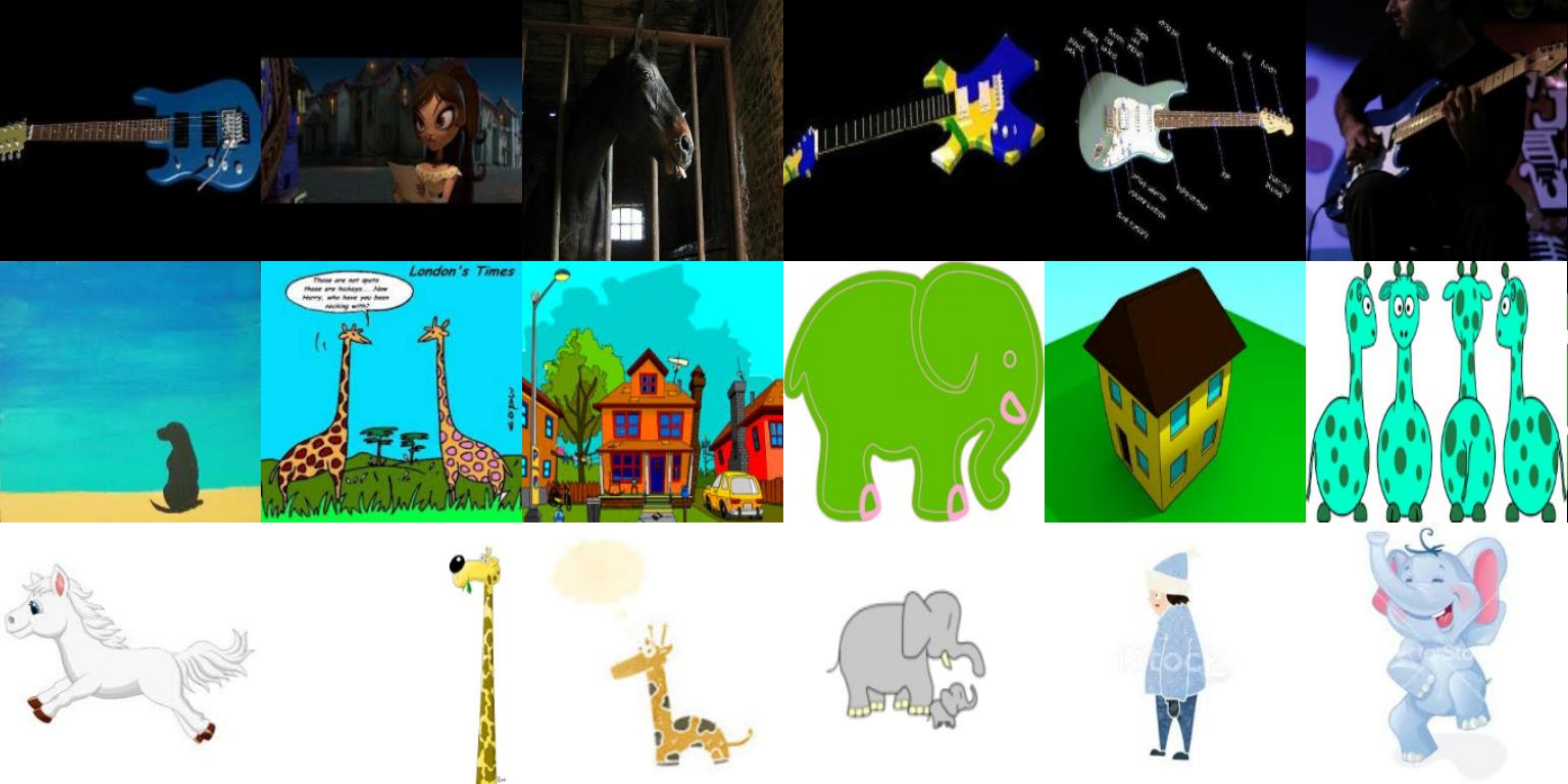}\label{fig:topk-sketch}}
  \caption{Top-6 images associated to each latent domain for the different sources\slash{}target combinations. Each row corresponds to a different latent domain.}
  \label{fig:top-k}
  \vspace{-5pt}
\end{figure*}

\begin{figure*}[t]
 \centering
 \subfloat[Cartoon and Sketch as sources]
  {\includegraphics[width=0.24\textwidth]{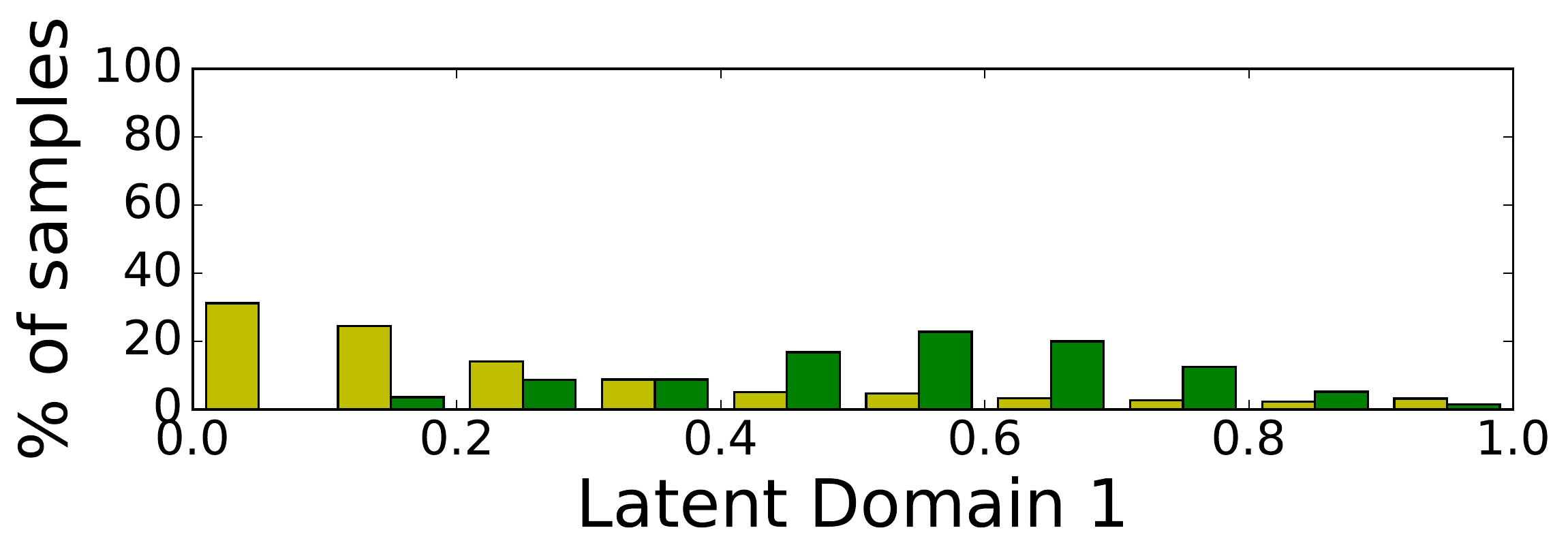}\label{fig:assignment-pa-s}} 
  \subfloat[Art and Photo as targets]{\includegraphics[width=0.25\textwidth,]{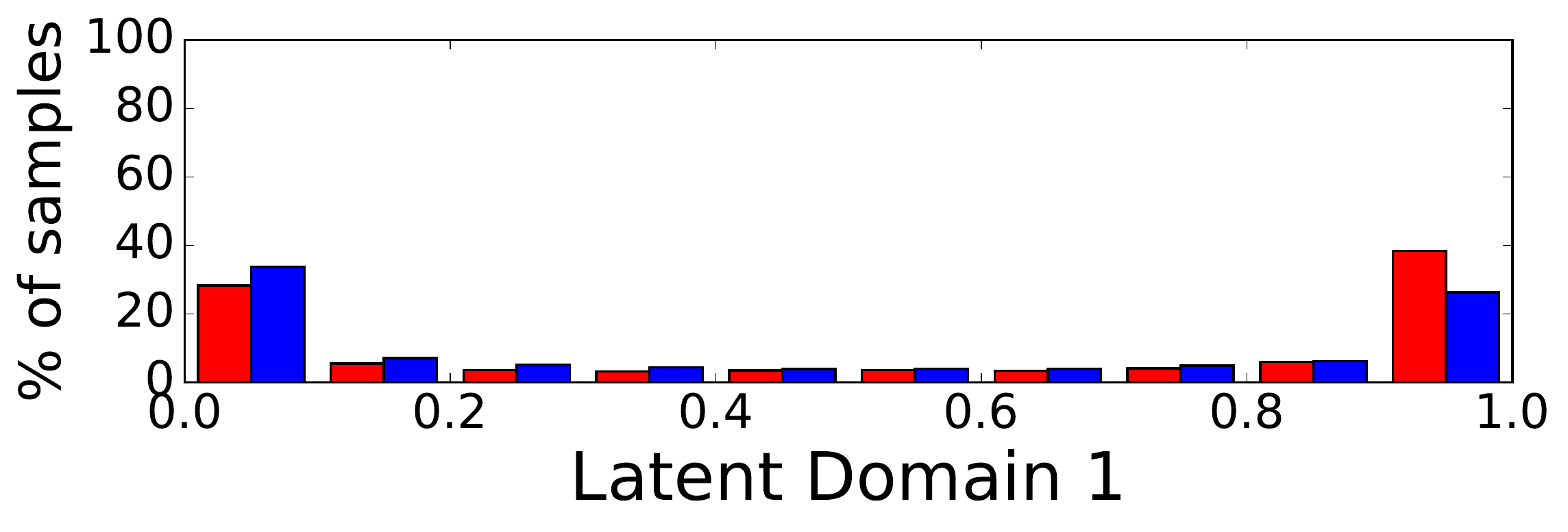}\label{fig:assignment-pa-t}}\hfill
  \subfloat[Art and Sketch as sources]
  {\includegraphics[width=0.24\textwidth]{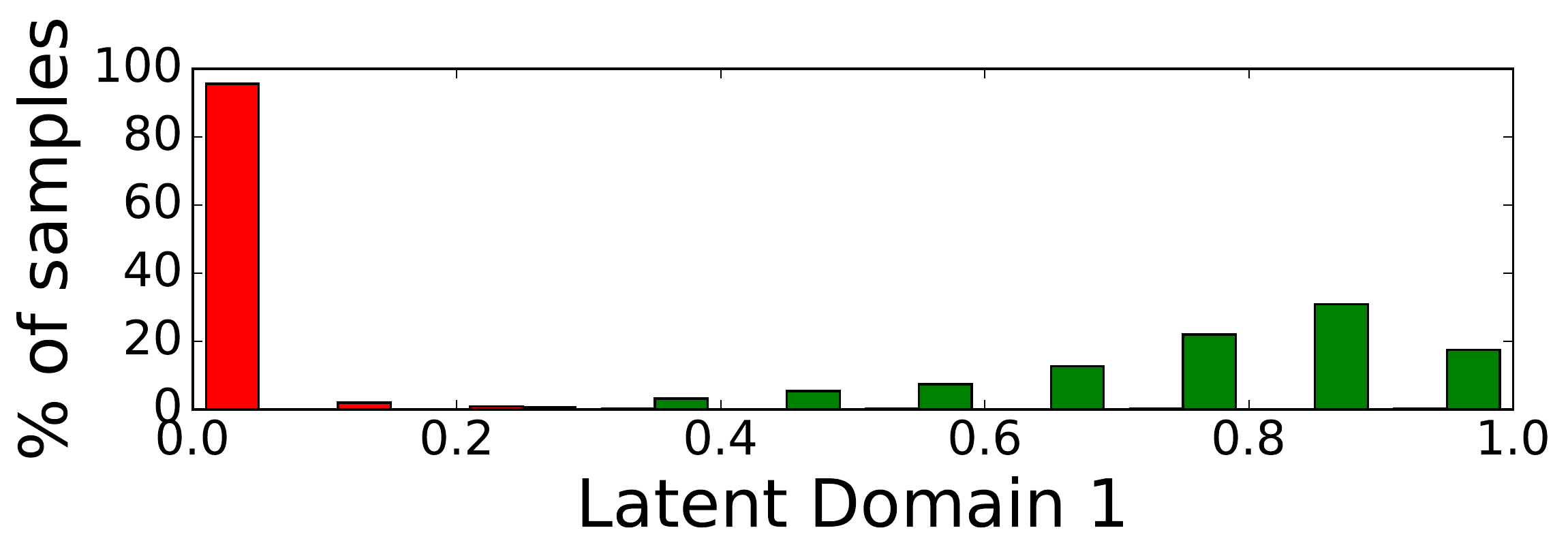}\label{fig:assignment-pc-s}}
    \subfloat[Cartoon and Photo as targets]{\includegraphics[width=0.25\textwidth,]{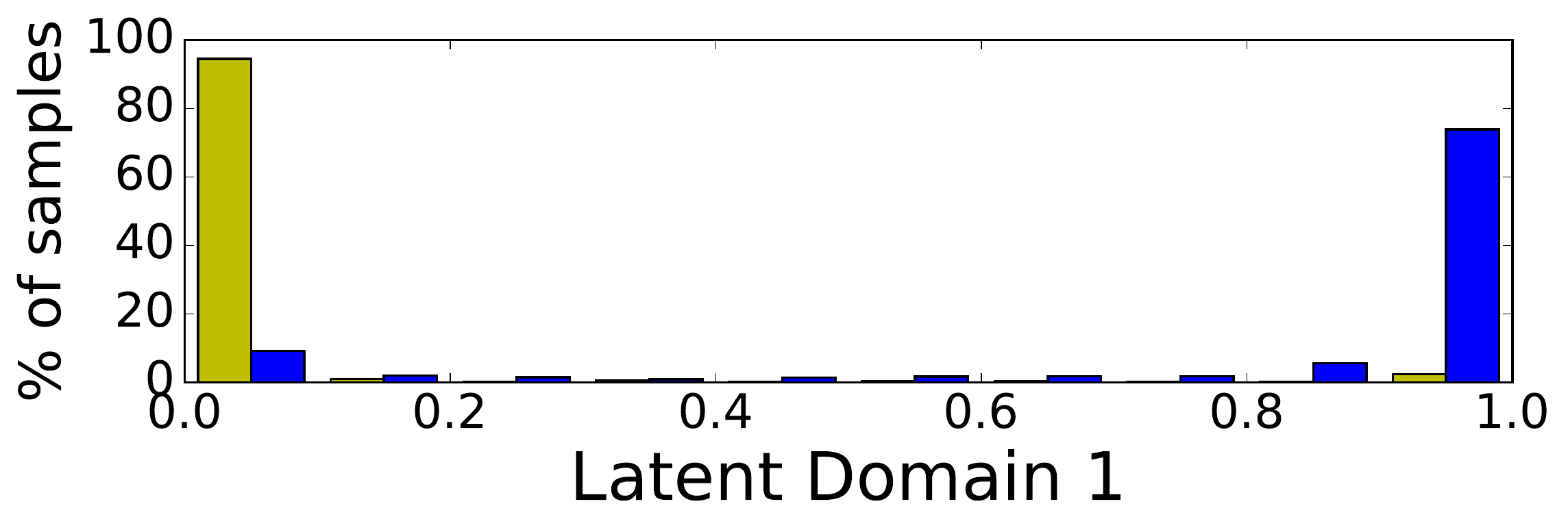}\label{fig:assignment-pc-t}}
  \\
    \subfloat[Art and Cartoon as sources]{\includegraphics[width=0.25\textwidth,]{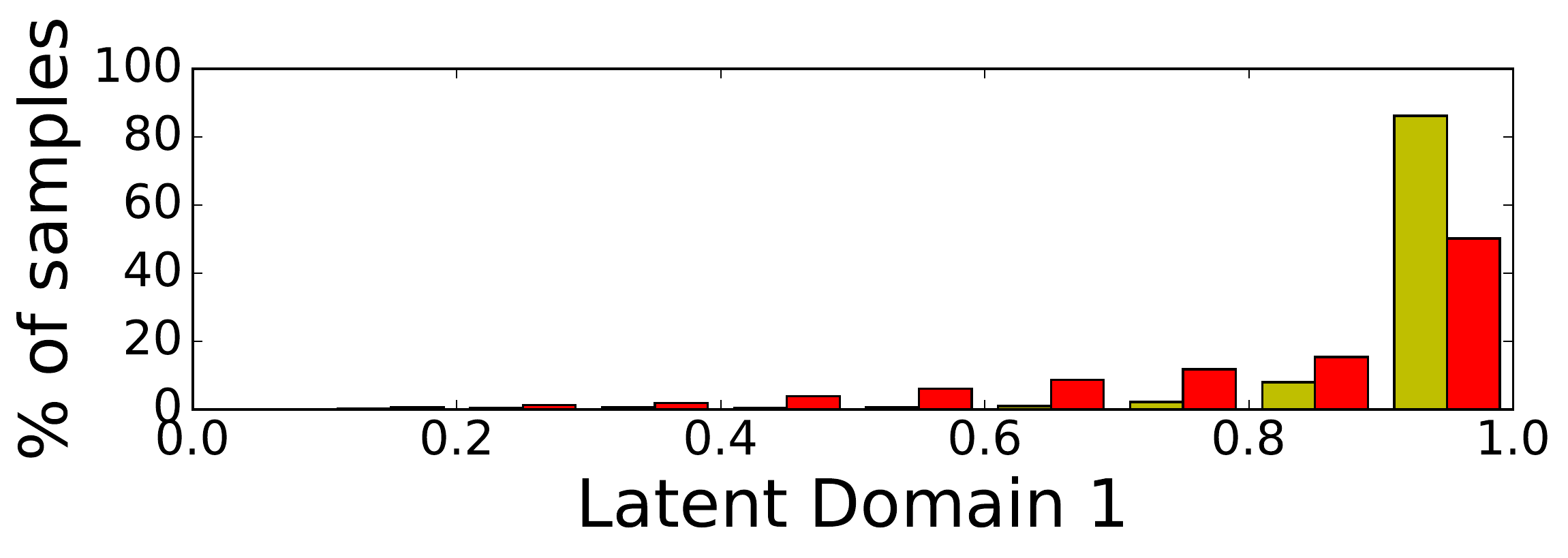}\label{fig:assignment-ps-s}}
  \subfloat[Photo and Sketch as targets]
  {\includegraphics[width=0.25\textwidth,]{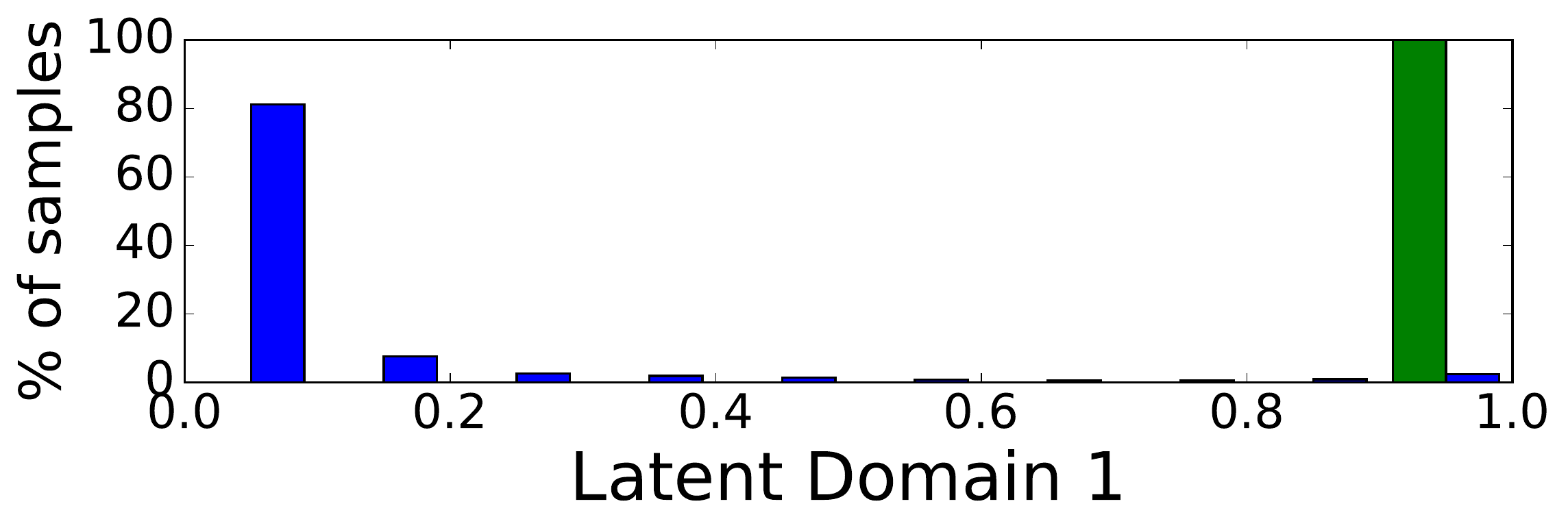}\label{fig:assignment-ps-t}}\hfill
   \subfloat[Photo and Sketch as sources]
  {\includegraphics[width=0.25\textwidth,]{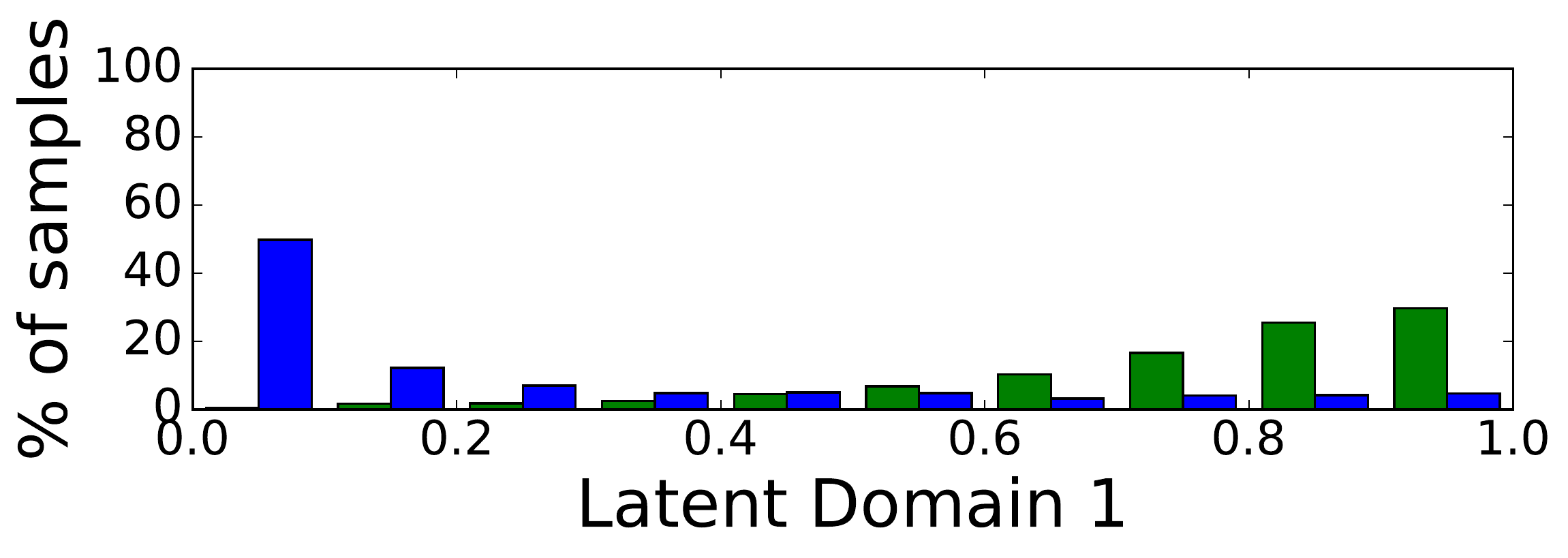}\label{fig:assignment-ac-s}}
    \subfloat[Art and Cartoon as targets]{\includegraphics[width=0.25\textwidth,]{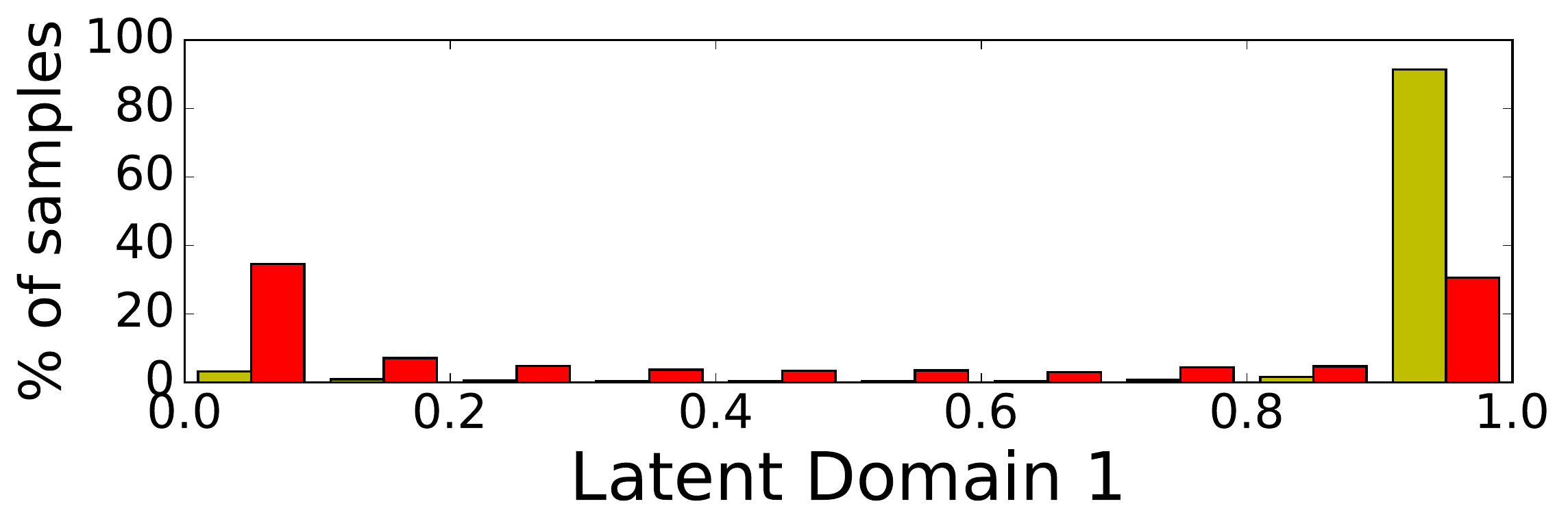}\label{fig:assignment-ac-t}}
 \\
 \subfloat[Cartoon and Photo as sources]
  {\includegraphics[width=0.25\textwidth,]{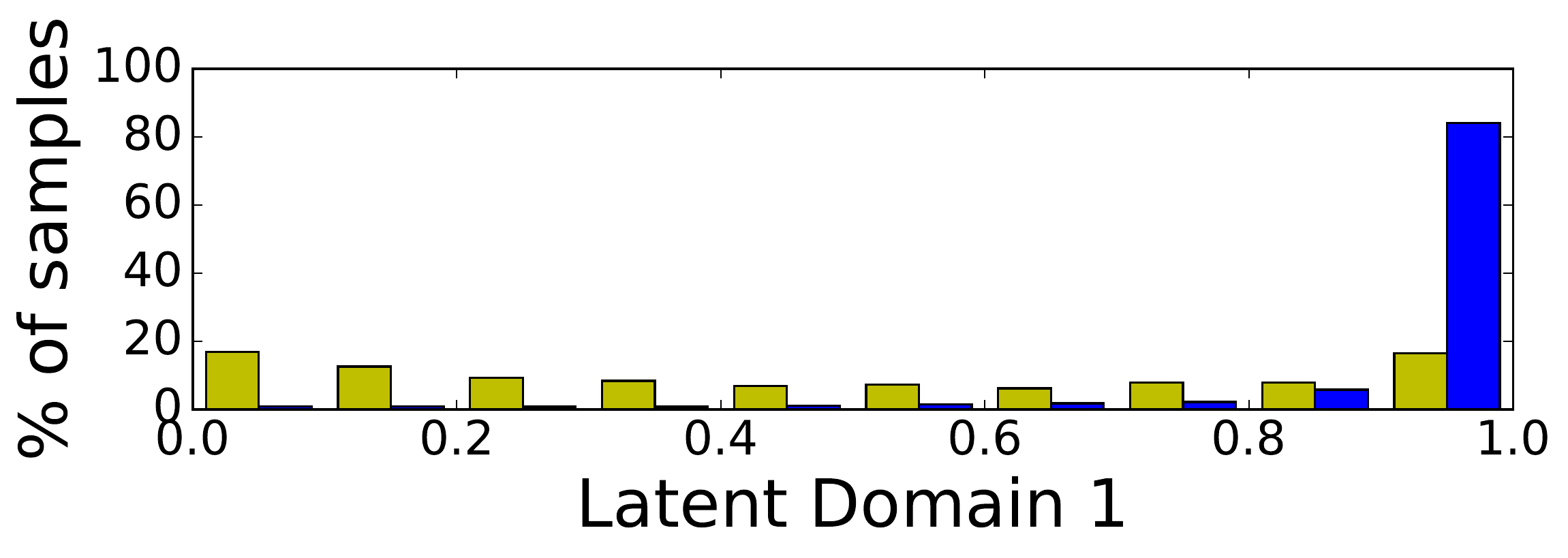}\label{fig:assignment-as-s}}
    \subfloat[Art and Sketch as targets]{\includegraphics[width=0.25\textwidth,]{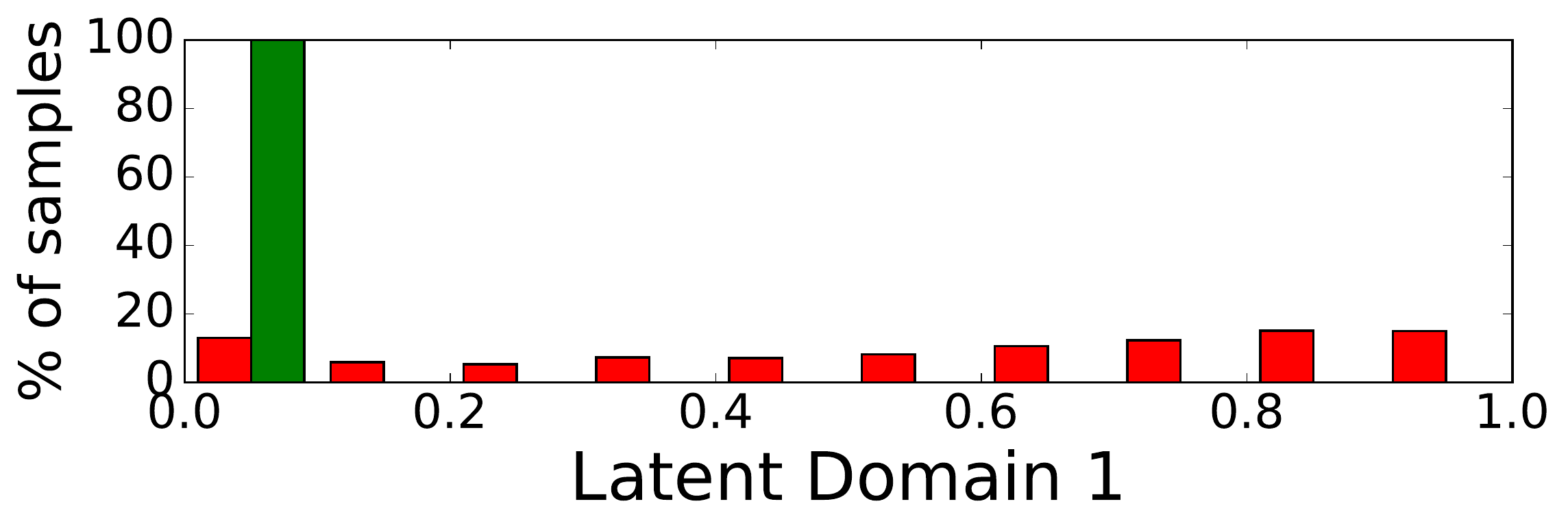}\label{fig:assignment-as-t}}\hfill
  \subfloat[Art and Photo as sources]
  {\includegraphics[width=0.25\textwidth,]{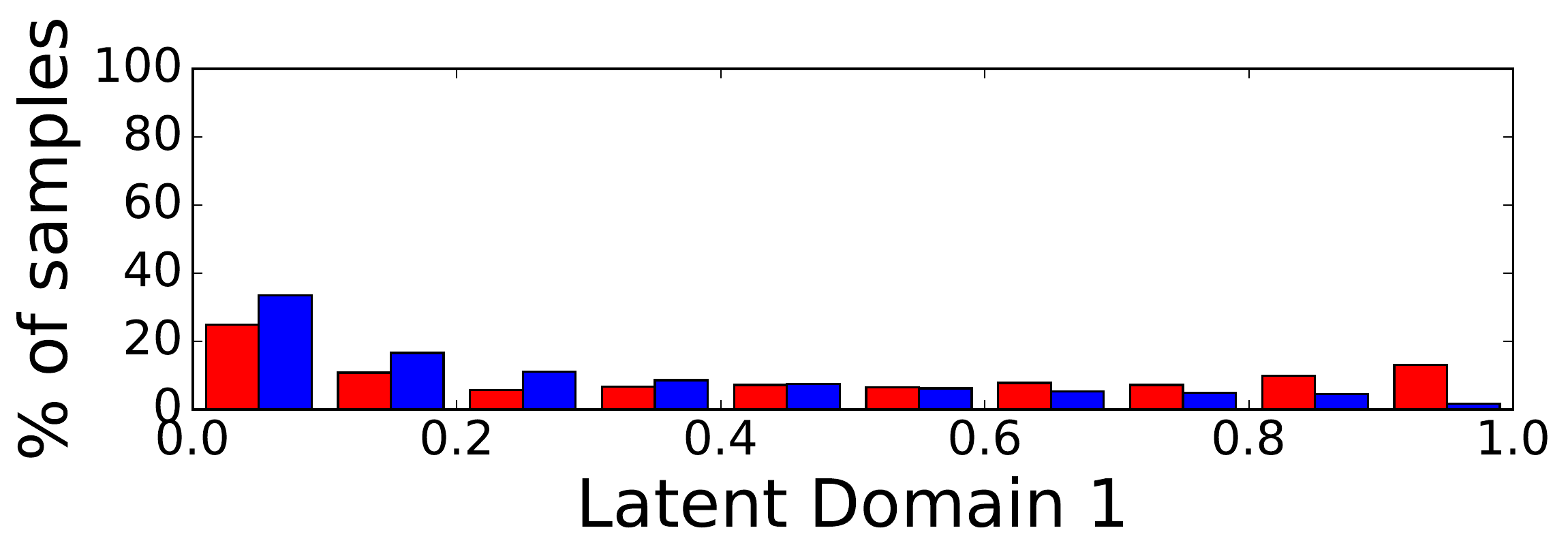}\label{fig:assignment-cs-s}}
    \subfloat[Cartoon and Sketch as targets]{\includegraphics[width=0.25\textwidth,]{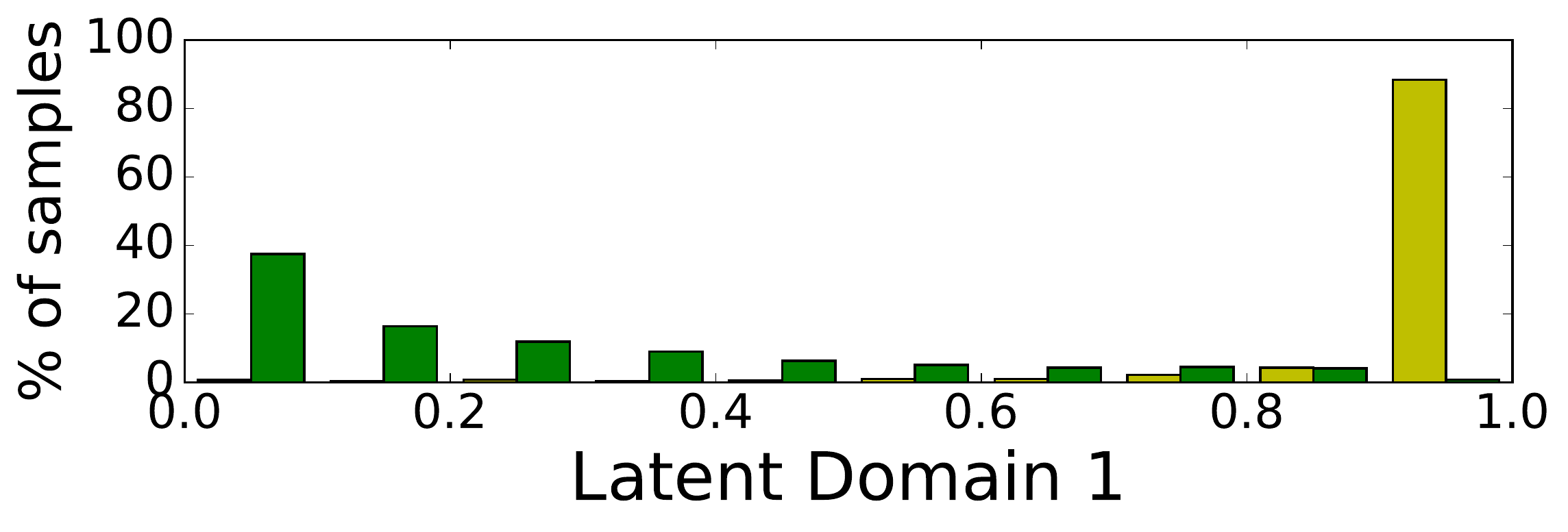}\label{fig:assignment-cs-t}}
  \\
  \caption{Distribution of the assignments produced by the domain prediction branch in all possible multi-target settings of the PACS dataset. Different colors denote different source domains (red: Art, yellow: Cartoon, blue: Photo, green: Sketch).
  }
  \vspace{-10pt}
  \label{fig:soft-assignment}
\end{figure*}

\begin{figure*}[t]
 \centering
 \subfloat[Cartoon and Sketch as sources]
  {\includegraphics[width=0.24\textwidth]{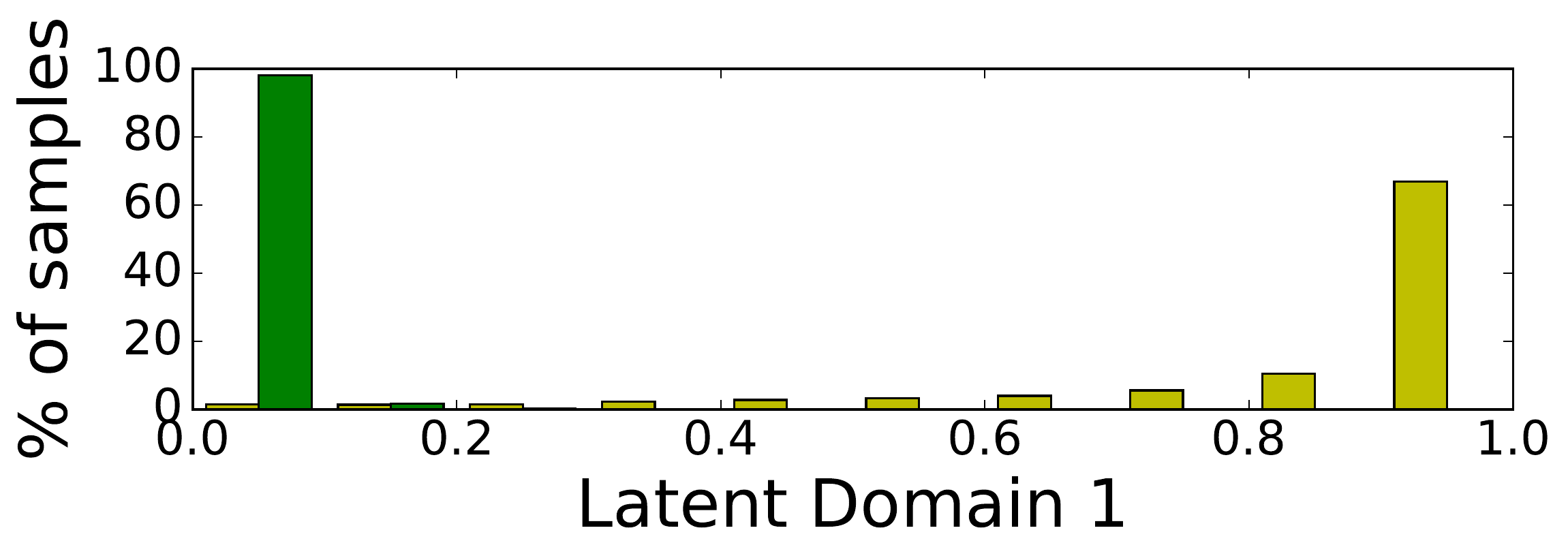}\label{fig:assignment-cluster-pa-s}} 
  \subfloat[Art and Photo as targets]{\includegraphics[width=0.25\textwidth,]{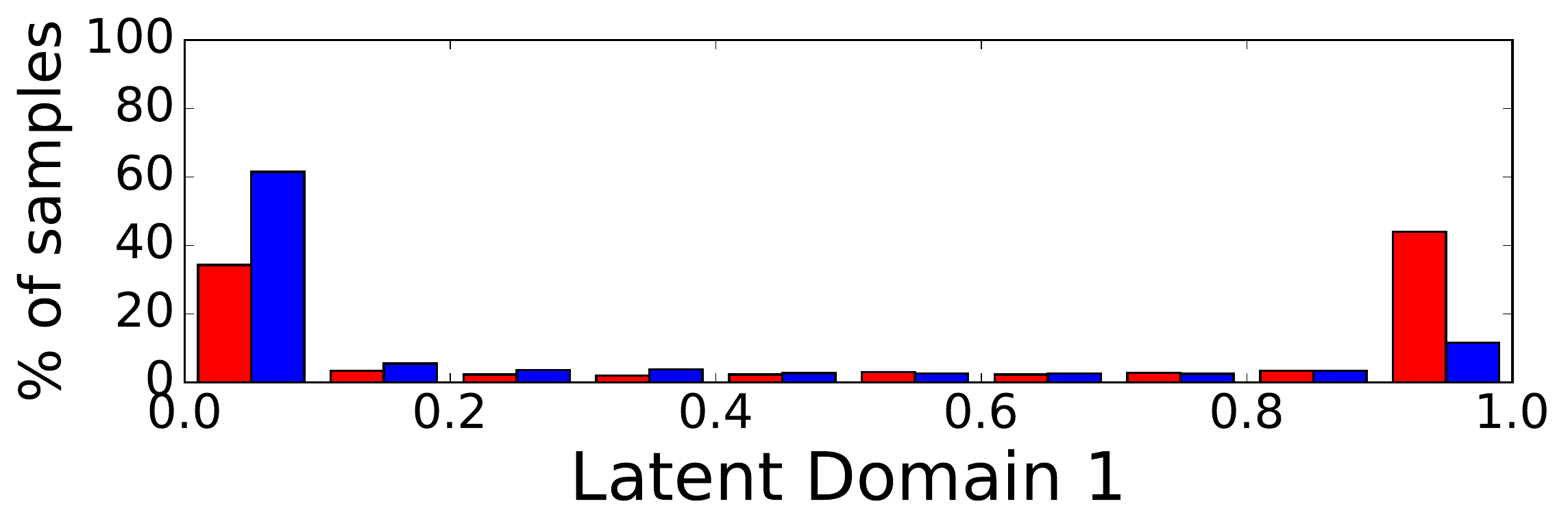}\label{fig:assignment-cluster-pa-t}}\hfill
  \subfloat[Art and Sketch as sources]
  {\includegraphics[width=0.24\textwidth]{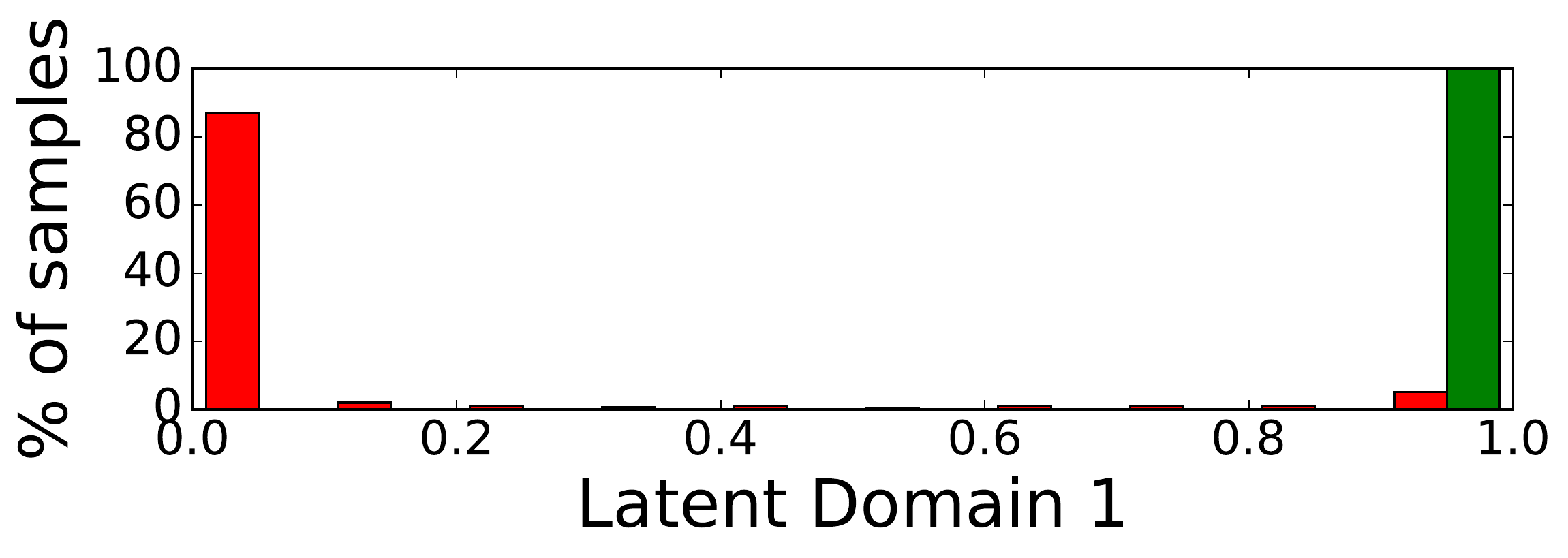}\label{fig:assignment-cluster-pc-s}}
    \subfloat[Cartoon and Photo as targets]{\includegraphics[width=0.25\textwidth,]{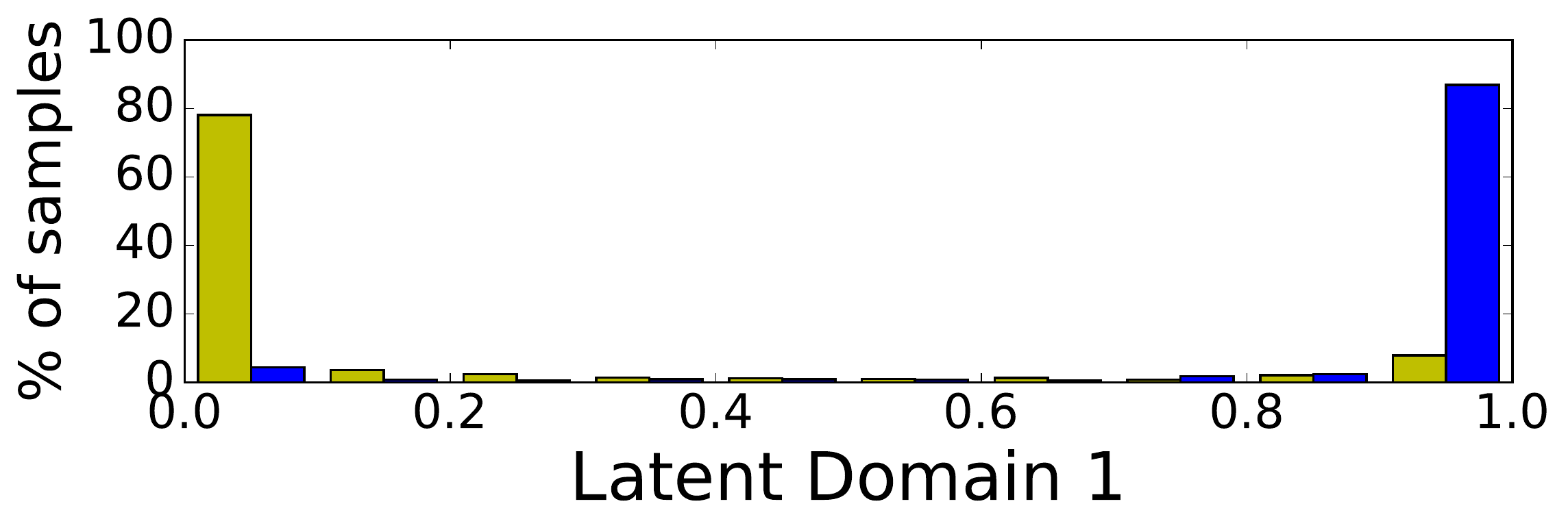}\label{fig:assignment-cluster-pc-t}}
  \\
    \subfloat[Art and Cartoon as sources]{\includegraphics[width=0.25\textwidth,]{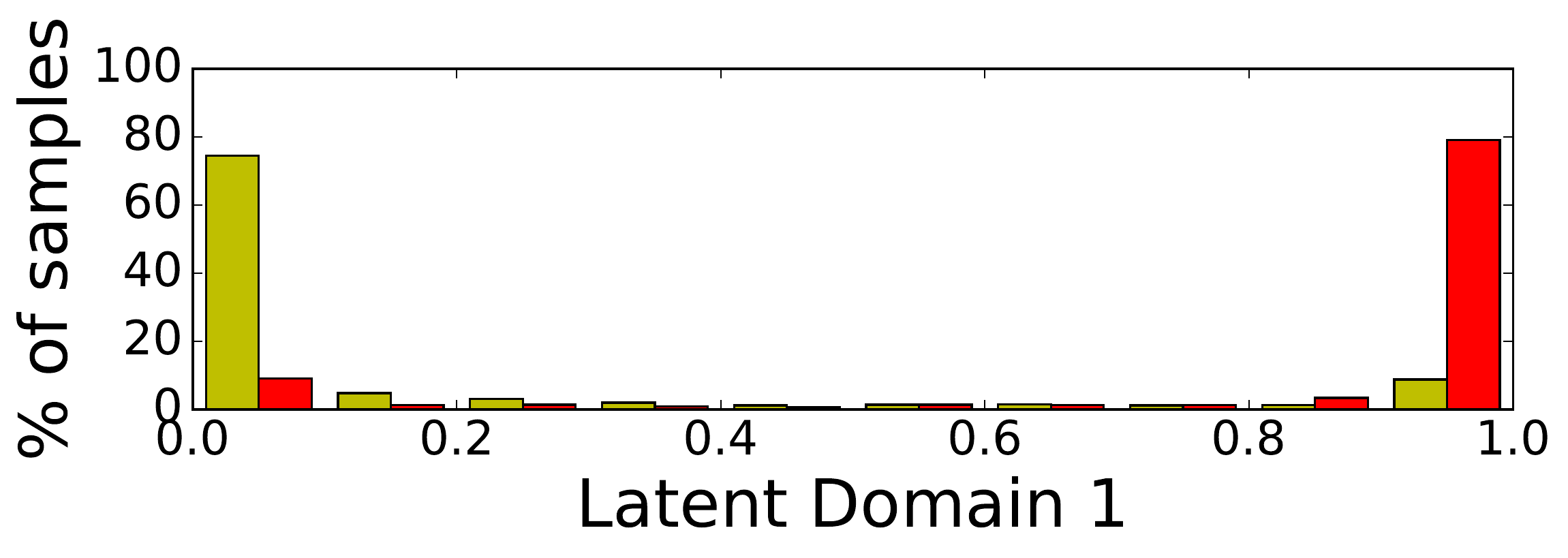}\label{fig:assignment-cluster-ps-s}}
  \subfloat[Photo and Sketch as targets]
  {\includegraphics[width=0.25\textwidth,]{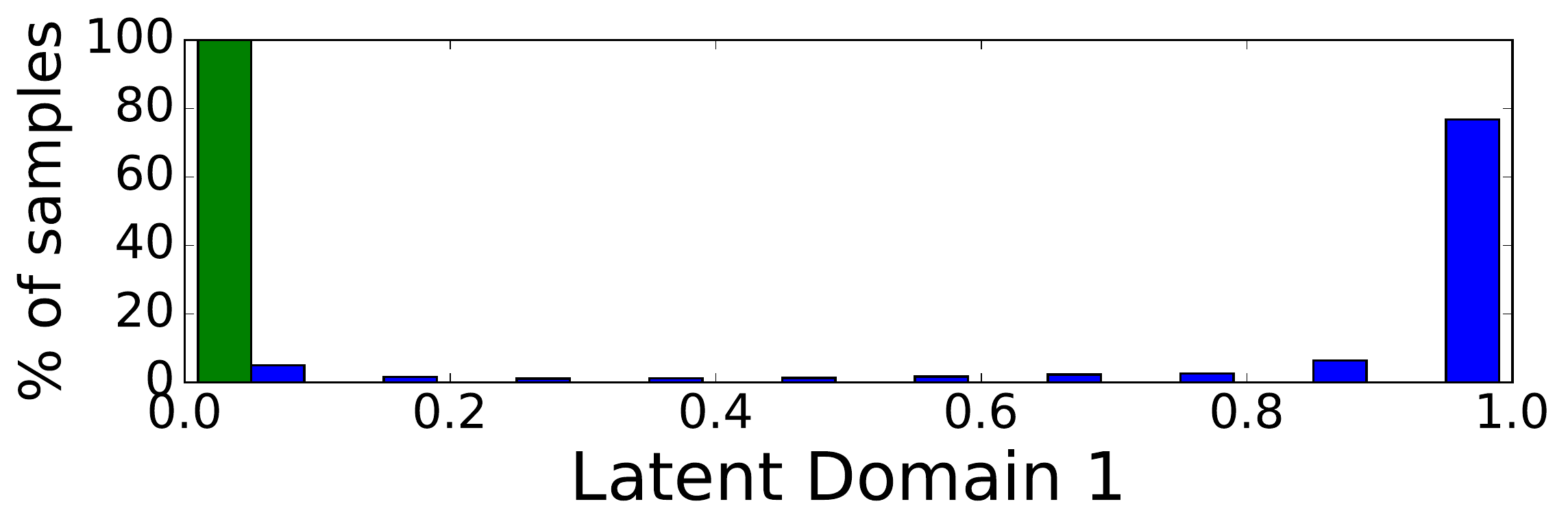}\label{fig:assignment-cluster-ps-t}}\hfill
   \subfloat[Photo and Sketch as sources]
  {\includegraphics[width=0.25\textwidth,]{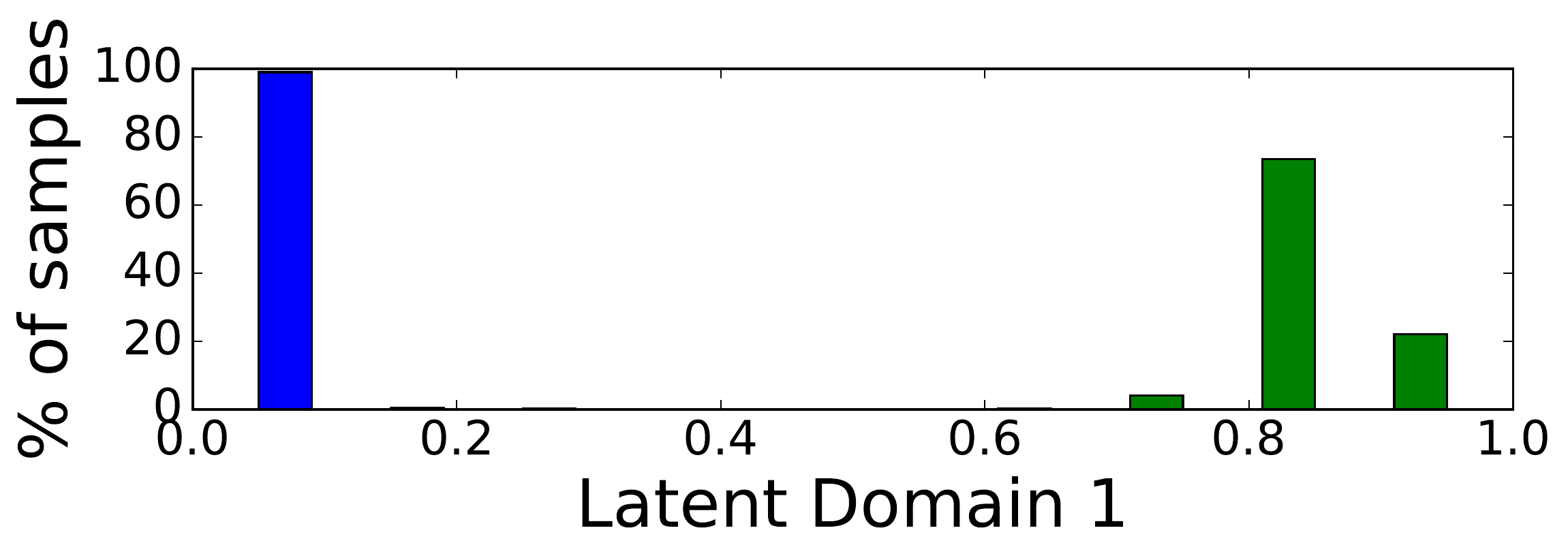}\label{fig:assignment-cluster-ac-s}}
    \subfloat[Art and Cartoon as targets]{\includegraphics[width=0.25\textwidth,]{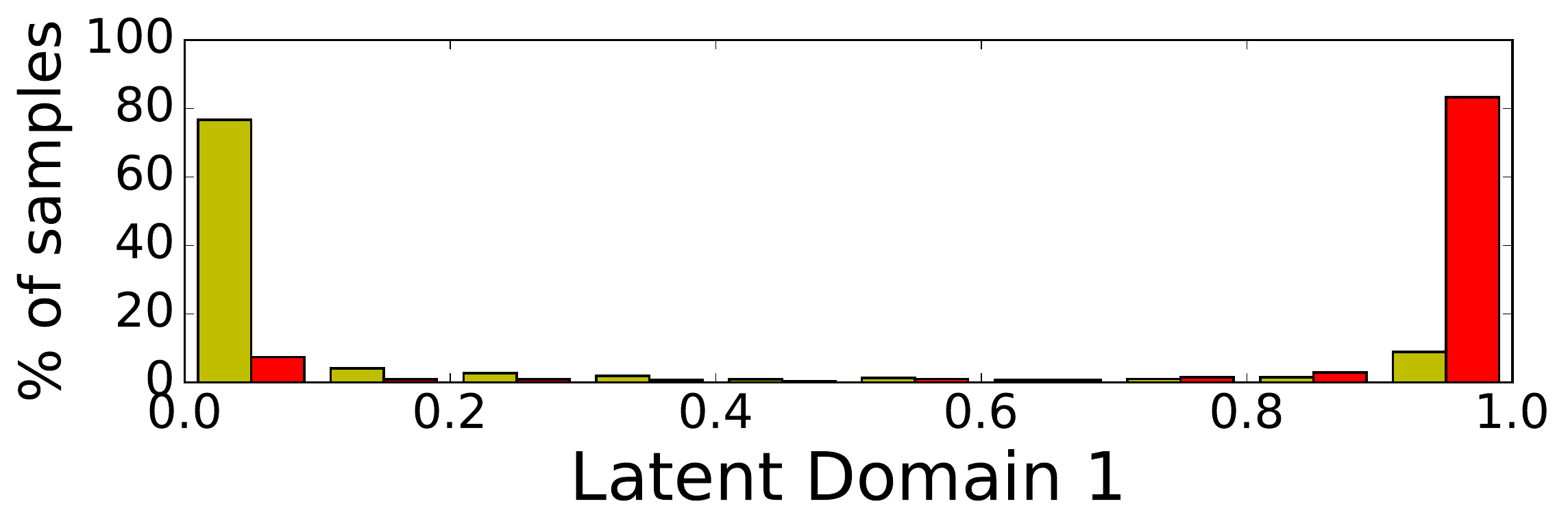}\label{fig:assignment-cluster-ac-t}}
 \\
 \subfloat[Cartoon and Photo as sources]
  {\includegraphics[width=0.25\textwidth,]{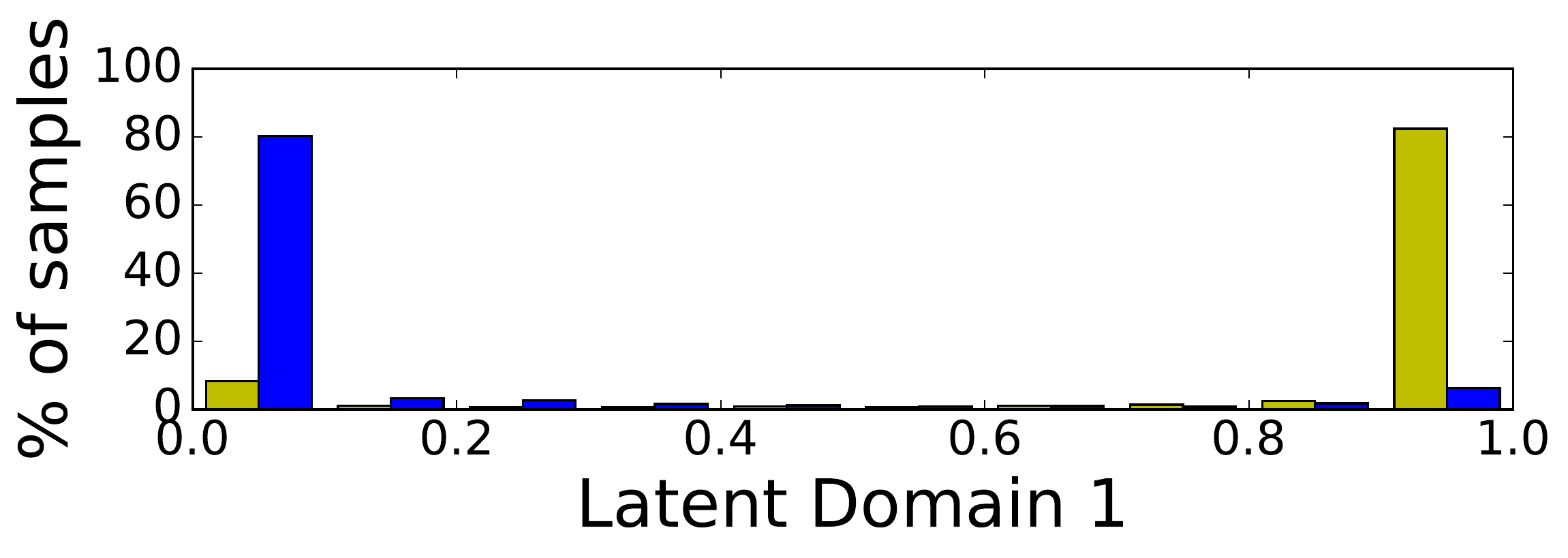}\label{fig:assignment-cluster-as-s}}
    \subfloat[Art and Sketch as targets]{\includegraphics[width=0.25\textwidth,]{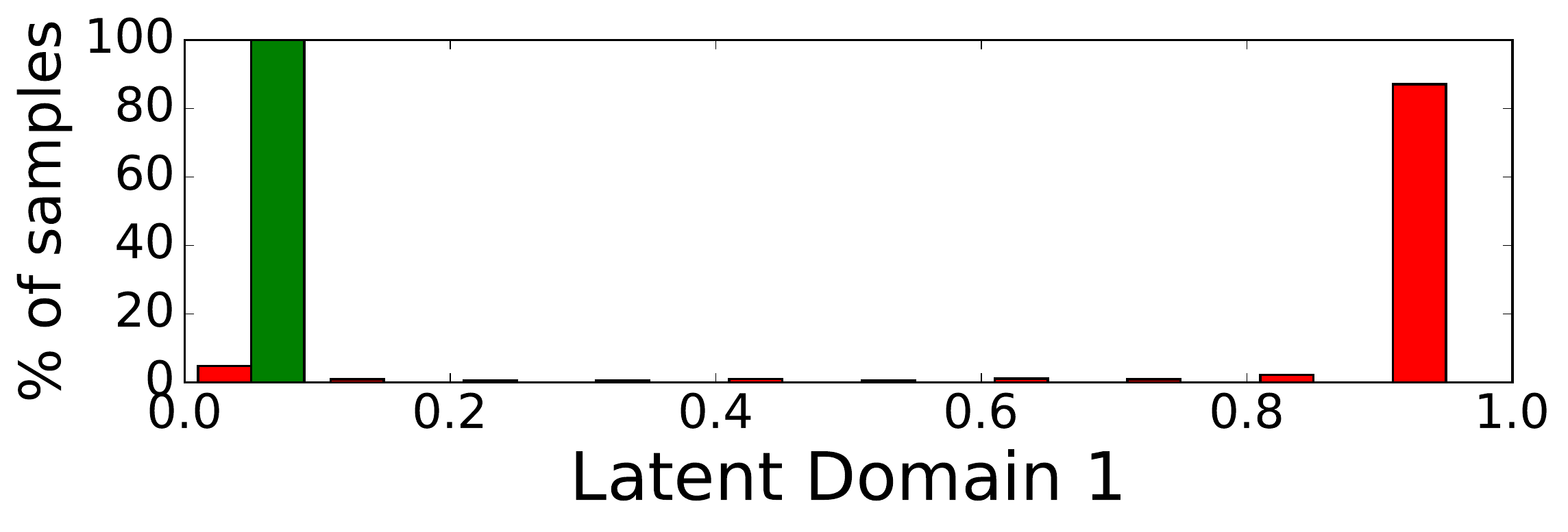}\label{fig:assignment-cluster-as-t}}\hfill
  \subfloat[Art and Photo as sources]
  {\includegraphics[width=0.25\textwidth,]{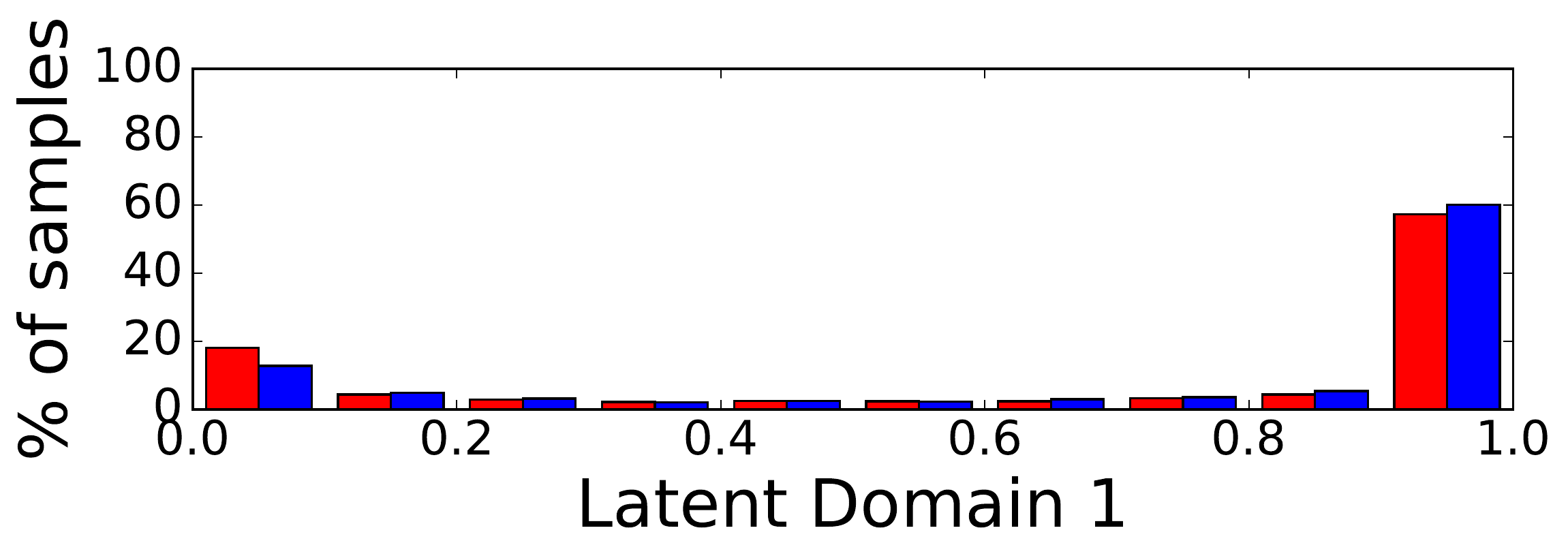}\label{fig:assignment-cluster-cs-s}}
    \subfloat[Cartoon and Sketch as targets]{\includegraphics[width=0.25\textwidth,]{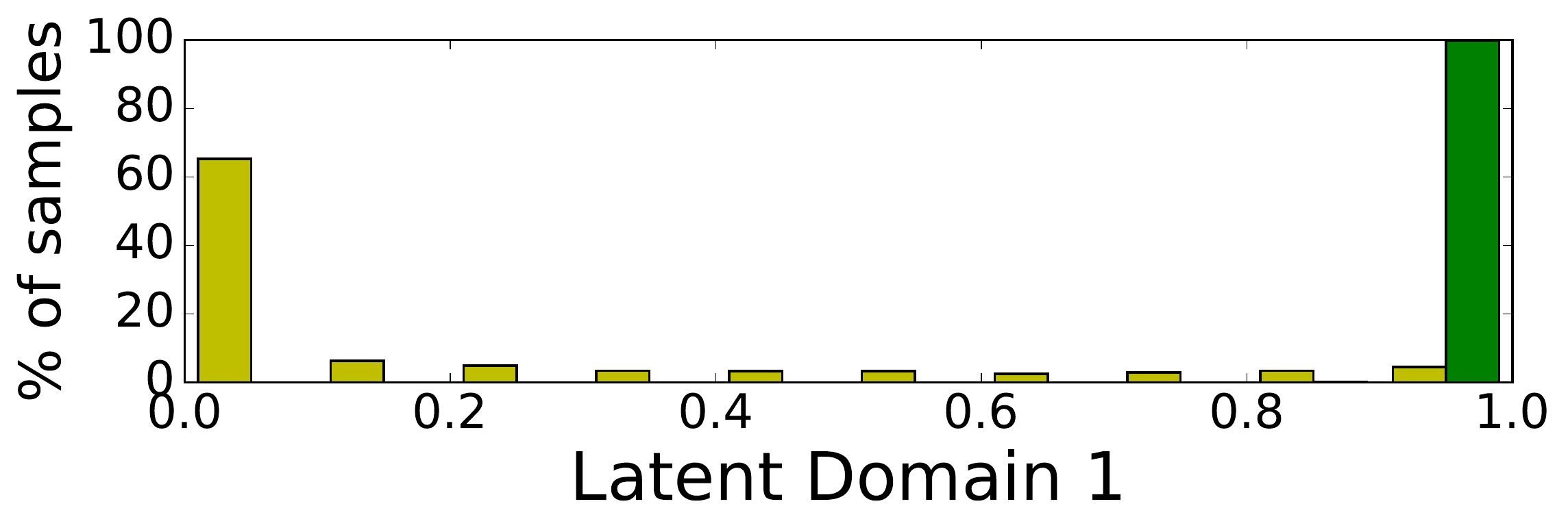}\label{fig:assignment-cluster-cs-t}}
  \\
  \caption{{Distribution of the assignments produced by the domain prediction branch trained with the additional constraint on the entropy loss in all possible multi-target settings of the PACS dataset. Different colors denote different source domains (red: Art, yellow: Cartoon, blue: Photo, green: Sketch).}
  }
  \vspace{-10pt}
  \label{fig:soft-assignment-cluster}
\end{figure*}

\begin{figure*}[t]
 \centering
 \subfloat[SVHN as target]
  {\includegraphics[width=0.25\textwidth]{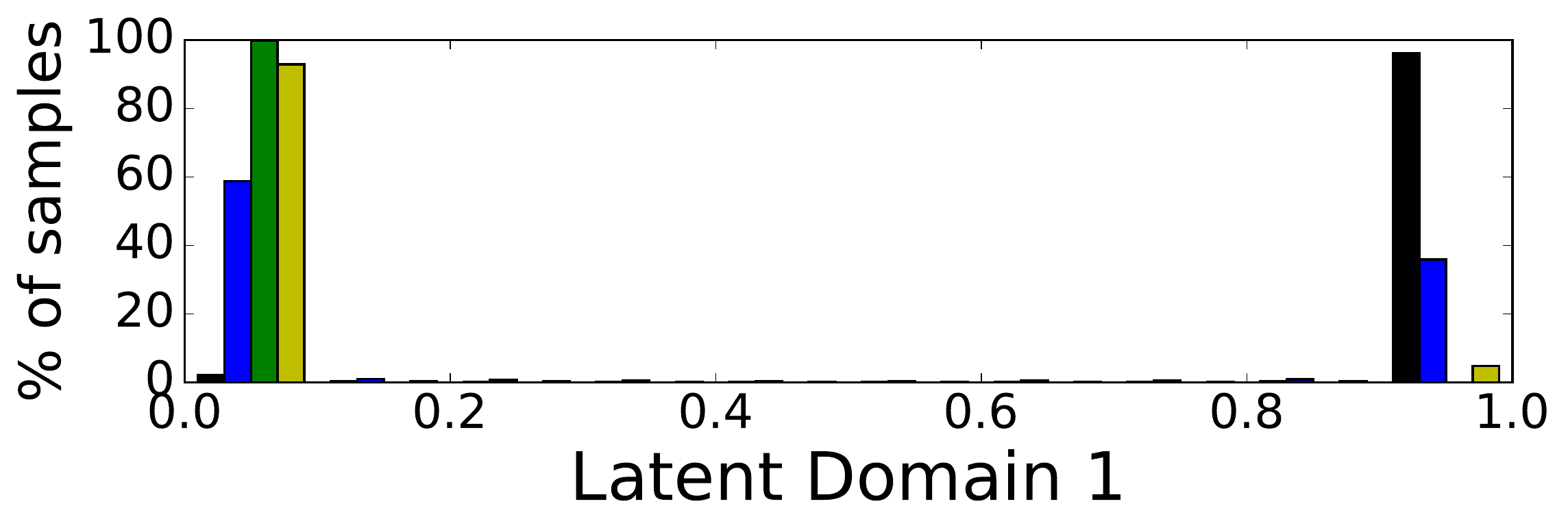}
  \includegraphics[width=0.25\textwidth,]{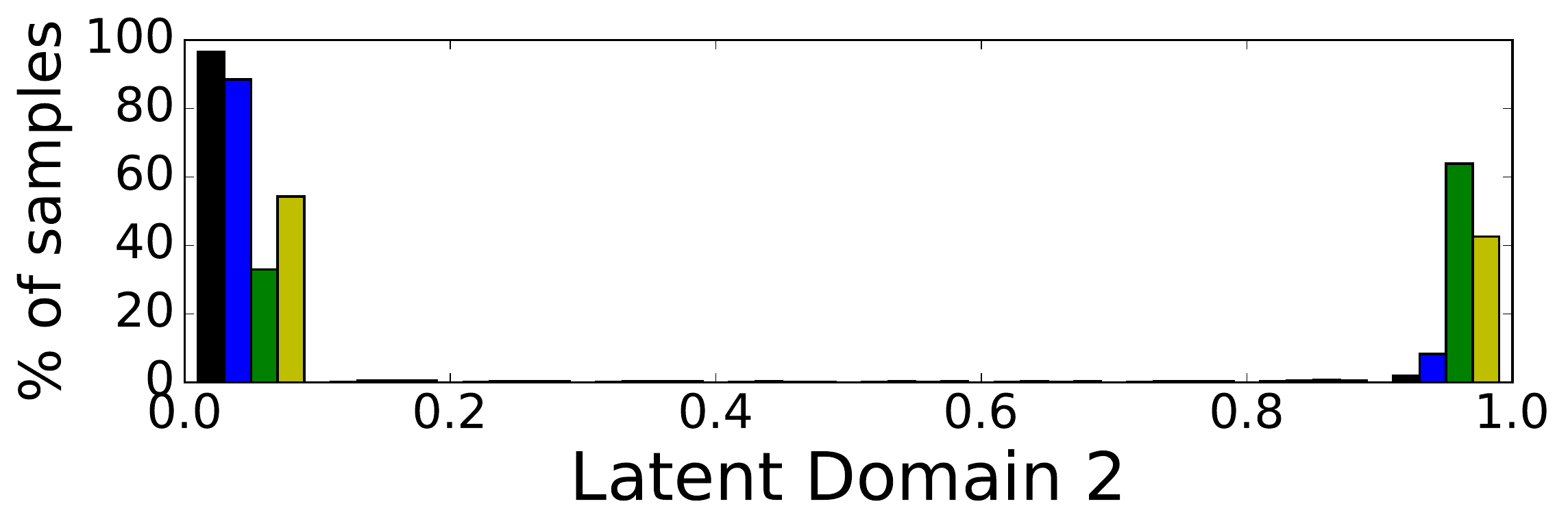}
  \includegraphics[width=0.25\textwidth]{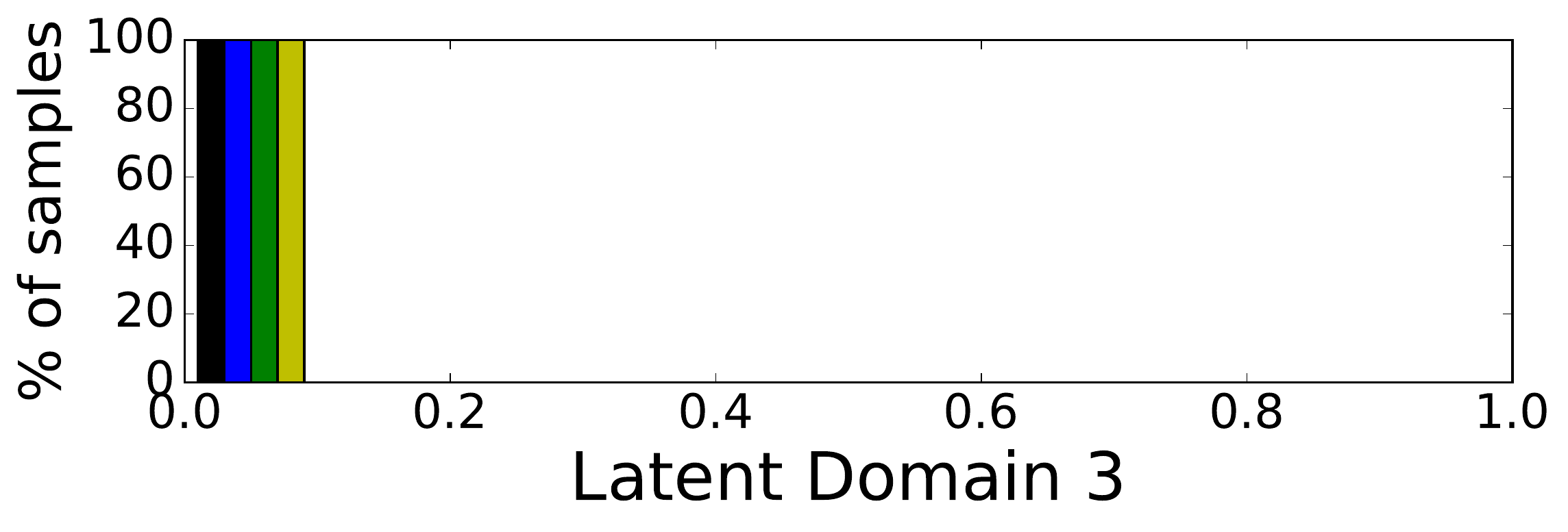}
   \includegraphics[width=0.25\textwidth,]{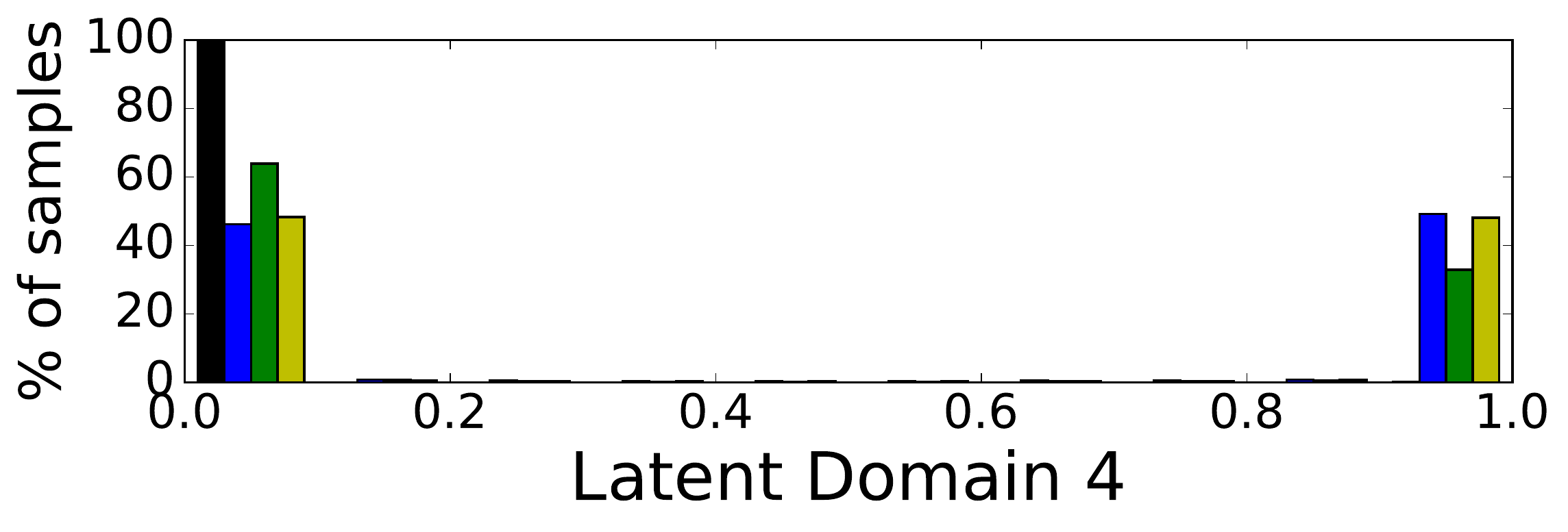}}
  \\
   \subfloat[MNIST-m as target]
 {\includegraphics[width=0.25\textwidth]{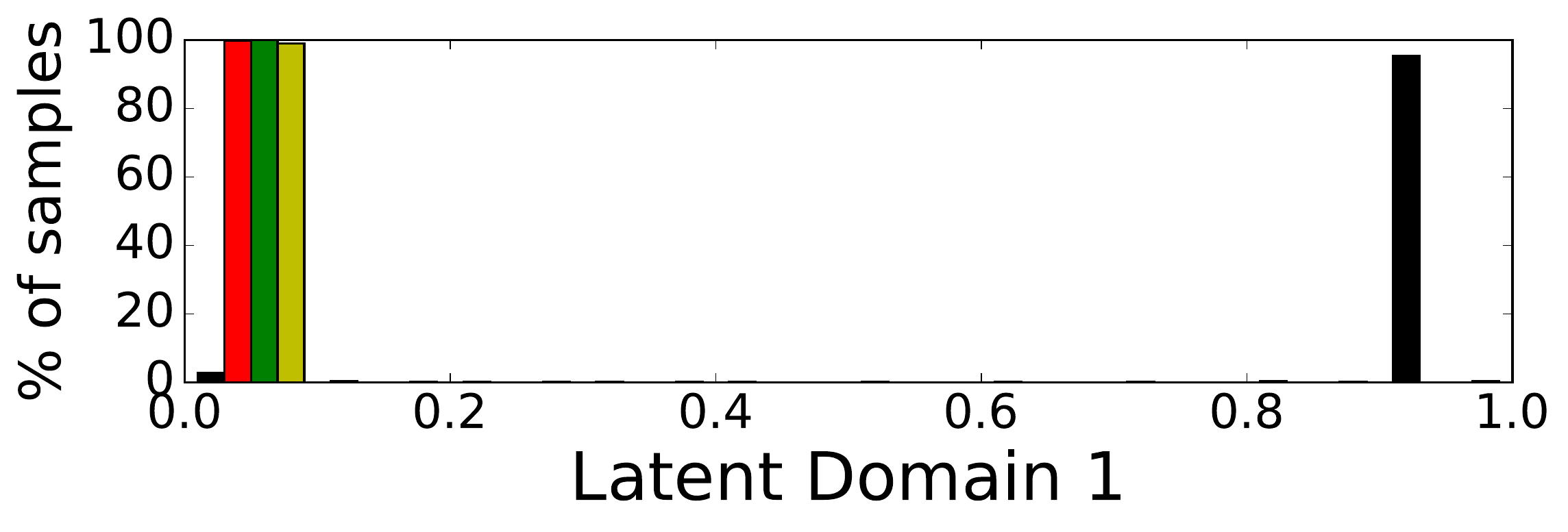}
  \includegraphics[width=0.25\textwidth,]{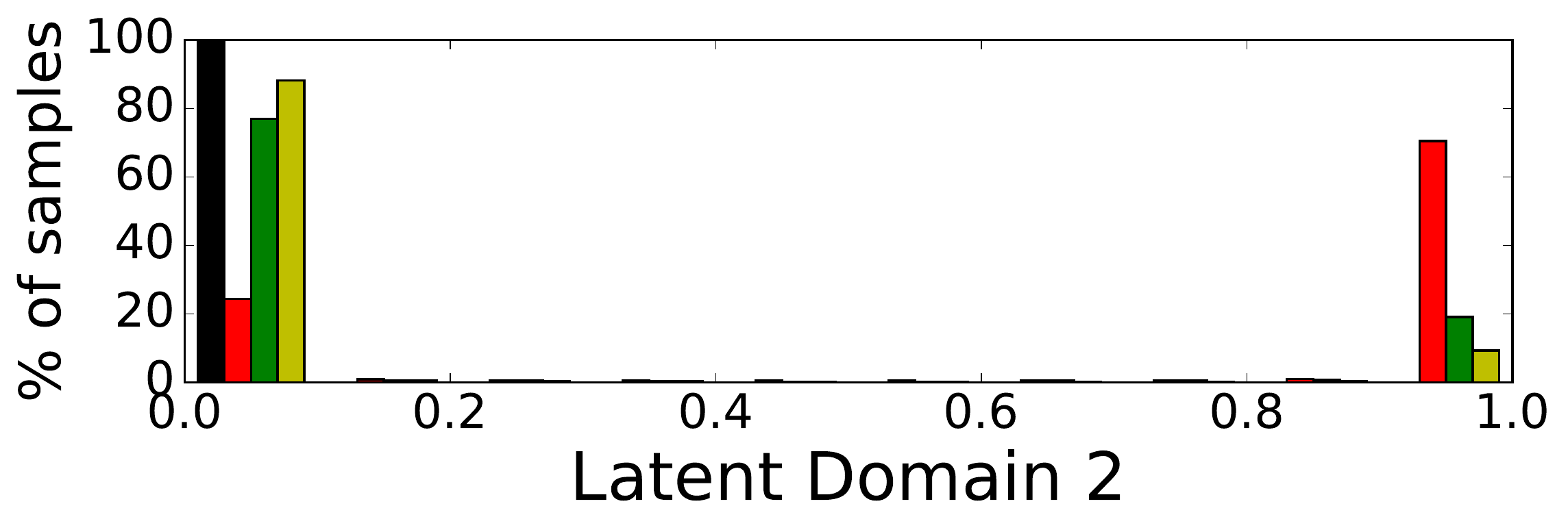}
  \includegraphics[width=0.25\textwidth]{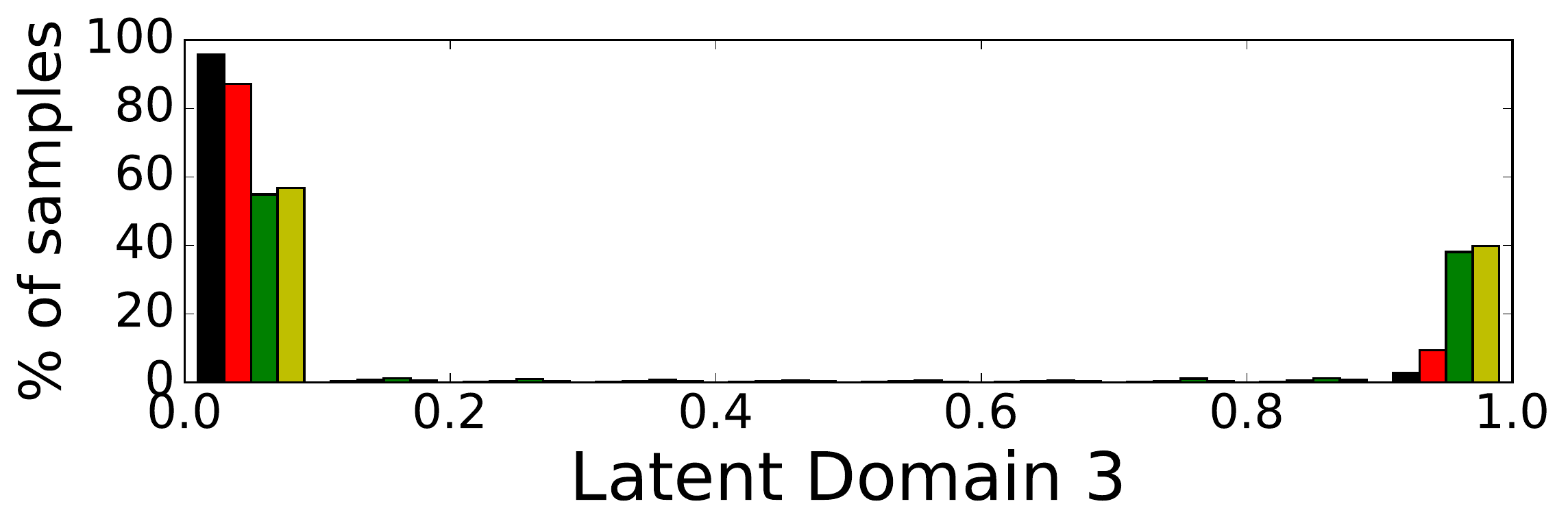}
   \includegraphics[width=0.25\textwidth,]{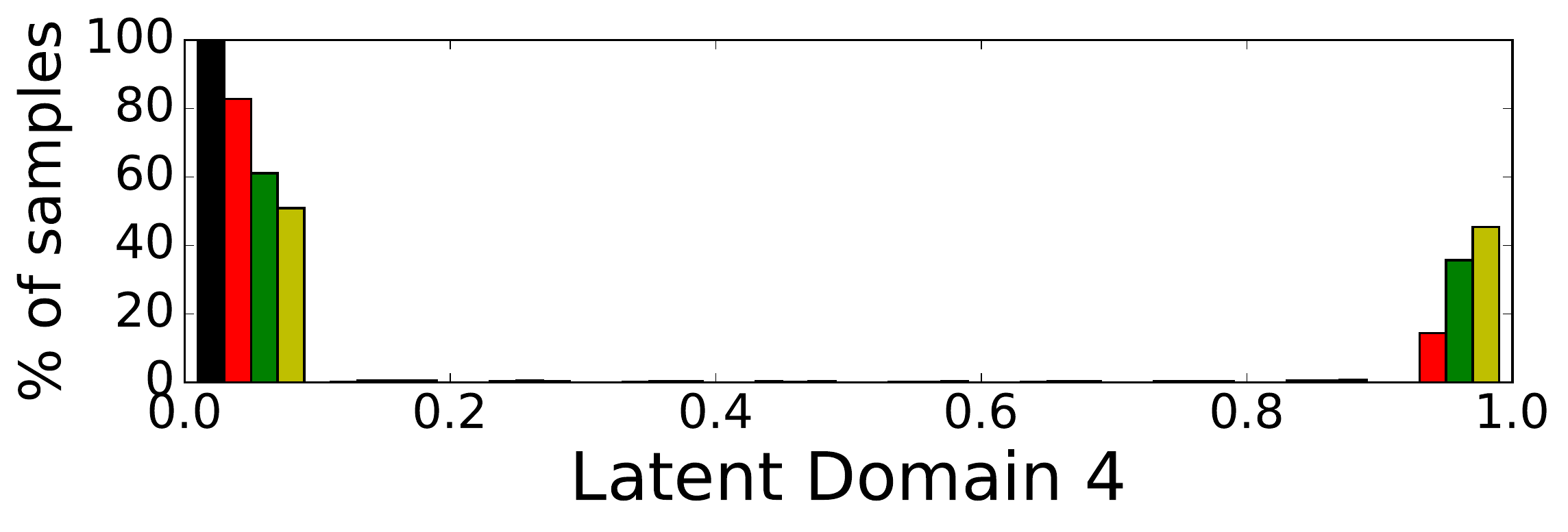}}
  
  \caption{{Distribution of the assignments produced by the domain prediction branch for each latent domain in all target settings of the Digits-five dataset. Different colors denote different source domains (black: MNIST, blue: MNIST-m, green: USPS, red: SVHN,  yellow: Synthetic numbers).}
  }
  \vspace{-10pt}
  \label{fig:soft-assignment-digits}
\end{figure*}

To analyze the performances of our approach in a multi-source multi-target scenario, we perform a second set of experiments on the PACS dataset considering 2 domains as sources and the other 2 as targets.
The results, shown in Table~\ref{tab:pacs-multitarget}, comprise the same baselines as in Table~\ref{tab:pacs}.
Note that, apart from the difficulty of providing useful domain assignments both in the source and target sets during training, the domain prediction step is required even at test time, thus having a larger impact on the final performances of the model.
The performance gap between DIAL and our approach increases in this setting compared to Table~\ref{tab:pacs}.
Our hypothesis is that not accounting for multiple domains has a larger impact on the \emph{unlabeled} target than on the \emph{labeled} source.
Looking at the partial results, when Photo is considered as one of the target domains there are no particular differences in the final performances of the various DA models: this may be caused by the bias of the pretrained network towards this domain. However, when the other domains are considered as targets, the gain in performances produced by our model are remarkable. When Sketch is one of the target domains, our model 
completely fills the gap between the unified source\slash{}target DA method and the multi-source multi-target upper bound with a gain of more then 7\% when Art and Cartoon considered as other target. {Setting $\lambda_B > 0$ in this setting allows to obtain a further boost of performances. This is evident in the scenario where Photo and Art are both the source or target domains, with Cartoon-Sketch correspond to the other pair. In this scenario the source/target pairs are quite close and enforcing a uniform assignment among the latent domains allows to obtain a better estimate of the latent domain.}

\myparagraph{Ablation study.}
We exploit the challenging multisource-multitarget scenario of Table \ref{tab:pacs-multitarget} in order to assess the impact of the various components of our algorithm. In particular we show how the performance are affected if (i) a random domain is assigned to each sample; (ii) no loss is applied to the domain prediction branch; (iii) no entropy loss is applied to the classification of unlabeled target samples. From Table \ref{tab:pacs-multitarget}  we can easily notice that if we drop either the domain prediction branch (random assignment) or the losses on top of it ($\lambda_E=\lambda_B=0$), the performances of the model become comparable to the ones obtain by the DIAL baseline. This shows not only the importance of discovering latent domains, but also that both the domain branch and our losses allow to extract meaningful subsets from the data. Moreover, this demonstrates the fact that our improvements are not only due to the introduction of multiple normalization layers, but also to the latent domain discovering procedure.
{For what concerns the classification branch, without the entropy component on unlabelled target samples ($\lambda_C=0$), the performance of the model significantly decreases (\ie from 82.6 to 76.4 in average). This confirms the findings of previous works \cite{carlucci2017autodial,carlucci2017just} about the impact that this loss for normalization based DA approaches. In particular, assuming that source and target samples of different domains are independently normalized, the entropy loss generates a gradient flow through unlabeled samples based in the direction of its most confident prediction. This is particularly important to learn useful features even for the target domain/s, for which no supervision is available.}

\myparagraph{In-depth analysis.}
The ability of our approach to discover latent domains is further investigated on PACS.
First, in Figure~\ref{fig:max-assignment}, we show how our approach assigns source samples to different latent domains in the single target setting.
The four plots correspond to a single run of the experiments of Table \ref{tab:pacs}.
Interestingly, when either Cartoon (Figure~\ref{fig:assignment-cartoon}) or Sketch (Figure~\ref{fig:assignment-sketch}) is the target, samples from Photo and Art tend to be associated to the same latent domain and, similarly, when either Photo (Figure~\ref{fig:assignment-photo}) or Art (Figure~\ref{fig:assignment-art}) is the target, samples from Cartoon and Sketch are mostly grouped together.
These results confirms the ability of our approach to automatically assign images of similar visual appearance to the same latent distribution.
In Figure~\ref{fig:top-k}, we show the top-6 images associated to each latent domain for each sources\slash{}target setting.
In most cases, images associated to the same latent domain have similar appearance, while there is high dissimilarity between images associated to different latent domains.
Moreover, images assigned to the same latent domain tend to be associated with one of the original domains.
For instance, the first row of Figure~\ref{fig:topk-photo} contains only images from Art, while the third contains only images from Sketch.
Note that no explicit domain supervision is ever given to our method in this setting.

\begin{table*}[t]
			\caption{PACS dataset: comparison of different methods using the ResNet architecture on the multi-source multi-target setting. The first row indicates the two target domains. }
		\centering
		\scalebox{.95}{
		\begin{tabular}{ l | c  c c  c c c | c } 
			\hline
			Method  & Photo-Art & Photo-Cartoon & Photo-Sketch&Art-Cartoon&Art-Sketch&Cartoon-Sketch & Mean\\
            	\hline
                ResNet \cite{he2016deep} &71.4&84.2&81.4&62.2&70.3&54.2&70.6\\
DIAL \cite{carlucci2017just} &{86.7} &86.5 &86.8 &77.1 &72.1 &67.7 &79.5 \\
Random assignment &86.6&86.7&85.9&76.2&69.1&69.4&79.1\\
Ours $\lambda_E=\lambda_B=0$ &86.8&86.5&86.7&78.6&73.8&68.7&80.2\\
Ours $\lambda_B=\lambda_C=0$& 82.4&85.0&83.7&71.7&74.0&68.8&76.4\\
Ours $\lambda_B=0$ &86.1&{87.9}&{87.9}&\textbf{79.3}&{79.9}&{74.9}&{82.6}\\


Ours &\textbf{87.2}&\textbf{88.1} &\textbf{88.7} &77.7&\textbf{81.3}&\textbf{77.0}&\textbf{83.3} \\\hline\hline
Multi-source/target DA & 87.7&88.9&86.8&79.0&79.8 &75.6&83.0\\ \hline 
		\end{tabular}
        }
         \vspace{-10pt}
		\label{tab:pacs-multitarget}
\end{table*}



In Figure~\ref{fig:soft-assignment}, we show the histograms of the domain assignment probabilities predicted by our model with $\lambda_B=0$ in the various multi-source, multi-target settings of Table~\ref{tab:pacs}.
As the figures shows, in most cases the various pairs of target domains tend to be very well separated: this justifies the large gain of performances produced by our model in this scenario.
The only cases where the separation is less marked is when Art and Photo, which have very similar visual appearance, are considered as targets.
On the other hand, source domains are not always as clearly separated as the targets.
In particular the pairs Photo-Cartoon, Art-Photo and Art-Cartoon, tend to receive similar assignments when they are considered as source.
A possible explanation is that the supervised source loss could have a stronger influence on the domain assignment than the unsupervised target one.
In any case, note that these results do not detract from the validity of our approach.
In fact, our main objective is to obtain a good classification model for the target set, independently from the actual domain assignments we learn.

{In Figure~\ref{fig:soft-assignment-cluster}, the same analysis is performed on our method with the additional constraint of having a uniform assignment distribution among domains. As the figure shows, this constraint allows to obtain a clearer domain separation in most of the cases, overcoming the difficulties that the domain prediction branch experienced in separating domain pairs such as Photo-Cartoon and Photo-Art.}

{We perform a similar analysis in another dataset, Digits-five. The results are reported in Figure~\ref{fig:soft-assignment-digits}. As the figure shows, when SVHN is the target domain, one of the latent domains (latent domain 1) receives very confident assignments for the samples of the MNIST dataset. The samples of the other source datasets receive assignment spread through all the latent domains, with the exceptions of USPS which receives the most confident predictions for the second latent domain and MNIST-m, which partially influences the first latent domain, the one with confidence assignments to MNIST. One latent domain does not receive assignments form any of the sources (latent domain three): this might happen if the entropy term overcomes the uniform assignment constraints in the early stages of training. Similarly, when MNIST-m is the target domain, the first two latent domains receive confident assignments for samples belonging to MNIST and SVHN datasets respectively, while the third and the fourth receive higher assignments for samples of the remaining source domains.}

\subsubsection{Experiments on Office-31}

In our Office-31 experiments we consider the following baselines, trained on the union of the source sets: (i) a plain AlexNet network; (ii) AlexNet with BN inserted after each fully-connected layer; and (iii) AlexNet + DIAL~\cite{carlucci2017just}.
Additionally, we consider single source domain adaptation approaches, using the results reported in~\cite{xu2018deep}.
The methods are Transfer Component Analysis (TCA) \cite{pan2011domain},  Geodesic Flow Kernel (GFK) \cite{gong2012geodesic}, Deep Domain Confusion (DDC) \cite{tzeng2015simultaneous}, Deep Reconstruction Classification  Networks (DRCN) \cite{ghifary2016deep} and  Residual  Transfer  Network  (RTN) \cite{long2016unsupervised}, as well as the  Reversed  Gradient (RevGrad) \cite{ganin2014unsupervised} and Domain Adaptation Network (DAN) \cite{long2015learning} algorithms considered in the digits experiments.
For these algorithms we report the performances obtained in the ``Best single source'' and ``Unified sources`` settings, as available from~\cite{xu2018deep}.
As in the previous experiments, Multi-source DA with perfect domain knowledge can be regarded as a performance upper bound for our method.
Finally, we include results reported in~\cite{xu2018deep} for different multi-source DA models: Deep Cocktail Network (DCTN) \cite{xu2018deep}, the two shallow methods in~\cite{xie2015learning} (sFRAME) and~\cite{gopalan2011domain} (SGF), and an ensemble of baseline networks trained on each source domain separately (Source only).
These results are summarized in Table~\ref{tab:ablation-office}.

We note that, in this dataset, the improvements obtained by adopting a multi-source model instead of a single-source one are small.
This is in accordance with findings in~\cite{li2017deeper}, where it is shown that the domain shift in Office-31, when considering deep features, is indeed quite limited if compared to PACS, and it is mostly linked to changes in the background (Webcam-Amazon, DSLR-Amazon) or acquisition camera (DSLR-Webcam).
This is further supported by the smaller gap between DIAL and our method in this case compared to the previous experiments  {(average $p^*$ of 0.54)}. {In this setting, introducing our uniform loss term does not provides boost in performances. We ascribe this behaviour to the fact that in this scenario, each batch is built with a non-uniform number of samples per domain (following \cite{carlucci2017autodial}) while our current objective assumes a balanced sampling among domains. }

\begin{figure}[ht]
  \centering
  \includegraphics[width=0.8\columnwidth,height=0.1\textheight]{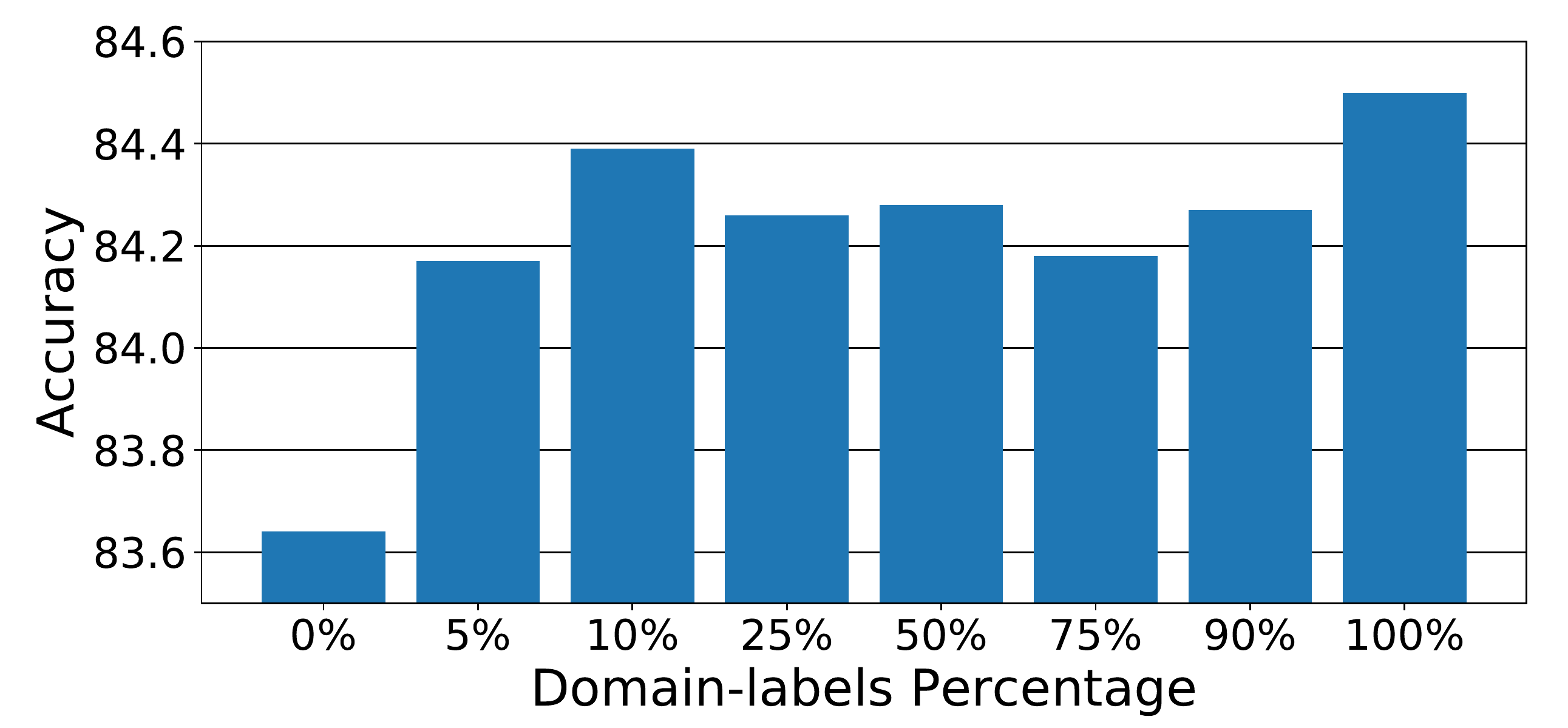}
  \caption{Office31 dataset. Performance at varying number of domain labels ($\%$) for source samples. 
  }
  \label{fig:office-bars}
  \vspace{-10pt}
\end{figure}

In a final Office-31 experiment, we consider a setting where the true domain of a subset of the source samples is known at training time. Figure~\ref{fig:office-bars} shows the average accuracy obtained when a different amount of domain labels are available.
Interestingly, by increasing the level of domain supervision the accuracy quickly saturates towards the value of Multi-source DA, completely filling the gap with as few as $5\%$ of the source samples.

\begin{table}[t]
			\caption{Office-31 dataset: comparison of different methods using AlexNet. In the first row we indicate the source (top) and the target domains (bottom).} 
		\centering
		\scalebox{.8}{
		\begin{tabular}{ l | l r | c | c | c | c  } 
			\hline
			& \multirow{2}{*}{Method}& Source& A-W & A-D & W-D  & \multirow{2}{*}{{Mean}}\\
           & & Target & D & W & A  & \\\hline
           \multirow{7}{*}{\parbox{1.5cm}{Best single source \cite{xu2018deep}}} 
           &TCA\cite{pan2011domain}&&95.2&93.2&51.6&68.8\\
           &GFK\cite{gong2012geodesic}&&95.0&95.6&52.4&68.7\\
           &DDC\cite{tzeng2015simultaneous}&&98.5&95.0&52.2&70.7\\
           &DRCN\cite{ghifary2016deep}&&99.0&96.4&56.0&73.6\\
           &RevGrad\cite{ganin2014unsupervised}&&99.2&96.4&53.4&74.3\\
           &DAN\cite{long2015learning}&&99.0&96.0&54.0&72.9\\
           &RTN\cite{long2016unsupervised}&&99.6&96.8&51.0&73.7\\\hline\hline
           \multirow{6}{*}{\parbox{1.5cm}{Unified sources}} &
           Source only from \cite{xu2018deep}&&98.1 &93.2 &50.2&80.5\\
           &Source only (ours) &		&94.6 &89.1	&49.1	&77.6\\
&{Source only+BN} (ours) &	&91.9&92.7	&46.5	&77.0\\
&RevGrad\cite{xu2018deep}&&\textbf{98.8}&\textbf{96.2}&54.6&83.2\\
&DAN\cite{xu2018deep}&&\textbf{98.8}&95.2&53.4&82.5\\
&Single BN &	&92.9&95.2	&60.1	&82.7\\
&DIAL \cite{carlucci2017just} &
&93.8 &94.3	&62.5	&83.5\\\hline
&Ours $\lambda_B=0$ &
	&93.7	&94.6 &\textbf{62.6}	&83.6\\
	&Ours &&93.6&93.6&62.4&83.2
	\\\hline\hline






            \hline
            \multirow{5}{*}{\parbox{1.5cm}{Multi-source}}&Source only \cite{xu2018deep}&&98.2&92.7&51.6&80.8\\
            &sFRAME\cite{xie2015learning}&&54.5&52.2&32.1&46.3\\
            &SGF\cite{gopalan2011domain}&&39.0&52.0&28.0&39.7\\
            &DCTN \cite{xu2018deep}&&99.6&96.9&54.9&83.8\\            
           &Multi-source DA	&	&94.8&95.8	&62.9	&84.5\\ \hline
		\end{tabular}
        }
        \label{tab:ablation-office}
\end{table}

\subsubsection{Comparison with S.o.t.A. on inferring latent domains}
In this section we compare the performance of our approach with previous works on DA which also consider the problem of inferring latent domains~\cite{hoffman2012discovering,xiong2014latent,gong2013reshaping}.
As stated in Section \ref{sec:related}, there are no previous works adopting deep learning models (i) in a multi-source setting and (ii) discovering hidden domains.
Therefore, the methods we compare to all employ handcrafted features.
For these approaches we report results taken from the original papers.
Furthermore, we evaluate the method of Gong~\etal~\cite{gong2013reshaping} using features from the last layer of the AlexNet architecture.
For a fair comparison, when applying our method we freeze AlexNet up to \texttt{fc7}, and apply mDA layers only after \texttt{fc7} and the classifier.

We first consider the Office-31 dataset, as this benchmark has been used in~\cite{hoffman2012discovering,xiong2014latent}, showing the results in Table~\ref{tab:sota-office}.
Our model outperforms all the baselines, with a clear margin in terms of accuracy.
Importantly, even when the method in~\cite{gong2013reshaping} is applied to features derived from AlexNet, still our approach leads to higher accuracy.  
For the sake of completeness, in the same table we also report results from previous multi-source DA methods~\cite{gopalan2014unsupervised,nguyen2015dash,lin2017cross} based on shallow models.
While these approaches significantly outperform \cite{hoffman2012discovering} and \cite{xiong2014latent}, still their accuracy is much lower than ours. 
{Moreover, introducing our novel loss term provides higher performances with respect to the our approach with $\lambda_B=0$. }

\begin{table}[t]
			\caption{Office-31: comparison with state-of-the-art algorithms. In the first row we indicate the source (top) and the target domains (bottom).} 
		\centering
		\scalebox{.9}{
		\begin{tabular}{ l@{\hspace{-6ex}} r | c | c | c | c } 
			\hline
			 \multirow{2}{*}{Method}& Sources &A-D & A-W & W-D  & \multirow{2}{*}{Mean}\\          
             &Target & W & D & A  & \\  \hline
            Hoffman \etal \cite{hoffman2012discovering}&&24.8	&42.7	&12.8	&26.8\\
            Xiong \etal \cite{xiong2014latent}&&29.3	&43.6	&13.3	&28.7\\
            \hline
            Gong \etal (AlexNet) \cite{gong2013reshaping}	&&91.8&{94.6}	&48.9	&78.4\\
         Ours $\lambda_B=0$	&&{93.1}&
         94.3&{64.2}& {83.9}\\
                  Ours	&&\textbf{94.5}&\textbf{94.9}&\textbf{64.9}&\textbf{84.8}\\
            \hline\hline
                             Gopalan \etal  \cite{gopalan2014unsupervised} &&51.3	&36.1	&35.8	&41.1\\
          Nguyen \etal  \cite{nguyen2015dash} &&64.5	&68.6	&41.8	&58.3\\
Lin \etal \cite{lin2017cross} &&73.2	&81.3	&41.1	& 65.2\\ \hline
		\end{tabular}
        }
		\label{tab:sota-office}
          
\end{table}

To provide a comparison in a multi-target scenario, we also consider the Office-Caltech dataset, comparing our model with \cite{hoffman2012discovering,gong2013reshaping}.
Following \cite{gong2013reshaping}, we test both single target (Amazon) and multi-target (Amazon-Caltech and Webcam-DSLR) scenarios.
As for the PACS multi-source\slash{}multi-target case, the assignment of each sample to the source or target set is assumed to be known, while the assignment to the specific domain is unknown.
We again want to remark that, since we do not assume to know the target domain to which a sample belongs, the task is even harder since we require a domain prediction step also at test time.
As in the Office-31 experiments, our approach outperforms all baselines, including the method in~\cite{gong2013reshaping} applied to AlexNet features.
{In this scenario, introducing our uniform loss provides a boost in performances in the multi-target setting, where the two source/target pairs have similar appearance. This is inline to what reported for the multitarget experiments on PACS (Table \ref{tab:pacs-multitarget}). }

\begin{table}[t]
 			\caption{Office-Caltech dataset: comparison with state-of-the-art algorithms. In the first row we indicate the source (top) and the target domains (bottom).
            } 
 		\centering
 		\scalebox{.85}{
 		\begin{tabular}{ l@{\hspace{-6ex}} r | c | c | c | c } 
        \hline
        		\multirow{2}{*}{Method}	 & Source& A-C & W-D & C-W-D  & \multirow{2}{*}{Mean}\\          
            & Target& W-D & A-C & A  & \\
 			\hline
                Gong \etal \cite{gong2013reshaping} - original& &41.7	&35.8	&41.0	&39.5\\
                Hoffman \etal \cite{hoffman2012discovering} - ensemble	&&31.7	&34.4	&38.9	& 35.0\\
               Hoffman \etal \cite{hoffman2012discovering} - matching &&39.6	&34.0	&34.6	&36.1\\
               Gong \etal \cite{gong2013reshaping} - ensemble &&38.7	&35.8	&42.8	&39.1\\
                Gong \etal \cite{gong2013reshaping}	- matching &&42.6	&35.5	&44.6	&40.9\\
             \hline
           Gong \etal (AlexNet) \cite{gong2013reshaping} - ensemble&&{87.8}	&87.9	&93.6	&89.8\\
{Ours $\lambda_B=0$}&&{93.5}&	{88.2}&{93.7}&{91.8}\\
{Ours}&&\textbf{95.0}&	\textbf{88.7}&\textbf{93.9}&\textbf{92.5}\\
             \hline
 		\end{tabular}
         }
 		\label{tab:sota-office-caltech}
 \end{table}


\section{Conclusions}
\label{sec:conclusions}
In this work we presented a novel deep DA model for automatically discovering latent domains within visual datasets. 
The proposed deep architecture is based on a side-branch which computes the assignment of source and target samples to their associated latent domain. These assignments are then exploited within the main network by novel domain alignment layers which reduce the domain shift by aligning the feature distributions of the discovered sources and the target domains.
Our experimental results demonstrate the ability of our model to efficiently exploit the discovered latent domains for addressing challenging domain adaptation tasks. 
Future works will investigate other architectural design choices for the domain prediction branch, as well as the possibility to integrate it into other CNN models for unsupervised domain adaptation \cite{ganin2014unsupervised}.


%



\ifCLASSOPTIONcompsoc
  \section*{Acknowledgments}
\else
 \section*{Acknowledgment}
\fi
\vspace{-4pt}We acknowledge financial support from ERC grant 637076 - RoboExNovo and project \textit{DIGIMAP}, grant 860375, funded by the Austrian Research Promotion Agency (FFG). This work was carried out under the "Vision and Learning joint Laboratory" between FBK and UNITN.
\vspace{-8pt}

\ifCLASSOPTIONcaptionsoff
  \newpage
\fi



%

\bibliographystyle{ieee}
\bibliography{root}

%
\vspace{-32pt}
\begin{IEEEbiography}[{\includegraphics[width=1in,height=1.25in,clip,keepaspectratio]{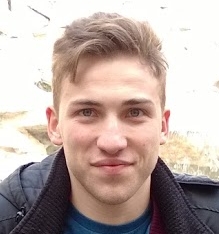}}]{Massimiliano~Mancini}
Massimiliano Mancini is a Ph. D. student with Sapienza University of Rome, in collaboration with Fondazione Bruno Kessler (FBK). He received is master degree in Artificial Intelligence and Robotics form Sapienza University of Rome in 2016. Currently, he is a member of the Visual Learning and Multimodal Applications Laboratory, led by Prof. Barbara Caputo, and the Technologies of Vision Laboratory in FBK. He was also visiting Ph.D. student in the Robotics, Perception and Learning Laboratory at KTH Royal Institute of Technology in Stockholm. 
\end{IEEEbiography}
\vspace{-33pt}
\begin{IEEEbiography}[{\includegraphics[width=1in,height=1.25in,clip,keepaspectratio]{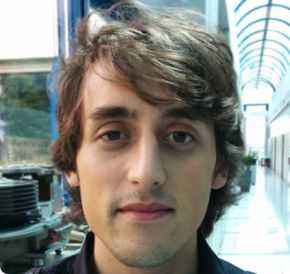}}]{Lorenzo~Porzi}
Lorenzo Porzi is a researcher with Mapillary Research, focusing on deep learning applied to semantic and 3D vision.
He received his PhD in computer science from the University of Perugia and Fondazione Bruno Kessler (FBK) in 2016.
Before joining Mapillary Research, Lorenzo held a post-doctoral research position at IRI-UPC in Barcelona, working with the group of Dr. Francesc Moreno-Noguer.
While in FBK, he participated in the VENTURI and REPLICATE EU projects.
\end{IEEEbiography}
\vspace{-33pt}
\begin{IEEEbiography}[{\includegraphics[width=1in,height=1.25in,clip,keepaspectratio]{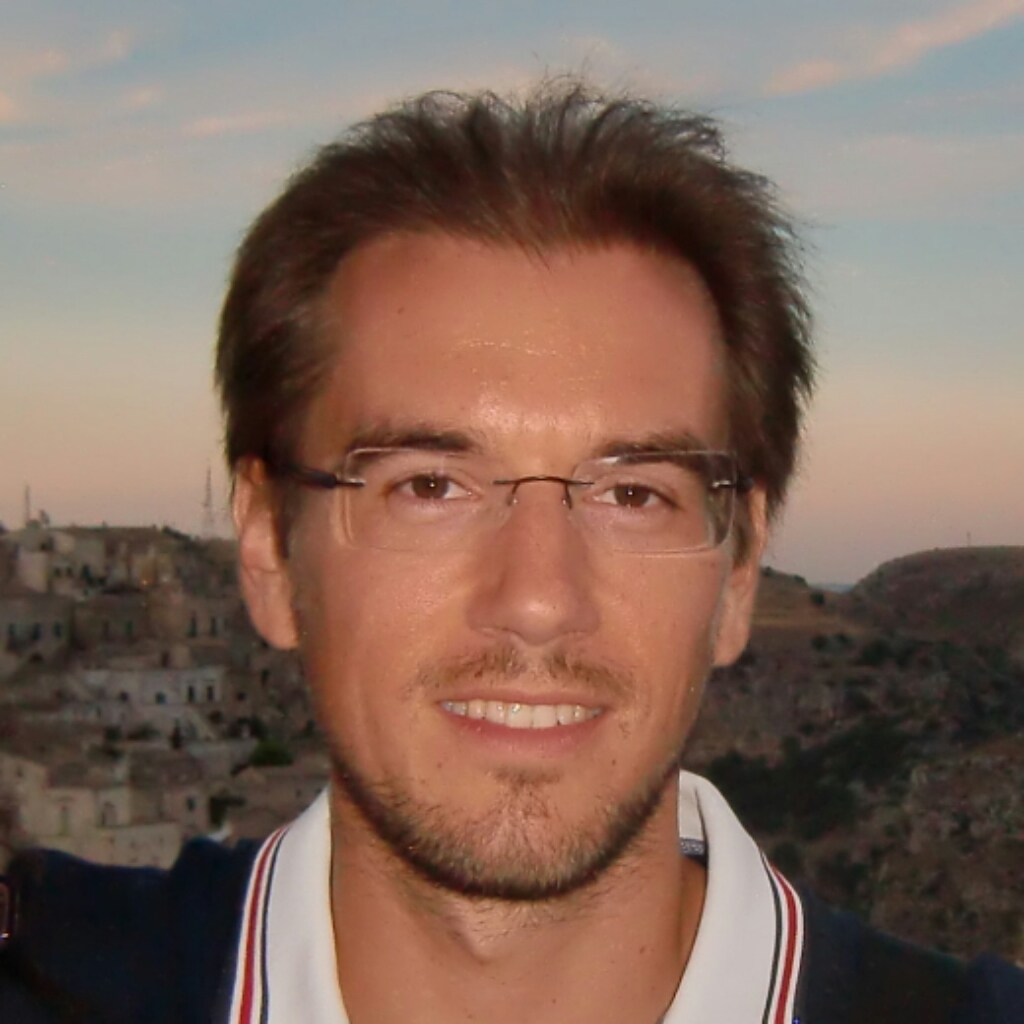}}]{Samuel~Rota~Bul\'o}
Samuel Rota Bul\'o received the PhD in computer science at University of Venice in 2009. He worked there as PostDoc and held teaching positions until 2013. He then became a researcher for FBK in computer vision and machine learning. Since 2017, he is a senior researcher with Mapillary Research. He was awarded the prestigious Marr Prize in 2015. He serves on the editorial board for "Pattern Recognition" and “International Journal of Machine Learning and Cybernetics”  and is regularly on the program committee of international conferences of his field. He participated to several  EU projects (SIMBAD, VENTURI, REPLICATE).
\end{IEEEbiography}
\vspace{-33pt}
\begin{IEEEbiography}[{\includegraphics[width=1in,height=1.25in,clip,keepaspectratio]{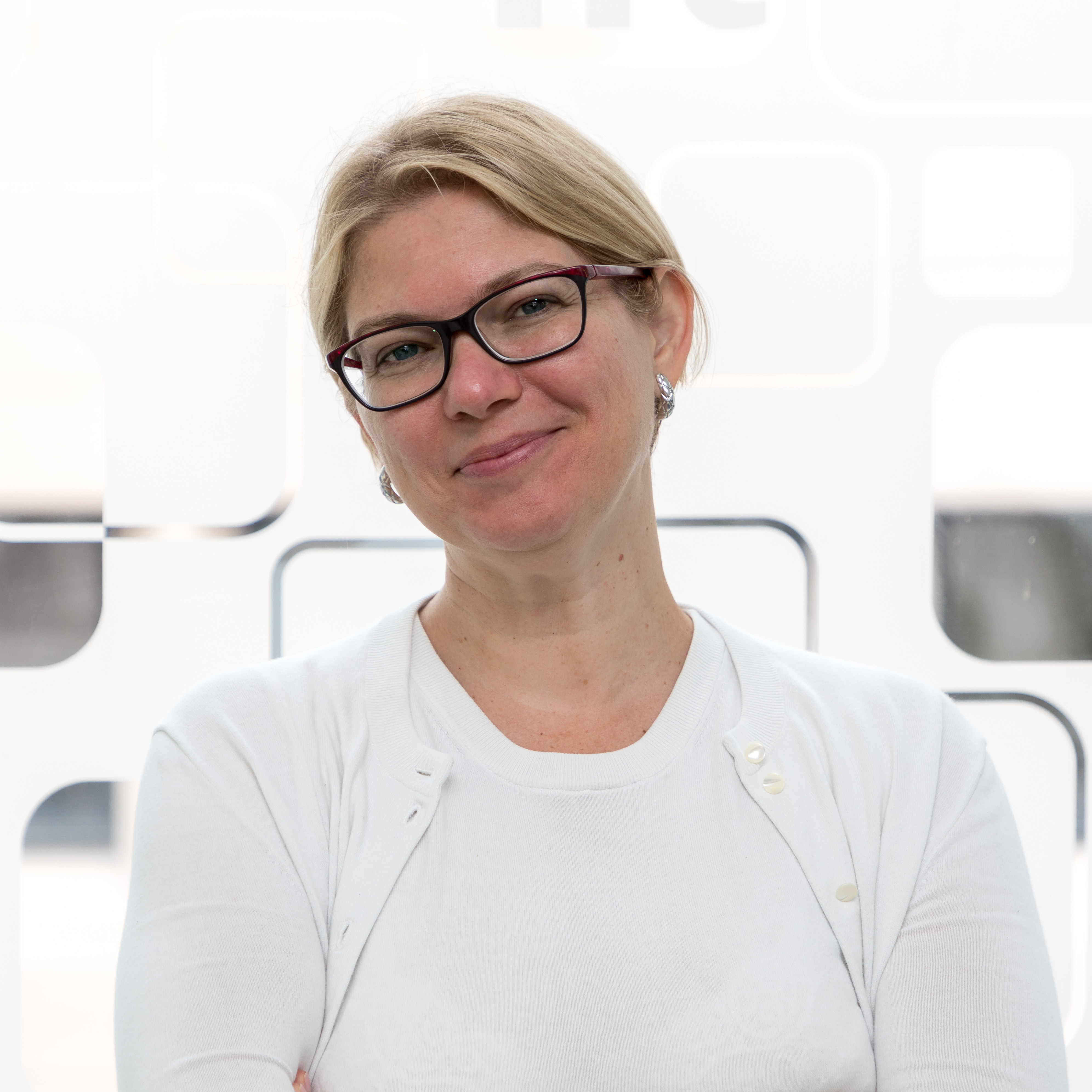}}]{Barbara~Caputo}
Barbara Caputo is Full Professor at the  DAUIN Department of Control and Computer Engineering of Politecnico di Torino and Principal Investigator at the Italian Institute of Technology (IIT), where she leads the Visual Learning and Multimodal Applications Laboratory (VANDAL).  Her main research interest is to develop algorithms for learning, recognition and categorization of visual and multimodal patterns for artificial autonomous systems. These features are crucial to enable robots to represent and understand their surroundings, to learn and reason about it, and ultimately to equip them with cognitive capabilities. Her research is sponsored by the Swiss National Science Foundation (SNSF), the Italian Ministry for Education, University and Research (MIUR), the European Commission (EC) and the European Research Council (ERC).
\end{IEEEbiography}
\vspace{-33pt}
\begin{IEEEbiography}[{\includegraphics[width=1in,height=1.25in,clip,keepaspectratio]{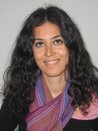}}]{Elisa~Ricci}
Elisa Ricci is an associate professor at University of Trento and a researcher at Fondazione Bruno Kessler. She received her PhD from the University of Perugia in 2008. She has since been a post-doctoral researcher at Idiap Research Institute and an assistant professor at University of Perugia. She was also a visiting researcher at University of Bristol. Her research interests
are mainly in the areas of computer vision and machine learning.
\end{IEEEbiography}

\appendix
\section{mDA layers formulas}
\label{sec:formulas}
From the main text, we have the output of our mDA layer denoted by
\begin{equation}
y_{i}=\mDA(x_i, \vct{w}_i; \vct{\hat{\mu}}, \vct{\hat{\sigma}})=\sum_{d\in \set{D}} w_{i,d} \hat{x}_{i,d},
\end{equation}
where, for simplicity:
\begin{equation}
\hat{x}_{i,d}=\frac{x_i - \hat{\mu}_d}{\sqrt{\hat{\sigma}_d^2 + \epsilon}},
\end{equation}
and the statistics are given by
\begin{equation}
\begin{aligned}
  \hat{\mu}_d &= \sum_{i=1}^{\con{b}} \hat{w}_{i,d} x_i, \\
  \hat{\sigma}_d^2&=\sum_{i=1}^{\con{b}} \hat{w}_{i,d} (x_i-\hat{\mu}_d)^2,
\end{aligned}
\end{equation}
where
$\hat{w}_{i,d}=w_{i,d}/\sum_{j=1}^{\con{b}}w_{j,d}$.

From the previous equations we can derive the partial derivative of the loss function with respect to both the input $x_i$ and the domain assignment probabilities $w_{i,d}$.
Let us denote $\frac{\partial L}{\partial y_i}$ the partial derivative of the loss function $L$ with respect to the output $y_i$ of the mDA layer.
We have:
\begin{equation}
\begin{aligned}
  \frac{\partial \hat{x}_{i,d}}{\partial \hat{\sigma}_{d^*}^2} &=
    -\ind{d=d^*} \frac{1}{2} (x_i - \hat{\mu}_{d^*}) \cdot (\hat{\sigma}_{d^*}^2 + \epsilon)^{-\frac{3}{2}}, \\
  \frac{\partial \hat{x}_{i,d}}{\partial \hat{\mu}_{d^*}} &=
    -\ind{d=d^*} (\hat{\sigma}_{d^*}^2 + \epsilon)^{-\frac{1}{2}},
\end{aligned}
\end{equation}
and
\begin{equation}
\begin{aligned}
  \frac{\partial \hat{\sigma}_d^2}{\partial x_{i}} &=
    2\,\hat{w}_{i,d}\cdot(x_i-\hat{\mu}_d), &
  \frac{\partial \hat{\mu}_d}{\partial x_{i}} &=
    \hat{w}_{i,d}.
\end{aligned}
\end{equation}
Thus, the partial derivative of $L$ w.r.t. the input $x_i$ is:
\begin{equation}
  \frac{\partial L}{\partial x_{i^*}} =
    \sum_{d\in \set{D}} \frac{w_{i^*,d}}{\sqrt{\hat{\sigma}_d^2 + \epsilon}} \left[
      \frac{\partial L}{\partial y_{i^*}} - A_d - \hat{x}_{i^*,d} B_d
    \right],
\end{equation}
where:
\begin{equation}
\label{eqn:a-b}
\begin{aligned}
  A_d &= \sum_{i=1}^\con{b} \hat{w}_{i,d} \frac{\partial L}{\partial y_{i}}, \\
  B_d &= \sum_{i=1}^\con{b} \hat{w}_{i,d} \hat{x}_{i,d} \frac{\partial L}{\partial y_{i}}.
\end{aligned}
\end{equation}
For the domain assignment probabilities $w_{i,d}$ we have:
\begin{align}
  \frac{\partial \hat{\mu}_d}{\partial \hat{w}_{i,d^*}} &= \ind{d=d^*} x_{i}, \\
  \frac{\partial \hat{\sigma}_d^2}{\partial \hat{w}_{i,d^*}} &= \ind{d=d^*} (x_i-\hat{\mu}_d)^2, \\
  \frac{\partial \hat{w}_{i,d}}{\partial w_{i^*,d^*}} &=
    \ind{d=d^*} \frac{\ind{i=i^*} \sum_{j=1}^{\con{b}}w_{j,d} - w_{i,d}}{(\sum_{j=1}^{\con{b}}w_{j,d})^2}.
\end{align}
Thus, the partial derivative of $L$ w.r.t. $w_{i,d}$ is:
\begin{equation}
  \frac{\partial L}{\partial w_{i^*,d}} = \hat{x}_{i^*,d} \left(
    \frac{\partial L}{\partial y_{i^*}} - A_d
  \right) - \frac{1}{2} \left(
    \hat{x}_{i^*,d}^2 - \frac{\sigma_d^2}{\sigma_d^2 + \epsilon}
  \right) B_d,
\end{equation}
where $A_d$ and $B_d$ are defined as in \eqref{eqn:a-b}.

\section{Training loss progress}
\begin{figure}[t]
 \centering
 \subfloat[MNIST-m as target]
  {\includegraphics[width=0.25\textwidth]{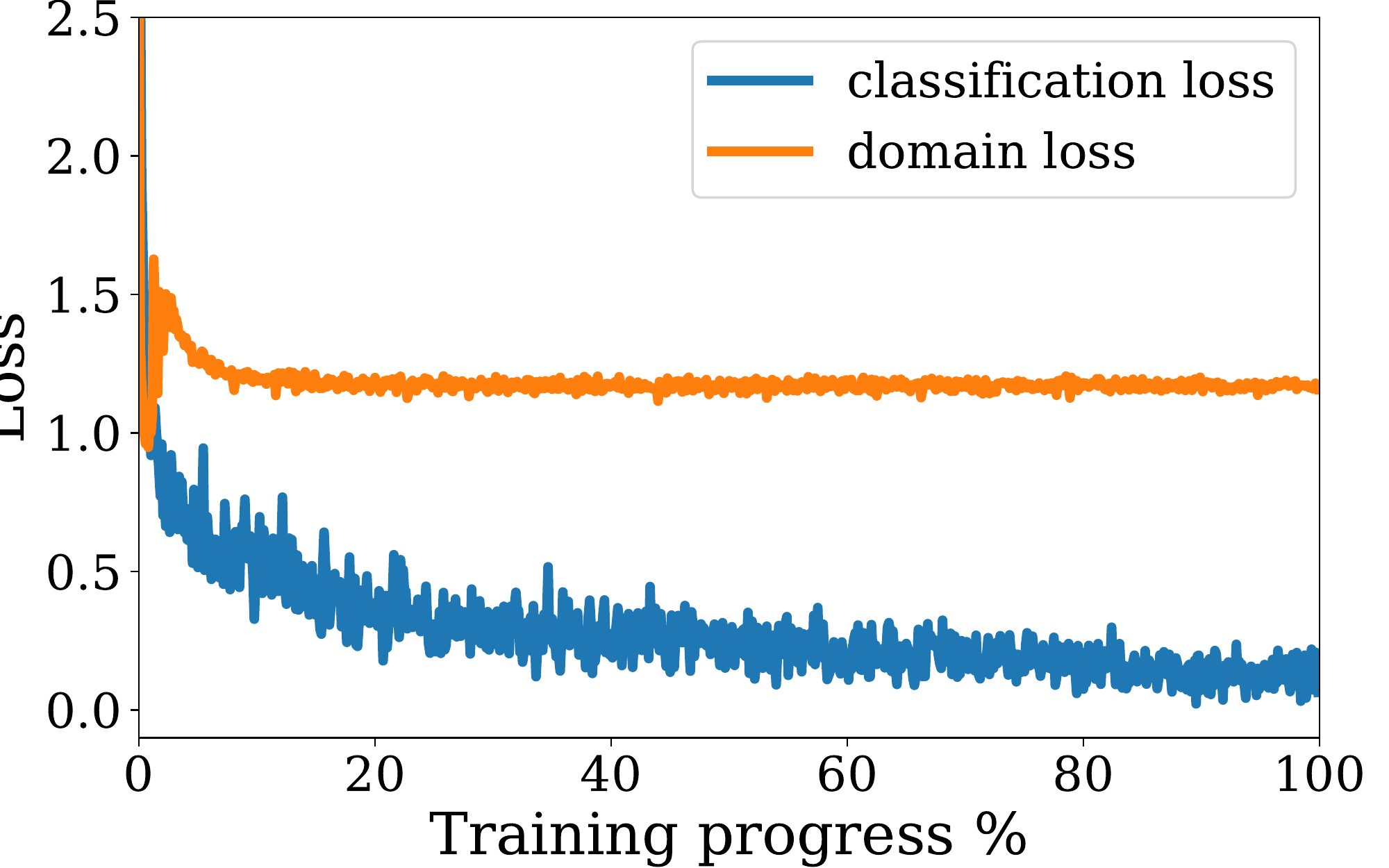}}
 \subfloat[SVHN as target]
  {\includegraphics[width=0.25\textwidth]{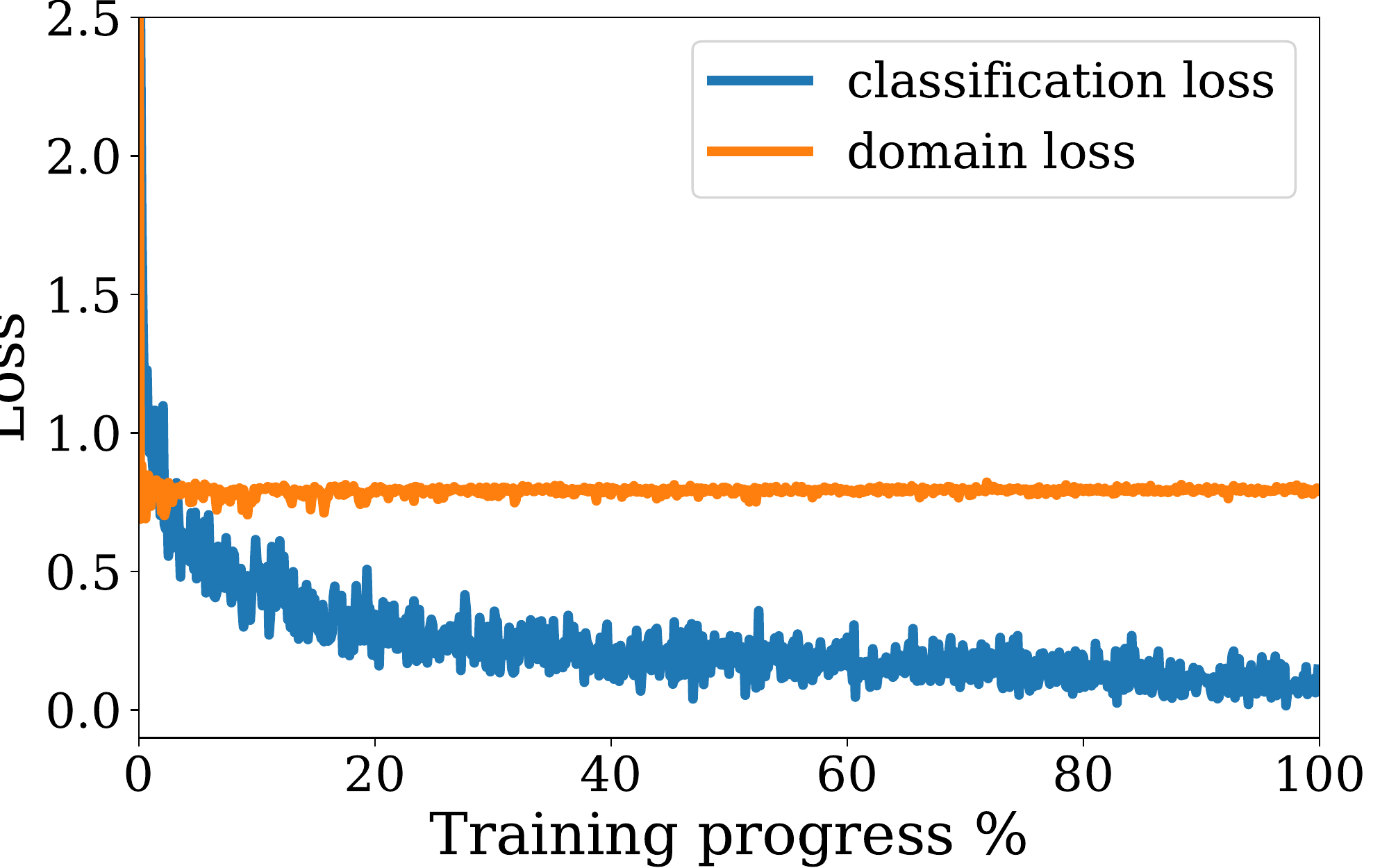}}
  \caption{{Digits-five: plots of the domain (orange) and classification (blue) losses during the training phase.}
  }
  \label{fig:losses-unified}
\end{figure}

\begin{figure}[t]
 \centering
 \subfloat[MNIST-m as target]
  {\includegraphics[width=0.25\textwidth]{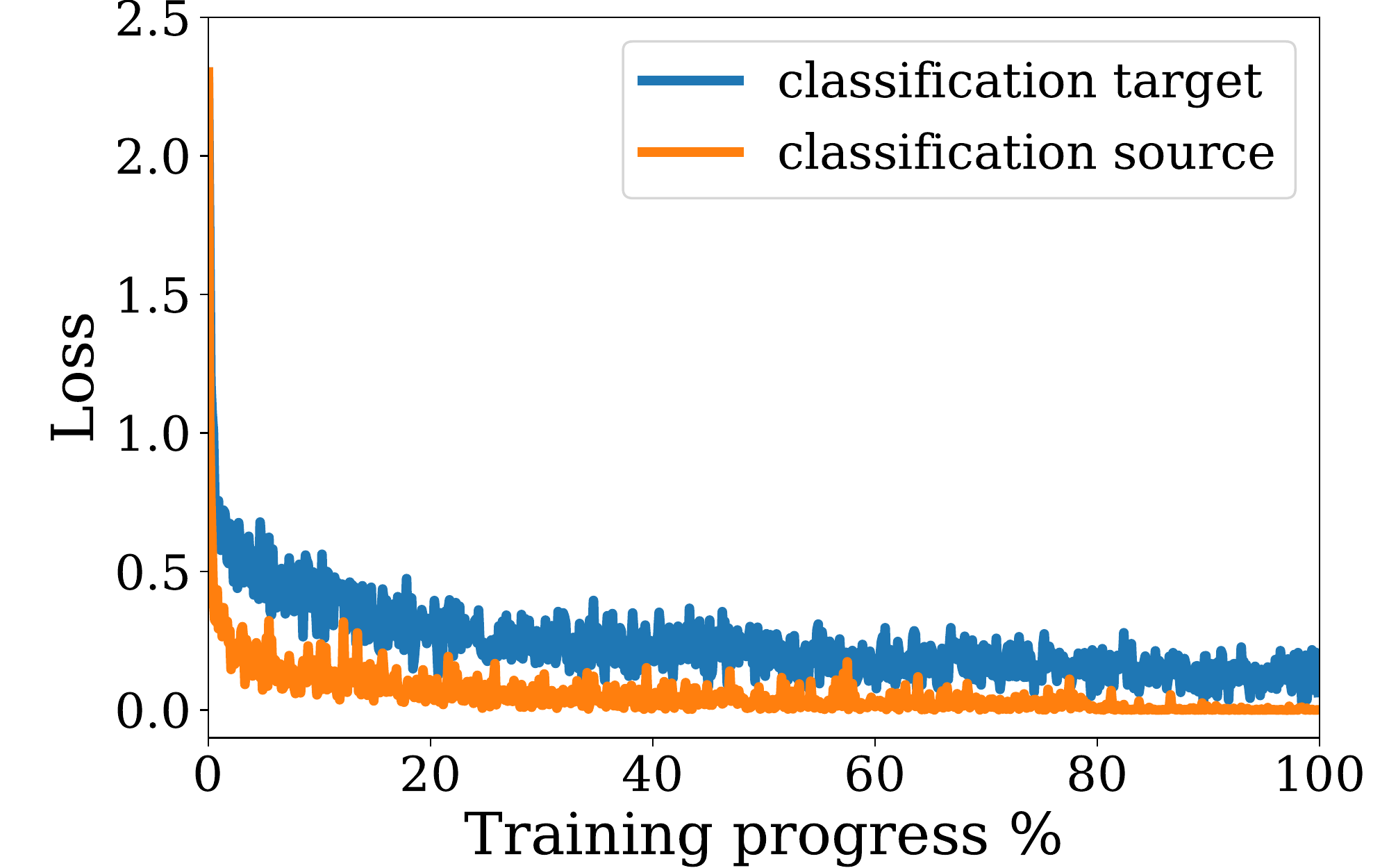}}
 \subfloat[SVHN as target]
  {\includegraphics[width=0.25\textwidth]{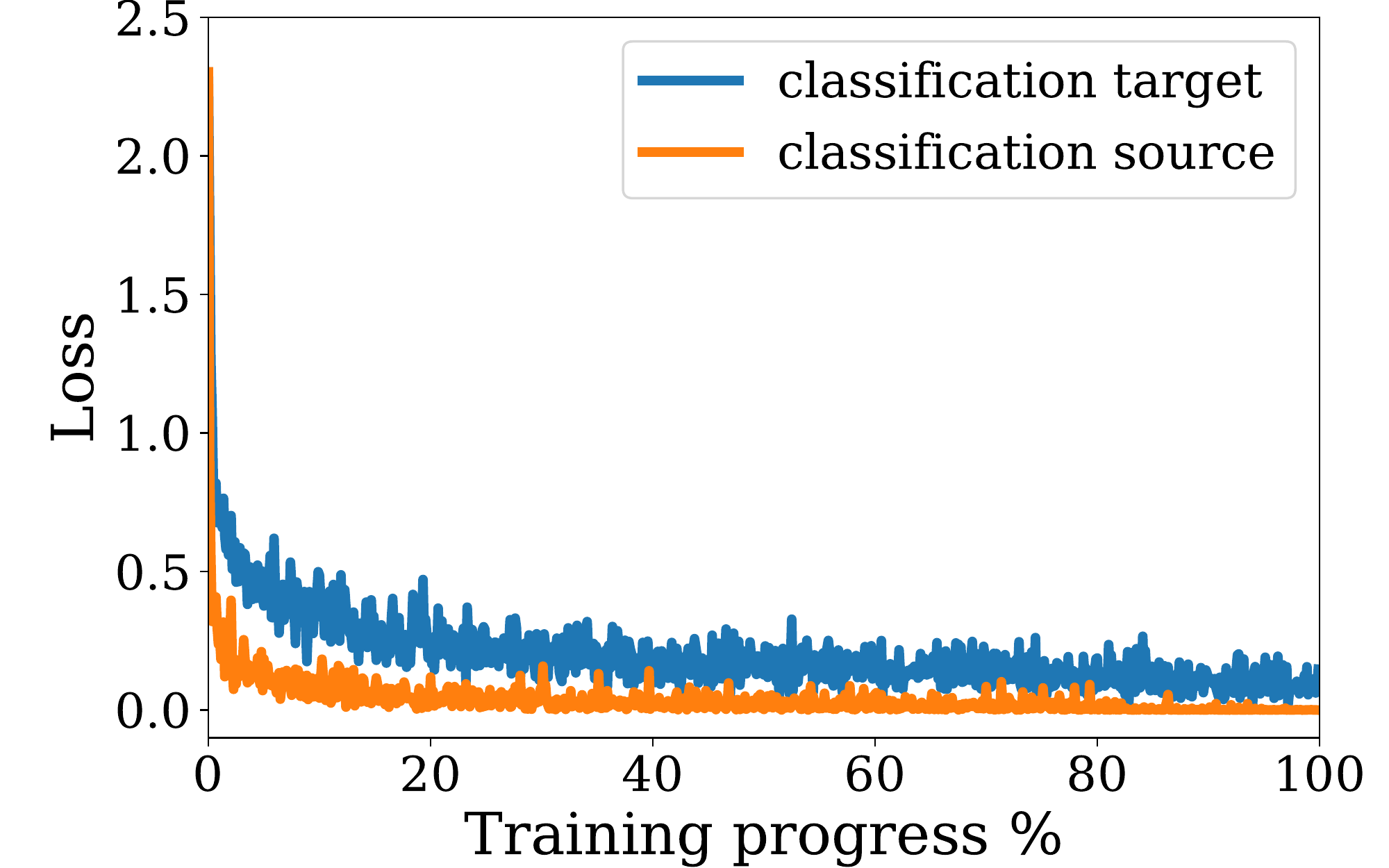}}
  \caption{{Digits-five: plots of the cross-entropy loss on source samples (orange) and entropy loss on target sample (blue) for the semantic classifier during the training phase.}
  }
  \label{fig:losses-semantic}
\end{figure}

\begin{figure}[h]
 \centering
 \subfloat[MNIST-m as target]
  {\includegraphics[width=0.25\textwidth]{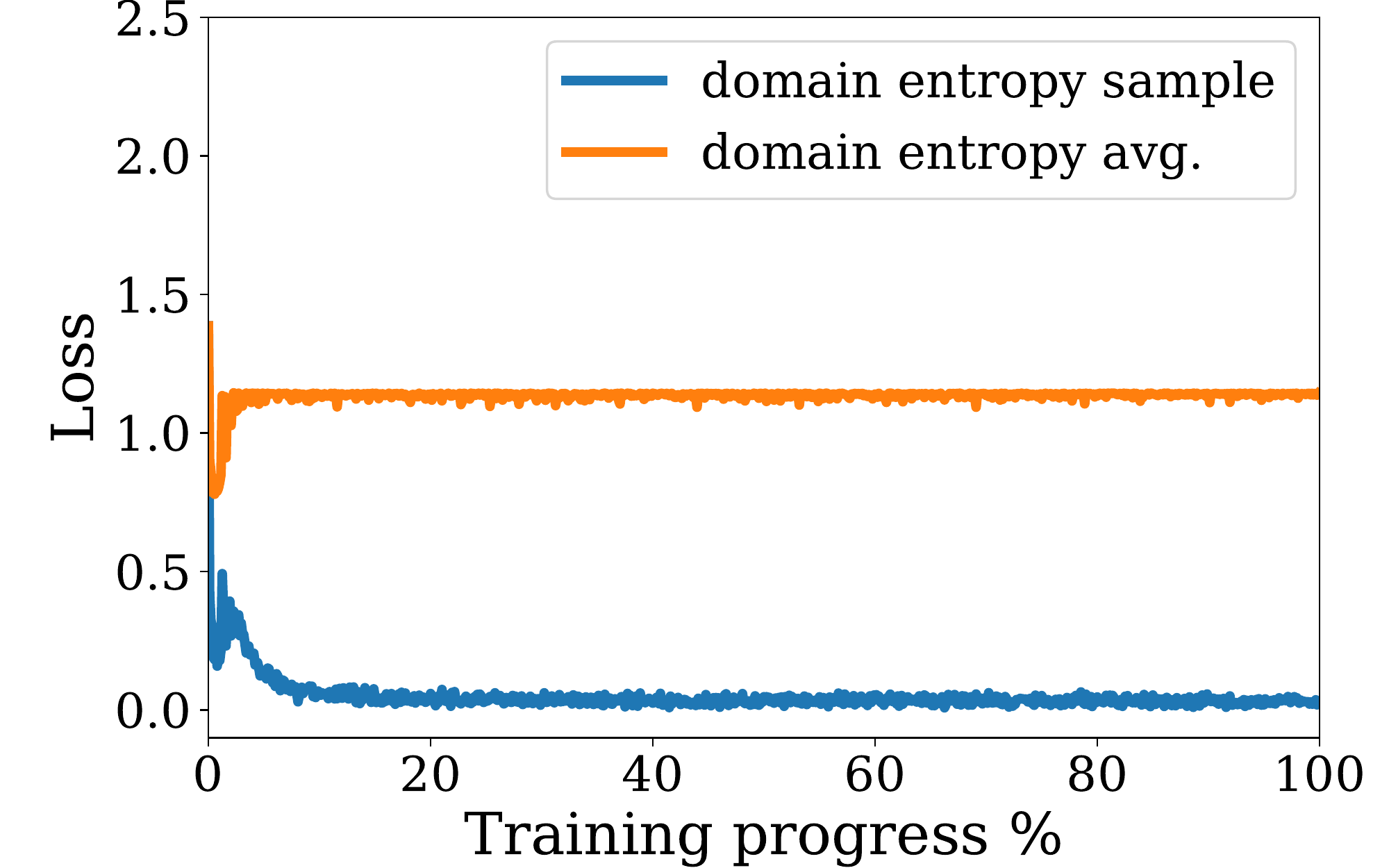}}
 \subfloat[SVHN as target]
  {\includegraphics[width=0.25\textwidth]{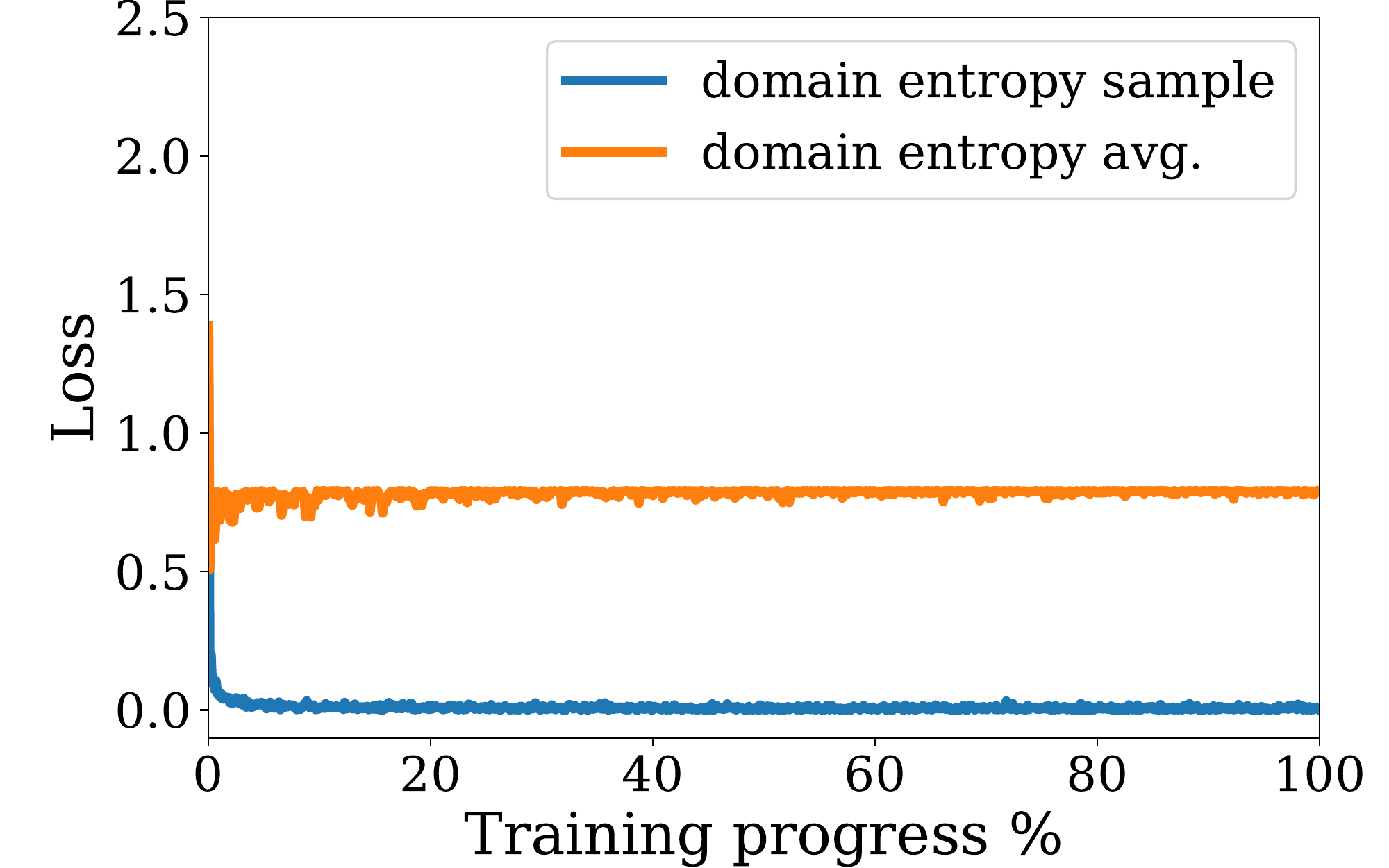}}
  \caption{{Digits-five: plots of the entropy loss on single sample (blue) and on the average batch assignments (orange) for the domain classifier during the training phase.}
  }
  \label{fig:losses-domain}
\end{figure}

{In this section, we plot the losses as the training progresses for the Digits-five experiments. The plots are shown in Figure \ref{fig:losses-unified}. For both MNIST-m and SVHN, the classification loss smoothly decreases, while the domain loss first decreases and then stabilizes around a fixed value. This is a consequence of the introduced balancing term on the domain assignments, which enforces the entropy to be low for the assignment of a single sample, but high for the assignments averaged across the entire batch. In Figures \ref{fig:losses-semantic} and \ref{fig:losses-domain} we plot the single components of the classification and domain loss respectively. For the semantic part (Figure \ref{fig:losses-semantic}), both the entropy loss on target sample and the cross-entropy loss on source samples decrease smoothly. For the domain assignment part (Figure \ref{fig:losses-domain}), we can see how the entropy loss on single samples rapidly decreases, while the average batch assignment keeps an high entropy, as expected. We highlight that when SVHN is used as target, the source domains are a bit closer to each other in appearance, thus the average batch entropy has a slightly lower value (\ie the assignments are less balanced) with respect to the MNIST-m as target case.

Finally, it is worth noticing that the domain loss reaches a stable value earlier than the classification components. This is a design choice, since we want to learn a semantic predictor on stable and confident domain assignments.}

\section{Additional Results on PACS}
{
A crucial problem in domain adaptation rarely addressed in the literature is how to tune model hyper-parameters. In fact, setting the hyper-parameters values based on the performance on the source domain is suboptimal, due to the domain shift. Furthermore, assuming the presence of a validation set for the target domain is not realistic in practice \cite{morerio2017minimal}: in unsupervised domain adaptation we only assume the presence of a set of unlabelled target data. Despite recent research in this direction \cite{morerio2017minimal}, there is no clear solution to this problem in the literature. This problem is more severe in our case, since it is not trivial to define a validation set for the latent domain discovery problem, due to the assumption that multiple source and target domains are mixed.

\begin{table}[t]
			\caption{PACS dataset: comparison of different methods using the ResNet architecture. The first row indicates the target domain, while all the others are considered as sources. {The numbers in parenthesis indicate the results using a target validation set for model selection.}} 
		\centering
		\scalebox{0.8}{
		\begin{tabular}{ l | c  c  c  c | c  } 
			\hline
			Method & Sketch & Photo & Art & Cartoon & Mean \\
            	\hline
                ResNet \cite{he2016deep} &60.1&92.9&74.7&72.4&75.0\\
DIAL \cite{carlucci2017just} &66.8 (71.3)&\textbf{97.0} (\textbf{97.4})&87.3 (87.5)&85.5 (87.0)&84.2 (85.8)\\
Ours   &\textbf{70.7} (\textbf{75.2})&\textbf{97.0} (97.3)&87.4 (\textbf{87.7})&86.3 (\textbf{87.2})&\textbf{85.4} (\textbf{86.9})\\\hline\hline
Multi-source DA & 71.6 (78.1)& 96.6 (97.2) & 87.5 (88.7) & 87.0 (87.4) & 85.7 (87.9) \\ \hline 
\hline
		\end{tabular}
        }
		\label{tab:pacs-val}
\end{table}

\begin{table*}[h!]
			\caption{PACS dataset: comparison of different methods using the ResNet architecture on the multi-source multi-target setting. The first row indicates the two target domains. {The numbers in parenthesis indicate the results using a target validation set for model selection.} }
		\centering
		\scalebox{.95}{
		\begin{tabular}{ l | c  c c  c c c | c } 
			\hline
			Method  & Photo-Art & Photo-Cartoon & Photo-Sketch&Art-Cartoon&Art-Sketch&Cartoon-Sketch & Mean\\
            	\hline
                ResNet \cite{he2016deep} &71.4&84.2&81.4&62.2&70.3&54.2&70.6\\
DIAL \cite{carlucci2017just} &{86.7} (87.5)&86.5 (87.1)&86.8 (88.2)&77.1 (78.7)&72.1 (74.2)&67.7 (70.4)&79.5 (81.0)\\


Ours &\textbf{87.2} (\textbf{87.7}) &\textbf{88.1} (\textbf{88.5})&\textbf{88.7} (\textbf{89.7})&77.7 (\textbf{79.6})&\textbf{81.3} (\textbf{82.2})&\textbf{77.0} (\textbf{79.3})&\textbf{83.3} (\textbf{84.5})\\\hline\hline
Multi-source/target DA & 87.7 (88.8)&88.9 (89.8)&86.8 (88.3)&79.0 (79.5)&79.8 (82.2)&75.6 (79.1)&83.0 (84.6)\\ \hline 
		\end{tabular}
        }
         \vspace{-10pt}
		\label{tab:pacs-multitarget-val}
\end{table*}

Nonetheless, for the sake of completeness, we analyze the performances of our model and the baselines if we assume the presence of a target validation set to perform model selection. We consider the PACS dataset, in both the single and multi-target scenarios. The results are reported in parenthesis in Table \ref{tab:pacs-val} and in Table \ref{tab:pacs-multitarget-val}. While our model and the baselines obviously benefit from the validation set, the overall trends remain the same, with our model achieving higher performances with respect to the baseline and close to the multi-source upper bound. Notice that a validation set is especially beneficial in the case of consistent domain shift: for instance, all the methods increase their results by almost 5\% in Table \ref{tab:pacs-val} when Sketch is the target domain.

As a final note, we underline that the use of a validation set on the target domain for unsupervised domain adaptation is not a common practice in the community, thus these results can be regarded as an upper bound with respect to our model. 
}




\end{document}